\documentclass[10pt]{article}
\usepackage{arxiv}
\usepackage[english]{babel} 

\usepackage[T1]{fontenc}    

\usepackage[utf8]{inputenc} 

\usepackage{multirow}
\usepackage{amsfonts}       
\usepackage{nicefrac}       
\usepackage{microtype}      
\usepackage{dsfont}         
\usepackage{mathrsfs}       
\usepackage{bm}             
\usepackage{xcolor}         
\usepackage{xspace}         

\usepackage{algorithmic}
\usepackage{algorithm}
\usepackage[]{amsthm, thmtools, thm-restate}
\usepackage[]{amssymb}      
\usepackage[]{mathtools}    

\usepackage[round, semicolon, authoryear]{natbib} 

\usepackage{physics}

\usepackage{cancel}         
\usepackage{centernot}      

\usepackage{booktabs}       
\usepackage{graphicx}       
\usepackage[]{tikz}         
\usepackage[]{pgfplots}     
\usepackage{adjustbox}

\usepackage{float}          
\usepackage{caption}
\usepackage{wrapfig}    
\usepackage{subcaption} 
\usepackage{url}                               
\usepackage{hyperref}                          
\hypersetup{
  colorlinks, citecolor=teal, linkcolor=blue
}

\usepackage{nameref}                           
\usepackage[capitalize, nameinlink]{cleveref}  
\usepackage{crossreftools}
\crefformat{footnote}{#2\footnotemark[#1]#3}          

\crefformat{problem}{#2 Prob.~(#1)#3}

\crefname{Problem}{Problem}{Problem}

\newcommand{\define}{\coloneqq}

\newcommand{\subto}{\text{subject to }}

\newcommand{\avg}[1]{\overline{#1}}

\DeclareMathOperator*{\E}{E} 
\DeclareMathOperator*{\I}{I}   

\DeclareMathOperator*{\argmin}{argmin}

\newcommand{\PP}{\mathbf{P}} 
\newcommand{\ZZ}{\mathbf{Z}} 
\newcommand{\RR}{\mathbf{R}} 

\newcommand{\eg}{{e.g.}\xspace}  
\newcommand{\ie}{{i.e.}\xspace}  
\newcommand{\etc}{{etc.}\xspace} 
\newcommand{\viz}{{viz.}\xspace} 

\AtBeginDocument{}

  {%
   \par\noindent{\bfseries\upshape Proof\ }%
  }%
  {}

\pgfplotsset{compat=1.17}

\title{Long-Term Fairness with Unknown Dynamics}

\author{%
  Tongxin Yin\footnotemark[1] \\
  Computer Science and Engineering\\
  University of Michigan \\
  Ann Arbor, MI 48109 \\
  \texttt{tyin@umich.edu}
  \And
  Reilly Raab\footnotemark[1] \\
  Computer Science and Engineering\\
  University of California, Santa Cruz\\
  Santa Cruz, CA 95064 \\
  \texttt{reilly@ucsc.edu} \\
  \AND
  Mingyan Liu \\
  Computer Science and Engineering\\
  University of Michigan \\
  Ann Arbor, MI 48109 \\
  \texttt{mingyan@umich.edu}
  \And
  Yang Liu \\
  Computer Science and Engineering\\
  University of California, Santa Cruz\\
  Santa Cruz, CA 95064 \\
  \texttt{yangliu@ucsc.edu}
}

\theoremstyle{plain}
\newtheorem{theorem}{Theorem}[section]
\newtheorem*{theorem*}{Theorem}

\newtheorem{lemma}[theorem]{Lemma}
\newtheorem{corollary}[theorem]{Corollary}
\theoremstyle{definition}
\newtheorem{definition}[theorem]{Definition}
\newtheorem{assumption}[theorem]{Assumption}
\theoremstyle{remark}

\newcommand{\ucbfair}[0]{{\texttt{L-UCBFair}}\xspace}
\newcommand{\drl}[0]{{\texttt{R-TD3}}\xspace}

\newcommand{\tp}{\texttt{tp}\xspace}
\newcommand{\tn}{\texttt{tn}\xspace}
\newcommand{\DP}{\texttt{DP}\xspace}
\newcommand{\EO}{\texttt{EO}\xspace}
\newcommand{\EOp}{\texttt{EOp}\xspace}
\newcommand{\QR}{\texttt{QR}\xspace}

\newcommand{\action}{a}
\newcommand{\state}{s}

\begin{document}

\maketitle
\renewcommand{\thefootnote}{\fnsymbol{footnote}}
\footnotetext[1]{These authors contributed equally to this work.}

\begin{abstract}
While machine learning can myopically reinforce
social inequalities, it may also be used to dynamically seek equitable outcomes.
In this paper, we formalize long-term fairness in the context of online
reinforcement learning. This formulation can accommodate dynamical control objectives,
such as driving equity inherent in the \emph{state} of a population, that cannot
be incorporated into static formulations of fairness.  We demonstrate that this
framing allows an algorithm to adapt to unknown dynamics by sacrificing
short-term incentives to drive a classifier-population system
towards more desirable  equilibria. For the proposed setting, we develop an
algorithm that adapts recent work in online learning.  We prove that this
algorithm achieves simultaneous probabilistic bounds on cumulative loss and
cumulative violations of fairness (as statistical regularities between
demographic groups). We compare our proposed algorithm to the repeated
retraining of myopic classifiers, as a baseline, and to a deep reinforcement
learning algorithm that lacks safety guarantees. Our experiments model human
populations according to evolutionary game theory and integrate real-world
datasets.
\end{abstract}

\section{Introduction} \label{sec:intro}
As machine learning (ML) algorithms are deployed for tasks with real-world social
consequences (\eg, school admissions, loan approval, medical interventions,
\etc), the possibility exists for runaway social inequalities
\citep{crawford2016there, chaney2018algorithmic, fuster2018predictably,
ensign2018runaway}. While ``fairness'' has become a salient ethical concern in
contemporary research, the closed-loop dynamics of real-world systems comprising ML policies and populations that mutually adapt to each other (\cref{fig:overview} in the supplementary material) remain poorly understood.

In this paper, our primary contribution is to consider the problem of
{\em long-term fairness}, or algorithmic fairness in the context of a dynamically
responsive population, as a reinforcement learning (RL) problem subject to
constraint. In our formulation, the central learning task is to develop a policy
that minimizes cumulative loss (\eg, financial risk, negative educational
outcomes, misdiagnoses, \etc) incurred by an ML agent interacting with a human
population up to a finite time horizon, subject to constraints on cumulative
``violations of fairness'', which we refer to in a single time step as
\emph{disparity} and cumulatively as \emph{distortion}.

Our central hypothesis is that an RL formulation of
long-term fairness may allow an agent to learn to \textbf{sacrifice short-term
utility in order to drive the system towards more desirable  equilibria}. The core
practical difficulties posed by our general problem formulation, however, are
the potentially unknown dynamics of the system under control, which must be
determined by the RL agent online (\ie, during actual deployment), and the
general non-convexity of the losses or constraints considered. Additionally, we
address continuous state and action spaces, in general, which preclude familiar
methods with performance guarantees in discrete settings.

Our secondary contributions are 1) to show that  long-term fairness can be solved within asymptotic, probabalistic bounds under certain dynamical
assumptions and 2) to demonstrate that the problem of long-term fairness can also be addressed more flexibly. For theoretical guarantees, we develop \ucbfair, an online RL method, and prove sublinear bounds on regret (suboptimiality of cumulative loss) and distortion
(suboptimality of cumulative disparity) with high probability (\cref{sec:ucbfair}). To demonstrate practical solutions without limiting assumptions, we apply \drl, an adaptation of a well-known deep reinforcement learning method (\viz, TD3) to a Lagrangian relaxation of the central problem with time-dependent regularization. We compare \ucbfair and \drl to a baseline, myopic policy in interaction with simulated
populations initialized with synthetic or real-world data and updated according to evolutionary game theory (\cref{sec:numerical}).

This paper considers fairness in terms of statistical regularities
across (ideally) socioculturally meaningful \emph{groups}. Acknowledging that
internal conflict exists between different statistical measures of fairness, we show that an RL approach to
long-term fairness can mitigate trade-offs between fairness defined on the
statistics of immediate policy decision \emph{outcomes} \citep{chen2022fairness},
(\eg, acceptance rate disparities \citep{dwork2012fairness, zemel2013learning,
feldman2015certifying}) and underlying distributional parameters (\eg,
qualification rate \citep{raab2021unintended,zhang2020fair}). 

\subsection{Related Work} \label{sec:related}
 Our effort to formalize long-term fairness as a reinforcement learning problem bridges recent work on
``fairness in machine learning'', which has developed in response to the
proliferation of data-driven methods in society, and ``safe reinforcement
learning'', which seeks theoretical safety guarantees in the control of
dynamical systems.

\textbf{Dynamics of Fairness in Machine Learning}~ We distinguish long-term fairness from the dynamics of fair allocation problems \cite{joseph2016fairness, jabbari2017fairness, tang2021bandit, liu2017calibrated} and emphasize side-effects of algorithmic decisions affecting future decision problems. By formalizing long-term fairness in terms of cumulative losses and disparities, we iterate on a
developing research trend that accounts for the dynamical response of a human
population to deployed algorithmic prediction: both as a singular reaction
\cite{liu2018delayed, hu2019disparate, perdomo2020performative} or as a sequence
of mutual updates to the population and the algorithm \cite{coate1993will,
d2020fairness, zhang2020fair, heidari2019on, wen2019fairness, liu2019disparate,
hu2018short, mouzannar2019fair, williams2019dynamic, raab2021unintended}.
In particular, \citet{perdomo2020performative} introduces the concept of ``performative prediction'', analyzing the fixed-points of interactions between a population and an algorithmic classifier, but with state treated as a pure function of a classifier's actions. For more realistic dynamics, \citet{mouzannar2019fair} and \citet{raab2021unintended} model updates to qualification rates that depend on both previous state and the classifier's actions, but only treat myopic classifiers that optimize immediate utility (subject to fairness constraints) rather than learning to anticipate dynamical population responses. \citet{wen2019fairness} adopted reinforcement learning to achieve long-term fairness. However, they explore a setting that is restricted to discrete state and action spaces. In particular, this recommends a tabular explore-then-commit approach to reinforcement learning that does not generalize to continuous spaces and thus cannot be directly used in our setting. In addition, we provide separate and tighter bounds for both utility and disparity.

\textbf{Safe Reinforcement Learning}~
\ucbfair furthers recent efforts in safe RL. While ``model-based'' approaches, in which the algorithm
learns an explicit dynamical model of the environment, constitute one thread of
prior work \cite{efroni2020exploration,
singh2020learning, brantley2020constrained, zheng2020constrained,
kalagarla2021sample, liu2021learning, ding2021provably}, such algorithms are
typified by significant time and space complexity. Among ``model-free''
algorithms, the unknown dynamics of our setting preclude the use of a simulator
that can generate
arbitrary state-action pairs \citet{xu2021crpo, ding2020natural, bai2022achieving}.  While \citet{wei2022triple} introduce a
model-free and simulator-free algorithm, the tabular setting considered is only applicable to discrete state and action spaces.  To tackle continuous state space,
\citet{ding2021provably, ghosh2022provably} consider linear dynamics:
\citet{ding2021provably} develop a primal-dual algorithm with safe exploration,
and \citet{ghosh2022provably} use a softmax policy design. Both algorithms are
based on the work of \citet{jin2020provably}, which proposed a least squares
value iteration method, using an Upper Confidence Bound (UCB) \cite{auer2002finite} to
estimate a state-action ``\(Q\)'' function. To our knowledge,
\ucbfair is the first model-free, simulator-free RL
algorithm that provides theoretical safety guarantees for both discrete and
\textbf{continuous} state and action spaces. Moreover, \ucbfair achieves bounds
on regret and distortion as tight
as any algorithm thus far with discrete action space \cite{ghosh2022provably}.
\vspace{-0.05in}

\section{Problem Formulation} \label{sec:formulation}

A binary classification task is an initial motivating
example for our formulation of long-term fairness, though the formal problem we propose is more widely applicable. To this initial task, we introduce ``fairness'' constraints, then population
dynamics, and then cumulative loss and ``disparity'', before formalizing the
problem of optimizing cumulative loss subject to constraints on cumulative
disparity.

We provide notation for the binary classification task: a random
individual, sampled i.i.d. from a population, has features \(X \in \RR^{d}\), a
label \(Y \in \{-1, 1\}\), and a demographic group \(G \in \mathcal{G}\) (where
\(\mathcal{G} = [n]\) for \(n \geq 2\)). Denote the joint distribution of
these variables in the population as \(\state \define \Pr(X, Y, G)\). The task
is to predict \(Y\) (as \(\hat{Y}\)) from \(X\) and \(G\). Specifically, the task is to
choose a classifier \(\action\), such that \(\hat{Y} \sim \action(X, G)\), that
minimizes some bounded loss \(\mathscr{L} \in [0, 1]\) over \(\state\). This \textbf{basic classification task} is
\[
  \min_{a} \quad \mathscr{L}(\state, \action)
\]

Typically, \(\mathscr{L}\) corresponds to the expectation value of a loss
function \(L\) (such as zero-one-loss) defined for individuals drawn from \(s\).
We consider this example, but, in general, allow arbitrary, (unit-interval)
bounded loss functions \(\mathscr{L}\):
\vspace{-0.05in}
\[
  \mathscr{L}(\state, \action) \stackrel{\eg}{=} \E_{\substack{X, Y, G \sim \state \\ \hat{Y} \sim \action(X, G)}} \big[L(Y, \hat{Y})\big]
\]
The standard \textbf{``fair'' classification task}
(\emph{without} a dynamically responsive population) is to constrain
classifier \(a\) such that the \emph{disparity}
\(\mathscr{D} \in [0, 1]\) induced on distribution \(\state\) by \(a\) is
bounded by some value \(c \in [0, 1]\):
\[
  \begin{aligned}
    \min_{\action} \quad\mathscr{L}(\state, \action) \quad
    \subto \quad \mathscr{D}(\state, \action) \leq c
  \end{aligned}
\]
A standard example of disparity is the expected divergence of group acceptance
rates \(\beta\). \eg, when \( \mathcal{G} = \{g_{1}, g_{2}\}\),
\[
  \begin{aligned}
    \mathscr{D}(\state, \action) &\stackrel{\eg}{=} \big| \beta_{\state, \action}(g_{1}) - \beta_{\state, \action}(g_{2}) \big| ^{2} \\
    \text{where } \beta_{\state, \action}(g) &\define \Pr_{\substack{X, Y, G \sim \state \\ \hat{Y} \sim \action(X, G)}} (\hat{Y} {=} 1 \mid G{=}g)
  \end{aligned}
\]
In this example, \(\mathscr{D}\) measures the violation of ``demographic
parity'': the condition under which a classifier achieves group-independent
positive classification rates
\(\forall g, \Pr(\hat{Y}{=}1 \mid G{=}g) = \Pr(\hat{Y}{=}1)\)
\cite{dwork2012fairness}.
In this paper, we also consider measures of fairness based on inherent population statistics (\eg, parity of group qualification rates \(\Pr(Y{=}1\mid G{=}g)\)), which must be driven dynamically \cite{raab2021unintended,zhang2020fair}. Such notions of disparity are well-suited to an RL formulation of long-term fairness.
\vspace{-0.1in}
\paragraph{State, Action, and Policy}
Considering the semantics of iterated classification tasks, we identify the distribution
\(\state \in \mathcal{S}\) of individuals in the population as a \emph{state}
and the classifier \(a \in \mathcal{A}\) as an \emph{action}. While state
space \(\mathcal{S}\) is arbitrary, we assume that action space
\(\mathcal{A}\) admits a Euclidean metric, under which it is closed (\ie,
\(\mathcal{A}\) is isomorphic to \([0, 1]^{m}, m \in \ZZ_{>0}\)). At a given
time \(\tau\), \(a_{\tau}\) is sampled stochastically according to the current
\emph{policy} \(\pi_{\tau}\):
$
  \action_{\tau} \sim \pi_{\tau}(\state_{\tau}).
$
We assume \(\state_{\tau}\) is fully
observable at time \(\tau\). In practice, \(\state_{\tau}\) must be approximated
from finitely many empirical samples, though this caveat introduces
well-understood errors that vanish in the limit of infinitely many samples.

\vspace{-0.1in}
\paragraph{Dynamics}
In contrast to a ``one-shot'' fair classification task, we assume
that a population may react to classification, inducing the distribution \(s\)
to change.  Importantly, such ``distribution shift'' is a well-known, real-world
phenomenon that can increase realized loss and disparity when deployed
classification policies are fixed \cite{chen2022fairness}. For classification
policies that free to change in response to a mutating distribution \(s\),
subsequent classification tasks depend on the (stochastic) predictions made in
previous tasks.  In our formulation, we assume the existence of dynamical kernel
\(\PP\) that maps a state \(\state\) and action \(\action\) at time \(\tau\) to
a \emph{distribution over} possible states at time \(\tau + 1\):
\vspace{-0.1in}
\begin{equation}
  \state_{\tau + 1} \sim \PP(\state_{\tau}, \action_{\tau})
\end{equation}
We stipulate that \(\PP\) may be
initially unknown, but it does not explicitly depend on time and may be
reasonably approximated ``online''. While real-world dynamics may depend on information other than the current distribution
\(\Pr(X, Y, G)\) (\eg, exogenous parameters,
history, or additional variables of state), we identify \(s\) with the current distribution for simplicity in
our treatment of long-term fairness.

\vspace{-0.1in}
\paragraph{Reward and Utility}
Because standard RL literature motivates \emph{maximizing
  reward} rather than \emph{minimizing loss}, let us define
the instantaneous reward
\(r \in [0, 1]\) and a separate, instantaneous ``utility'' \(g \in [0, 1]\) for
an RL agent as
\begin{align}
  r(\state_{\tau}, \action_{\tau}) &\define 1 - \mathscr{L}(\state_{\tau}, \action_{\tau}) \\
  g(\state_{\tau}, \action_{\tau}) &\define 1 - \mathscr{D}(\state_{\tau}, \action_{\tau})
\end{align}
where \(r\) and \(g\) do not explicitly depend on time \(\tau\).
\paragraph{Value and Quality Functions}
Learnable dynamics inspire us to 
optimize anticipated \emph{cumulative} reward, given constraints on
anticipated \emph{cumulative} utility. Let \(j\) represent either reward \(r\) or utility \(g\). We use the letter \(V\) (for ``value'') to denote the future expected accumulation of \(j\) over steps \([h, ..., H]\) (without time-discounting) starting from state \(\state\), using policy \(\pi\). Likewise, we denote the ``quality'' of an action
\(\action\) in state \(\state\) with the letter \(Q\). For \(j \in \{r, g\}\),
\begin{align}
    V_{j, h}^\pi(s) &= \E \Big[\sum_{\tau=h}^H j\big(s_\tau, a_\tau\big)| s_h = s\Big] \\
    Q_{j, h}^\pi(s, a) &= \E\left[\sum_{\tau=h}^H j\big(s_\tau, a_\tau)\big) \mid s_h=s, a_h=a\right]
\end{align}
By the boundedness of \(r, g \in [0, 1]\), \(V\) and \(Q\) belong to the interval \([0, H - h + 1]\).

\vspace{-0.05in}
\paragraph{``Long-term fairness'' via reinforcement learning}
The central problem explored in this paper is
\begin{align}\label{eq:optim}
    \max_\pi \quad V_{r, 1}^\pi(s) \quad
    \subto  \quad V_{g, 1}^\pi(s) \geq \tilde{c}
\end{align}
We emphasize that this construction of long-term fairness considers a finite time horizon of \(H\) steps 
and denote the optimal value of \(\pi\) as \(\pi^{\star}\).
\begin{restatable}[Slater's Condition]{assumption}{slaters-condition} \label{asm:slaters-condition}
\ie, ``strict primal feasibility'': $\exists$ $\gamma>0$,
$\bar{\pi}$, such that $V_{g, 1}^{\bar{\pi}}\left(s\right) \geq$
$\tilde{c}+\gamma$.
\end{restatable}
\vspace{-0.05in}
Slater's condition is also adopted in \cite{efroni2020exploration, ding2021provably, ghosh2022provably}.

\vspace{-0.05in}
\paragraph{The Online Setting} Given initially unknown dynamics, we
formulate long-term fairness for the ``online'' setting, in which
learning is only possible through actual ``live'' deployment of policy, rather
than through simulation. As it is not possible to
unconditionally guarantee constraint satisfaction in \cref{eq:optim} over a finite number of
episodes, we instead measure two types of \emph{regret}: one that measures
the suboptimality of a policy with respect to cumulative incurred loss, which we
will continue to call ``regret'', and one that measures the suboptimality of a
policy with respect to cumulative induced disparity, which we will call
``distortion''. Note that we define regret and distortion in \cref{eq:regret-simple} by marginalizing over the stochasticity
of state transitions and the sampling of actions:
\begin{equation} \label{eq:regret-simple}
\begin{aligned}
    \operatorname{Regret}(\pi, s_{1}) \define V_{r, 1}^{\pi^*}\left(s_1\right)-V_{r, 1}^{\pi}\left(s_1\right)\\
    \operatorname{Distortion}(\pi, s_{1}) \define \max\left[ 0, \tilde{c}-V_{g, 1}^{\pi}\left(s_1\right)\right]
\end{aligned}
\end{equation}

\section{Algorithms and Analysis}

We show that it is possible to provide guarantees for long-term fairness in the
online setting. Specifically, we develop an algorithm, \ucbfair, and prove that
it provides probabilistic, sublinear bounds for regret and distortion under a
linear MDP assumption (\cref{asm:linear-mdp}) and properly chosen parameters
(\cref{proof:thmbound}, in the supplementary material).  \ucbfair is
the first model-free algorithm to provide such guarantees in the online setting
with continuous state and action spaces.
\subsection{\ucbfair} \label{sec:ucbfair}

\paragraph{Episodic MDP}
  \ucbfair inherits from a family of algorithms that treat an episodic Markov
decision process (MDP) \cite{jin2020provably}. Therefore, we first map the
long-term fairness problem to the episodic MDP setting, which we denote  as
$\text{MDP}(\mathcal{S}, \mathcal{A}, H, \PP, \mathscr{L}, \mathscr{D})$. The algorithm runs for \(K\) \emph{episodes}, each consisting of
\(H\) time steps. At the
beginning of each episode, which we index with \(k\), the agent commits to a
sequence of policies \(\pi^{k} = (\pi_{1}^{k}, \pi_{2}^{k}, ..., \pi_{H}^{k})\)
for the next \(H\) steps. At each step \(h\) within an episode, an action
\(\action_{h}^{k} \in \mathcal{A}\) is sampled according to policy
\(\pi_{h}^{k}\), then the state \(s_{h+1}^{k}\) is sampled according to the
transition kernel \(\PP(\state_{h}^{k}, \action_{h}^{k})\). \(\mathscr{L}\) and
\(\mathscr{D}\) are the loss and disparity functions. 
\(s_1^k\) is sampled arbitrarily with each episode.

\vspace{-0.1in}
\paragraph{Episodic Regret}
Because \ucbfair predetermines its policy for an entire episode, we amend our
definition of regret and distortion for the all \(HK\) time steps by breaking it into a sum over \(K\) episodes of length \(H\).
\begin{equation}\label{eq:regret}
\begin{aligned}
    \operatorname{Regret}(K) = \sum_{k=1}^K\left(V_{r, 1}^{\pi^*}\left(s_1^k\right)-V_{r, 1}^{\pi^k}\left(s_1^k\right)\right)\\
    \operatorname{Distortion}(K) = \max\left[0, \sum_{k=1}^K\left(\tilde{c}-V_{g, 1}^{\pi^k}\left(s_1^k\right)\right)\right]
\end{aligned}
\end{equation}

\paragraph{The Lagrangian}
Consider the Lagrangian function
\(\mathcal{L}\) associated with \cref{eq:optim}, with dual variable \(\nu \geq 0\):
\begin{equation}\label{eq:lagrangian}
    \mathcal{L}(\pi, \nu):=V_{r, 1}^\pi\left(s\right)+\nu\left(V_{g, 1}^\pi\left(s\right)-\tilde{c}\right)
\end{equation}
\ucbfair approximately solves the primal problem \(\max_{\pi} \min_{\nu} \mathcal{L}(\pi, \nu)\), which is
non-trivial, since the objective function is seldom concave in practical
parameterizations of \(\pi\). Let \(\nu^{*}\) to denote the optimal
value of $\nu$.
\begin{restatable}[Boundedness of $\nu^{*}$]{lemma}{boundedy} \label{asm:boundedy} For \(\Bar{\pi}\) and \(\gamma > 0\) satisfying Slater's Condition
(\cref{asm:slaters-condition}),
    \begin{equation*}
        \nu^{*} \leq \frac{V_{r, 1}^{\pi *}\left(s_1\right)-V_{r, 1}^\pi\left(s_1\right)}{\gamma} \leq \frac{H}{\gamma} \define \mathscr{V}
    \end{equation*}
  \end{restatable}
  \cref{asm:boundedy}, defines \(H/\gamma = \mathscr{V}\) as an upper bound for the optimal dual variable \(\nu^{*}\). \(\mathscr{V}\) is an input to \ucbfair.

\begin{restatable}[Linear MDP]{assumption}{linearmdp} \label{asm:linear-mdp}
    $\text{MDP}(\mathcal{S}, \mathcal{A}, H, \PP, \mathscr{L}, \mathscr{D})$ is
a linear MDP with feature map
$\phi: \mathcal{S} \times \mathcal{A} \rightarrow \mathbb{R}^d$: For any $h$,
there exist $d$ signed measures
$\mu_h=\left\{\mu_h^1, \ldots, \mu_h^d\right\}$ over $\mathcal{S}$, such that, for
any $\left(s, a, s^{\prime}\right) \in \mathcal{S} \times \mathcal{A} \times \mathcal{S}$,
$$
\mathbb{P}_h\left(s^{\prime} \mid s, a\right)=\left\langle\phi(s, a), \mu_h\left(s^{\prime}\right)\right\rangle,
$$
In addition, there exist vectors $\theta_{r, h}, \theta_{g, h} \in \mathbb{R}^d$, such
that, for any $(s, a) \in \mathcal{S} \times \mathcal{A}$,
$$
 r\big(s, a\big)=\left\langle\phi(s, a), \theta_{r, h}\right\rangle; \quad g(s, a)=\left\langle\phi(s, a), \theta_{g, h}\right\rangle
$$
\end{restatable}

\cref{asm:linear-mdp} addresses the curse of dimensionality when state space \(\mathcal{S}\)
is the space of distributions over \(X, Y, G\). This assumption is also used in
\cite{jin2020provably, ghosh2022provably}, with a similar assumption made in \cite{ding2021provably}. 

\subsubsection{Explicit Construction}
\ucbfair, or ``\texttt{LSVI-UCB} for Fairness
'' (\cref{alg:UCBFair}) is based on an
optimistic modification of a Least-Squares Value Iteration, where optimism is
realized by an Upper-Confidence Bound, as in LSVI-UCB \cite{jin2020provably}.
For each \(H\)-step episode \(k\), \ucbfair maintains estimates for \(Q^{k}_{r}, Q^{k}_{g}\)
and a time-indexed policy $\pi^{k}$. In each episode \(k\), \ucbfair updates \(Q^{k}_{r}, Q^{k}_{g}\),
interacts with the environment, and updates the dual variable $\nu_{k}$ (constant in \(k\)).

\vspace{-0.1in}
\paragraph{LSVI-UCB \cite{jin2020provably}} The estimation of \(Q\) is
challenging. Consider the Bellman equation:
\(\forall(s, a) \in \mathcal{S} \times \mathcal{A}, \)
 \begin{equation*}
     Q_h^{\star}(s, a) \leftarrow\left[r_h+\E_{s^{\prime} \sim \PP_h(\cdot \mid s, a)} \max _{a^{\prime} \in \mathcal{A}} Q_{h+1}^{\star}\left(\cdot, a^{\prime}\right)\right](s, a)
 \end{equation*}
It is impossible to iterate over all \(s, a\) pairs, since \(\mathcal{S}\) and
\(\mathcal{A}\) are continuous. Instead, LSVI parameterizes \(Q_h^{\star}(s, a)\)
by the linear form  \(\mathrm{w}_h^{\top} \phi(s, a)\), which is supported by \citet{jin2020provably}. As an additional complication, \(\PP\) is unknown is estimated by LSVI
from empirical samples. Combining these
techniques, the Bellman equation can be replaced by a least squares problem.
 \begin{align*}
     \mathbf{w}_h \leftarrow \underset{\mathbf{w} \in \mathbb{R}^d}{\operatorname{argmin}} \sum_{\tau=1}^{k-1} & \bigg[r_h\left(s_h^\tau, a_h^\tau\right)+\max _{a \in \mathcal{A}} Q_{h+1}\left(s_{h+1}^\tau, a\right)
     -\mathbf{w}^{\top} \phi\left(s_h^\tau, a_h^\tau\right)\bigg]^2+\varsigma\|\mathbf{w}\|^2
 \end{align*}
 In addition, a ``bonus term''
\(\beta\left(\phi^{\top} \Lambda_h^{-1} \phi\right)^{1 / 2}\) is added to the estimate of Q to encourage exploration.
 
\textbf{Adaptive Search Policy} Compared to the works of
\citet{ding2021provably} and \citet{ghosh2022provably}, the major challenge we
face for long-term fairness is a continuous action space \(\mathcal{A}\), which
renders the independent computation of \(Q^{k}_{r}, Q^{k}_{g}\) for each action
impossible. To handle this issue, we propose an adaptive search policy: Instead
of drawing an action directly from a distribution over continuous values,
\(\pi\) represents a distribution over finitely many, finite regions of
\(\mathcal{A}\).  After sampling a region from a Softmax scheme, the agent draws action \(\action\) uniformly at random from it.
\begin{definition}\label{def}
    Given a set of distinct actions
$I = \{I_0, \cdots, I_M\} \subset \mathcal{A}$, where $\mathcal{A}$ is a closed
set in Euclidean space, define
$\mathcal{I}_i = \{a \colon \|a - I_i\|_2 \leq \|a - I_j\|_2, \forall i < j\}$
as the subset of actions closer to \(I_i\) than to \(I_{j}\), \ie, the Voronoi
region corresponding to locus \(I_{i}\), with tie-breaking imposed by the order of indices \(i\). Also define the locus function
$I(a) = \min_i \argmin_{I_i} \|a - I_i\|_2$.
\end{definition}
\begin{lemma} The Voronoi partitioning described above satisfies
$\mathcal{I}_i \cap \mathcal{I}_j = \varnothing, \forall i\neq j$ and
$\cup_{i=1}^{M} \mathcal{I}_i = \mathcal{A}$.
\end{lemma}
\begin{theorem}
If the number \(M\) of distinct loci or regions partitioning \(\mathcal{A}\) is
sufficiently large, there exists a set of loci \(I\) such that
$\forall a \in \mathcal{I}_i, i\in M, \|a - I_i\|_2 \leq \epsilon_I$.
\end{theorem}
In addition to defining a Voronoi partitioning of the action space for the
adaptive search policy of \ucbfair, we make the following assumption:
\begin{assumption} \label{asm:lipschitz} There exists \(\rho > 0\), such that
       \(\|\phi(s, a) - \phi(s, a^{\prime})\|_2 \leq \rho \| a - a^{\prime}\|_2.\)
\end{assumption}
This assumption is used to bound the distance of estimated action-value function and value function, thus bound adaptive search policy and the optimal policy. 

\textbf{Dual Update} For \ucbfair, the update method for the dual variable $\nu$
in \cref{eq:lagrangian} is also essential. Since \(V_{r, 1}^\pi\left(s\right)\)
and \(V_{g, 1}^\pi\left(s\right)\) are unknown, we use
\(V_{r, 1}^k\left(s\right)\) and \(V_{g, 1}^k\left(s\right)\) to estimate them. $\nu$ is iteratively updated by
minimizing \cref{eq:lagrangian} with step-size $\eta$, and $\mathscr{V}$ is an upper
bound for $\nu$. A similar method is also used in \citet{ding2021provably,
ghosh2022provably}.

\subsubsection{Analysis}
We next bound the regret and distortion, defined in \cref{eq:regret}, realizable
by \ucbfair. We then compare \ucbfair with existing algorithms for discrete
action spaces and discuss the importance of the number of regions $M$ and the
distance $\epsilon_I$.
\begin{restatable}[Boundedness]{theorem}{bound} \label{thm:bound}
 With probability $1-p$, there exists a constant \(b\) such that \ucbfair
achieves
\begin{align}
\text{Regret}(K) &\leq \big( b \zeta H^2 \sqrt{d^{3}}+ (\mathscr{V} + 1) H \big) \sqrt{K} \\
\text { Distortion }(K) &\leq \frac{b \zeta (1+\mathscr{V}) H^2 \sqrt{d^{3}}}{\mathscr{V}}  \sqrt{K}
\end{align}
when selecting parameter values
$\varsigma{=}1$ $\epsilon_I \leq \frac{1}{2\rho (1 + \mathscr{V}) KH \sqrt{d}}$,
$\zeta=\log (\log (M) 4 d H K / p)$, and $\beta= \mathcal{O}( d H \sqrt{\zeta})$.
\end{restatable}

For a detailed proof of \cref{thm:bound}, refer to \cref{proof:thmbound}.
\cref{thm:bound} indicates that \ucbfair reaches
$\tilde{\mathcal{O}}\left(H^2 \sqrt{d^3 K}\right)$ bounds for both regret and
distortion with high probability. Compared to the algorithms introduced by
\citet{ding2021provably, ghosh2022provably}, which work with discrete action
space, \ucbfair guarantees the same asymptotic bounds on regret and distortion.

\begin{algorithm}[htb]
   \caption{\ucbfair}
   \label{alg:UCBFair}
\begin{algorithmic}
  \STATE {\bfseries Input:} A set of points $\{I_0, I_1, \cdots, \I_M\}$ satisty \cref{def}. $\epsilon_I = \frac{1}{2\rho (1 + \chi) KH}$.\\
  $\nu_{1}{=}0$.
  $w_{r, h}{=}w_{g,h}{=}0$.
  $\alpha{=}\frac{\log (M) K}{2(1+\chi+H)}$.
  $\eta{=}\chi / \sqrt{K H^{2}}$.
  $\beta{=}C_{1} d H \sqrt{\log (4 \log M d T / p)}$, $\varsigma=1$.
\FOR{episode $k=1,2,...,K$}
    \STATE Receive the initial state $s^k_1$.
    \FOR{step $h = H, H - 1, \cdots, 1$}
        \STATE $\Lambda_{h}^{k} \leftarrow \sum_{\tau=1}^{k-1} \phi\left(s_{h}^{\tau}, a_{h}^{\tau}\right) \phi\left(s_{h}^{\tau}, a_{h}^{\tau}\right)^{T}+\varsigma \mathbf{I}$
        \FOR{$j\in\{r, g\}$}
        \STATE $w_{j, h}^{k} \leftarrow \left(\Lambda_{h}^{k}\right)^{-1}\left[\sum_{\tau=1}^{k-1} \phi\left(s_{h}^{\tau}, a_{h}^{\tau}\right)\big(j\left(s_{h}^{\tau}, a_{h}^{\tau}\right)+V_{j, h+1}^{k}\left(s_{h+1}^{\tau}\right)\big)\right]$
        \ENDFOR
        \FOR{iteration $i = 1, \cdots, M$ and index $j \in \{r, g\}$}
            \STATE $\xi_{i,j} \leftarrow \left(\phi(\cdot, I_i)^{T}\left(\Lambda_{h}^{k}\right)^{-1} \phi(\cdot, I_i)\right)^{1 / 2}, \quad Q_{j, h}^{k}(\cdot, I_i){\leftarrow}\min \left[\Big\langle w_{j, h}^{k}, \phi(\cdot, I_i)\Big\rangle+\beta \xi_{i,j}, H\right]$
        \ENDFOR
        \STATE $\text{SM}_{h, k}(I_i \mid \cdot)=\frac{\exp \left(\alpha\left(Q_{r, h}^{k}(\cdot, I_i)+\nu_{k} Q_{g, h}^{k}(\cdot, I_i)\right)\right)}{\sum_{j} \exp \left(\alpha\left(Q_{r, h}^{k}(\cdot, I_j)+\nu_{k} Q_{a, h}^{k}(\cdot, I_j)\right)\right)}$
        \STATE $\pi_{h}^k(a \mid \cdot)\leftarrow \frac{1}{\int_{b\in\mathcal{I}(a)} db} \text{SM}_{h, k}(I(a) \mid \cdot)$
        \STATE $V_{r, h}^{k}(\cdot)\leftarrow \int_{a\in\mathcal{A}} \pi_{h}^k(a \mid \cdot) Q_{r, h}^{k}(\cdot, a) da, \quad V_{g, h}^{k}(\cdot)\leftarrow\int_{a\in\mathcal{A}} \pi_{h}^k(a \mid \cdot) Q_{g, h}^{k}(\cdot, a) da$
    \ENDFOR
    \FOR{step $h = 1, \cdots, H$}
        \STATE Compute $Q_{r, h}^{k}\left(s_{h}^{k}, I_i\right), Q_{g, h}^{k}\left(s_{h}^{k}, I_i\right), \pi\left(I_i \mid s_{h}^{k}\right)$.
        \STATE Take action $a_{h}^{k} \sim \pi_{h}^k\left(\cdot \mid s_{h}^{k}\right)$ and observe $s_{h+1}^{k}$.
    \ENDFOR
    \STATE $\nu_{k+1}=\max \left\{\min \left\{\nu_{k}+\eta\left(\Tilde{c}-V_{g, 1}^{k}\left(s_{1}\right)\right), \mathscr{V}\right\}, 0\right\}$
\ENDFOR
\end{algorithmic}
\end{algorithm}

\subsection{\drl} \label{sec:drl}

Because assumptions made in \cref{sec:ucbfair} (\eg, the linear MDP assumption)
are often violated in the real world, we also consider more flexible and arguably powerful deep
reinforcement learning methods. Concretely, we utilize ``Twin-Delayed
Deep Deterministic Policy Gradient'' (TD3) \cite{fujimoto2018addressing} with
the implementation and default parameters provided by the open-source package
``Stable Baselines 3'' \cite{stable-baselines3} on 
a (Lagrangian) relaxation of the long-term fairness problem. We term this specific algorithm for long-term fairness \drl.

In general, methods such as TD3 lack provable safety guarantees, however they
may still confirm our hypothesis that agents trained via RL
can learn to sacrifice short-term utility in order to drive dynamics towards
preferable long-term states and expliictly incorporate dynamical control objectives provided as functions of state.

To treat the long-term fairness problem (\cref{eq:optim}) using
unconstrained optimization techniques (\ie, methods like TD3), we consider a time-dependent 
Lagrangian relaxation of \cref{eq:optim}. We
train \drl to optimize
\begin{equation}\label{eq:drloptim}
    \min_{\pi} \E_{\action_{\tau} \sim \pi(\state_{\tau})} \left[\sum_{\tau=1}^{H} \left[ \kappa_\tau \mathscr{L}(\state_{\tau}, \action_{\tau}) + \lambda_{\tau} \mathscr{D}(\state_{\tau}, \action_{\tau}) \right]\right]
\end{equation}
where \(\state_{\tau+1} \sim \PP(\state_{\tau}, \action_{\tau})\), \(\lambda_{\tau} = \tau / H\), and \(\kappa_\tau = 1 - \lambda_\tau\).

Strictly applied,
myopic fairness constraints can lead to undesirable dynamics and equilibria \cite{raab2021unintended}. Relaxing these constraints (hard \(\to\) soft) for the near future while emphasizing them long-term, we hope to develop classifiers that learn to transition to more favorable equilibria.

\subsection{Greedy Baseline} \label{sec:baseline}

In our experiments, we compare \ucbfair and \drl to a ``Greedy Baseline'' agent
as a proxy for a myopic status quo in which policy is repeatedly determined
by optimizing for immediate utility, without regard for the population dynamics
induced by algorithmic actions. Our chosen algorithm for the greedy baseline is
simply gradient descent in \(f\), defined as loss regularized by disparity, performed anew with each time step with fixed parameter \(\lambda\).
\begin{equation}\label{eq:greedy}
  f_{\tau}(\pi) = \E_{a_{\tau} \sim \pi}\left[(1 - \lambda) \mathscr{L}(\state_{\tau}, \action_{\tau}) + \lambda \mathscr{D}(\state_{\tau}, \action_{\tau}) \right]
\end{equation}
While such an algorithm does not guarantee constraint satisfaction, it is
nonetheless ``constraint aware'' in precisely the same way as a firm that
(probabilistically) incurs penalties for violating constraints.
\vspace{-0.1in}
\section{Simulated Environments} \label{sec:numerical}
We describe
our experiments with the algorithms we have detailed for long-term fairness as an RL problem: We consider a series of binary (\(Y \in \{-1, 1\}\) classification tasks on a population of two groups \(\mathcal{G} = \{g_{1}, g_{2}\}\) modeled according to evolutionary game theory (using replicator dynamics). We consider two families of distributions of real-valued features for the population: One that is purely synthetic, for which
\(X \sim \mathcal{N}(Y, 1)\), independent of group \(G\), and one that is based
on a logistic regression to real-world data. Both families of distributions are
parameterized by the joint distribution \(\Pr(Y, G)\).
RL agents are trained on episodes of length \(H\) initialized with randomly sampled states.
\subsection{Setting}
The following assumptions simplify our hypothesis space for classifiers in order to better handle continuous state space. 
These assumptions appeared in \citet{raab2021unintended}.
\begin{restatable}[Well-behaved feature]{assumption}{well-behaved} \label{asm:well-behaved}
For purely synthetic data, we require \(X\) to be a ``well-behaved'' real-valued
feature or ``score'' within each group. That is,
  \begin{equation*}
    \forall g, \quad \Pr(Y{=}1 \mid G{=}g, X{=}x) \text{ strictly increases in } x
  \end{equation*}
\end{restatable}
As an intuitive example of \cref{asm:well-behaved}, if \(Y\) represents qualification for a fixed loan and \(X\) represents credit-score, we require higher credit scores to strongly imply higher likelihood that an individual is qualified for the loan.

\begin{restatable}[Threshold Bayes-optimality]{theorem}{threshold} \label{thm:threshold}
  For each group \(g\), when \cref{asm:well-behaved} is satisfied, the
Bayes-optimal, deterministic binary classifier is a threshold policy
\vspace{-0.05in}
  \begin{equation*}
    \hat{Y} = 1 \text{ if } x \geq A_g \text{ and } -1 \text{ otherwise}
  \end{equation*}
\vspace{-0.05in}
  where \(A_g\) is the feature threshold for group \(g\).
\end{restatable}
As a result of \cref{thm:threshold}, we consider our action 
space to be the space of group-specific thresholds, and denote
an individual action as the vector
$
  \mathbf{A} \define (A_{1}, A_{2}, ..., A_{n})$.
  
\subsection{Replicator Dynamics} \label{sec:replicator}
Our use of replicator dynamics closely mirrors that of \cite{raab2021unintended}
as an ``equitable'' model of a population, in which individuals my be modeled
identically, independently of group membership, yet persistent outcome
disparities may nonetheless emerge from disparate initial conditions between
groups. In particular, we parameterize the evolving distribution
\(\Pr(X, Y \mid G)\), assuming constant group sizes, in terms of ``qualification
rates''
$
  q_{g} \define \Pr(Y{=}1 \mid G{=}g)
$
and update these qualification rates according to the discrete-time replicator
dynamics:
\[
  q_{g}[t + 1] = q_{g}[t] \frac{W_{1}^{g}[t]}{\avg{W}^{g}[t]}; \quad \avg{W}^{g}[t] \define W_{1} q_{g} + (1 - q_{g}) W_{{-}1}
\]
In this model, the \emph{fitness} \(W_{y}^{g} > 0\) of label \(Y{=}y\) in group
\(G{=}g\) may be interpreted as the ``average utility to the individual'' in
group \(g\) of possessing label \(y\), and thus relative \emph{replication rate}
of label \(y\) in group \(g\), as agents update their labels by mimicking
the successful strategies of in-group peers. Also following \citet{raab2021unintended}, we model \(W_{y}^{g}\) in terms of
acceptance and rejection rates with a group-independent utility matrix \(U\):
\[
  W_{y}^{g} = \sum_{\hat{y} \in \{{-}1, 1\}} U_{y,\hat{y}} \Pr(\hat{Y}{=}\hat{y} \mid Y{=}y, G{=}g)
\]
We choose the matrix \(U\) to eliminate dominant strategies (\ie, agents prefer
one label over another, independent of classification), assert that agents
always prefer acceptance over rejection, and to imply that the costs of
qualification are greater than the costs of non-qualification among accepted
individuals. While other parameterizations of \(U\) are valid, this choice of
parameters guarantees internal equilibrium of the replicator dynamics for a
Bayes-optimal classifier and ``well-behaved'' scalar-valued feature \(X\), such
that \(\Pr(Y{=}1 \mid X{=}x)\) is monotonically increasing in \(x\)
\cite{raab2021unintended}.
\vspace{-0.05in}
\subsection{Data Synthesis and Processing}\label{sec:synthesis}
In addition to a synthetic distribution, for which we assume
\(X \sim \mathcal{N}(Y, 1)\), independent of \(G\), for all time, we also
consider real-world distributions in simulating and comparing algorithms for
``long-term fairness''. In both cases, as mentioned above, we wish to
parameterize distributions in terms of qualification rates \(q_{g}\). As we perform binary classification on
discrete groups and scalar-valued features, in addition to parameterizing a
distribution in terms of \(q_{g}\), we desire a scalar-valued feature for each
example, rather than the multi-dimensional features common to real-world data. Our solution to parameterize a distribution of groups and scalar features is to use an additional learning step for
``preprocessing'': Given a static dataset \(\mathcal{D}\) 
from which \((X',Y,G)\) is drawn i.i.d., (\eg, the ``Adult Data Set'' \citet{Dua:2019}), at each time-step, we train a stochastic binary
classifier \(\tilde{a}\), such that \(\hat{Y}' \sim \tilde{a}(X', G)\) with a loss that re-weights examples by label value, in order to simulate the desired \(q_{g}\):
$
  \min_{\tilde{a}} \quad \E_{\tilde{a}, \mathcal{D}} [w(X', Y, G) L(Y, \hat{Y}')]
$, where $  w(X', Y, G)  =
  [ (1 - Y) / 2 + Y q_g ]/ \E_{\mathcal{D}}[Y | G]
  $, \(L\) is zero-one loss, and, in our experiments, we choose \(\tilde{a}\) according to logistic regression. We interpret \(\Pr(\hat{Y}{=}1)\) as a new, scalar feature value \(X \in \RR\) mapped from from higher-dimensional
features \(X'\) as the output of a learned ``preprocessing'' function \(\tilde{a}\),
\cref{asm:well-behaved} is as hard to satisfy in general as solving the
Bayes-optimal binary classification task over higher-dimensional features.
Nonetheless, we expect \cref{asm:well-behaved} to be approximately satisfied by
such a ``preprocessing'' pipeline.
\vspace{-0.05in}
\subsection{Linearity of Dynamics}
\ucbfair relies on \cref{asm:linear-mdp}, which asserts the existence of some Hilbert space in which the state dynamics \(\PP\) are linear. Such linearity for real-world (continuous time) dynamics holds only in infinite-dimensional Hilbert space \cite{brunton2021modern} and is not computationally tractable.
In addition, the ``feature map'' \(\phi\) that maps state-action pairs to the aforementioned Hilbert space must be learned by the policy maker. In experiment, we use a neural network to estimate a feature map $\hat{\phi}$ which approximately satisfies the linear MDP assumption. We defer details to \cref{app:exp_details}. 
\vspace{-0.05in}
\section{Experimental Results}\label{sec:results}
Do RL agents learn to seek favorable equilibria against short-term utility? Is a Lagrangian relaxation
of long-term fairness sufficient to encourage this behavior? We give positive demonstrations for both questions.
\vspace{-0.1in}
\subsection{Losses and Disparities Considered}\label{sec:losses}
Our experiments consider losses \(\mathscr{L}\) which combine
true-positive and true-negative rates, where, for \(\alpha, \beta \in [0, 1]\),
\begin{equation}\label{eq:loss_function}
  \mathscr{L}(\state, \action) = 1 - \alpha \tp(\state, \action) + \beta \tn(\state, \action),
\end{equation}
where \(\tp(\state, \action) = \Pr_{\state, \action}(\hat{Y}{=}1, Y{=}1)\) and \(\tn(\state, \action) = \Pr_{\state, \action}(\hat{Y}{=}{-}1, Y{=}{-}1)\).
For disparity \(\mathscr{D}\), we consider demographic parity (\DP) \cite{dwork2012fairness}, 
equal opportunity (\EOp) \cite{hardt2016equality}, and qualification rate (\QR): 
\begin{table}[h]
    \centering
    \resizebox{0.6\linewidth}{!}{
    \begin{tabular}{lcc}
    \hline
         Func.&  Form& \(\xi^y_{\state, \action}(g) = \Pr_{\state, \action} (\cdot)\)\\
         \hline
         \DP & \multirow{3}{9em}{\(\big| \xi_{\state, \action}(g_{1}){-}\xi_{\state, \action}(g_{2}) \big| ^{2} / 2\)} & \(\hat{Y} {=} 1 \mid G{=}g\) \\
         \QR& &\(\hat{Y} {=} 1 \mid G{=}g\)\label{eq:DP-2}\\
         \EOp& &\(\hat{Y} {=} 1 \mid G{=}g\)\\
         \hline
         \EO& \(\sum_y\big| \xi^{y}_{\state, \action}(g_{1}) - \xi^y_{\state, \action}(g_{2}) \big| ^{2} / 2\)& \(\hat{Y}{=}\hat{y} \mid Y{=}y, G{=}g\)\\
         \hline
    \end{tabular}}
    \label{eq:notion}
\end{table}
\vspace{-0.1in}

\QR does not matter to myopic fair
classification, which does not consider mutable population state.

\begin{figure}[htb]
  \centering
    \includegraphics[width=0.49\linewidth]{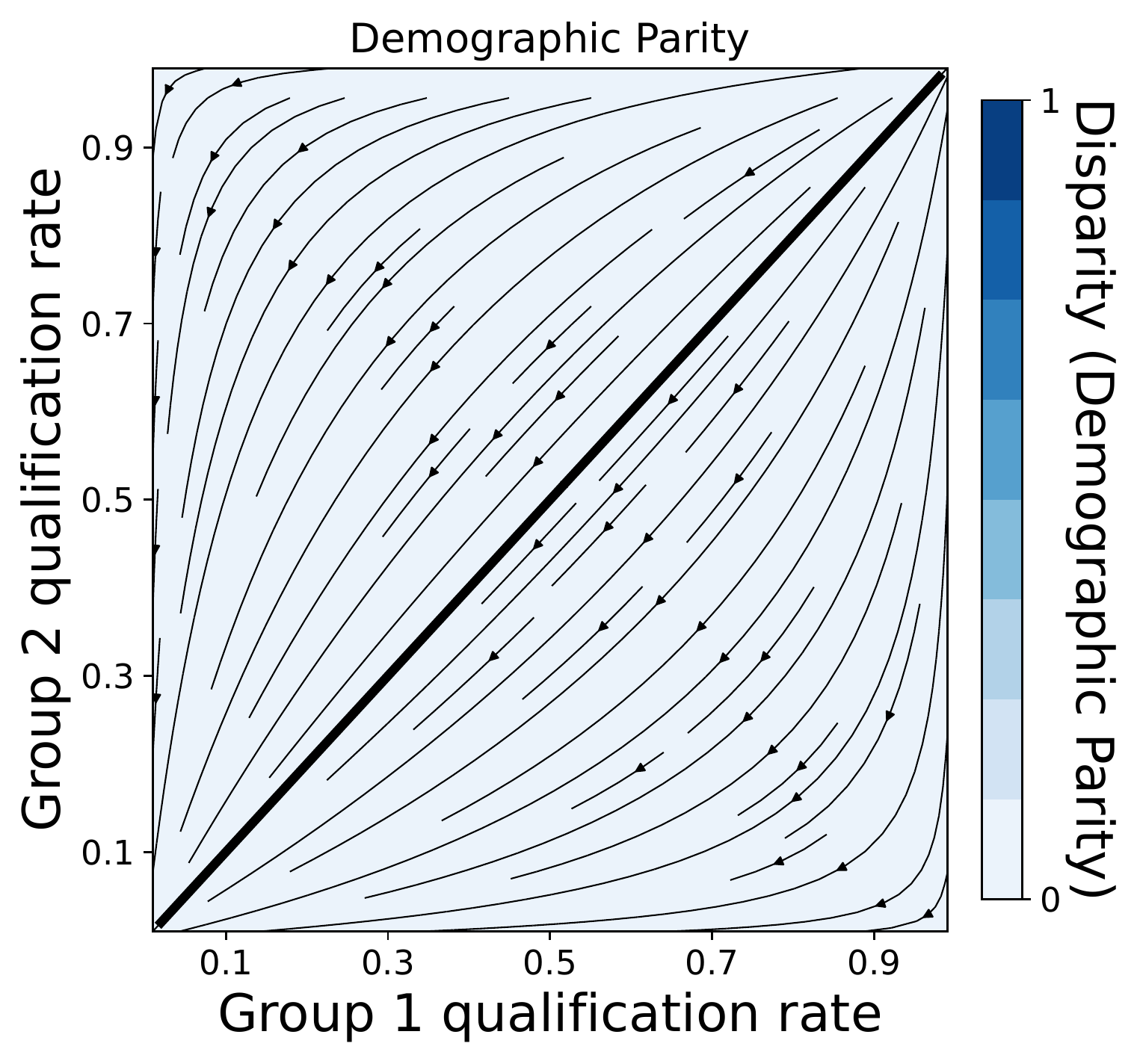}
    \includegraphics[width=0.49\linewidth]{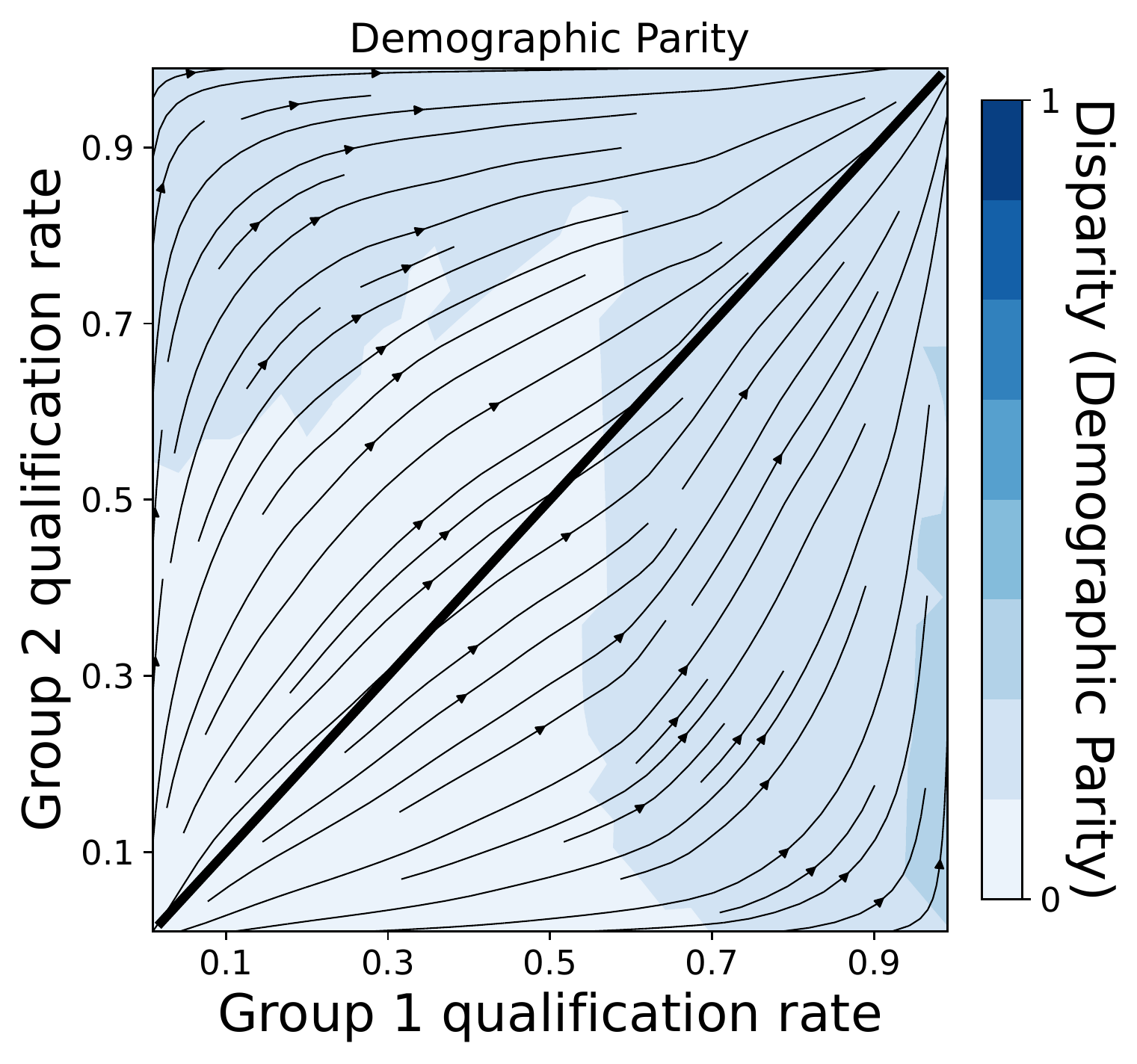}
    \includegraphics[width=0.88\linewidth]{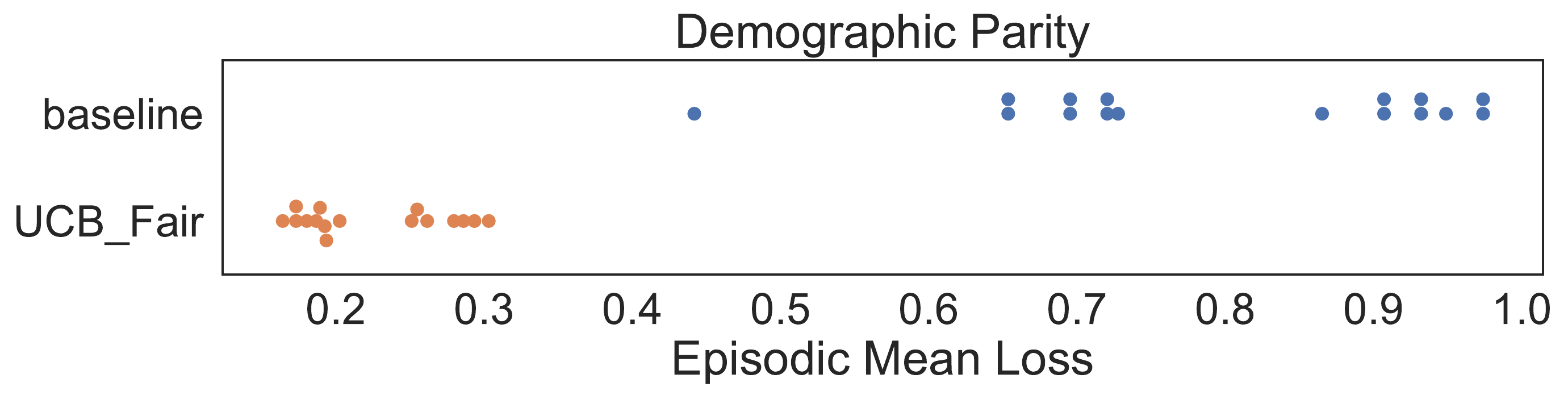}
    \caption{The greedy baseline algorithm (left) and \ucbfair (right) are
tasked to maximize the fraction of true-positive classifications
(\(\mathscr{L}=1-\tp\), \cref{eq:loss_function}), subject to demographic parity
(\(\mathscr{D}{=}\DP\), \cref{eq:DP-2}). The greedy algorithm uses
\(\lambda{=}0.5\) in \cref{eq:greedy}, while \ucbfair is trained for 2,000 steps
on episodes of length 100 prior to generating this ``phase portrait''. We depict
the expected dynamics (averaged over 20 policy iterations for each state) of the
classifier-population system, parameterized by the time-evolving qualification
rate in each group (1 on the horizontal, 2 on the vertical). Each group is of
equal size and identically modeled by the standard normal
\(X \sim \mathcal{N}(Y, 1)\).  Note that states in the left plot attract to
universal non-qualification \(\Pr(Y{=}1){=}0\), while the right plot converges
to universal qualification. The lower plot shows average loss over pairs of randomly sampled episodes.}
  \label{fig:1-inline}
  \vspace{-0.1in}
\end{figure}

\subsection{Results}
Our experiments show that algorithms trained with an RL formulation of long-term fairness can drive a reactive population
toward states with higher utility and fairness, even when short-term utility is \emph{misaligned} with desirable dynamics. Our central
hypothesis, that long-term fairness via RL may induce an algorithm
to sacrifice short-term utility for better long-term outcomes, is concretely
demonstrated by \cref{fig:1-inline}, in which a greedy classifier and \ucbfair,
maximizing true positive rate \(\tp\) (\cref{sec:losses}) subject to demographic
parity \(\DP\) (\cref{eq:DP-2}), drive a population to universal
non-qualification (\(\Pr(Y{=}1) \to 0)\) and universal qualification
(\(\Pr(Y{=}1) \to 1)\), respectively. Each phase plot
shows the dynamics of qualification rates \(q_g = \Pr(Y{=}1\mid G{=}g) \), which parameterize the population state \(\state\) and define the axes, with streamlines; color depicts averaged disparity \(\mathscr{D}\) incurred by \(\action\) in state \(\state\).

\begin{figure}[htb]
  \centering
  \includegraphics[width=0.49\linewidth]{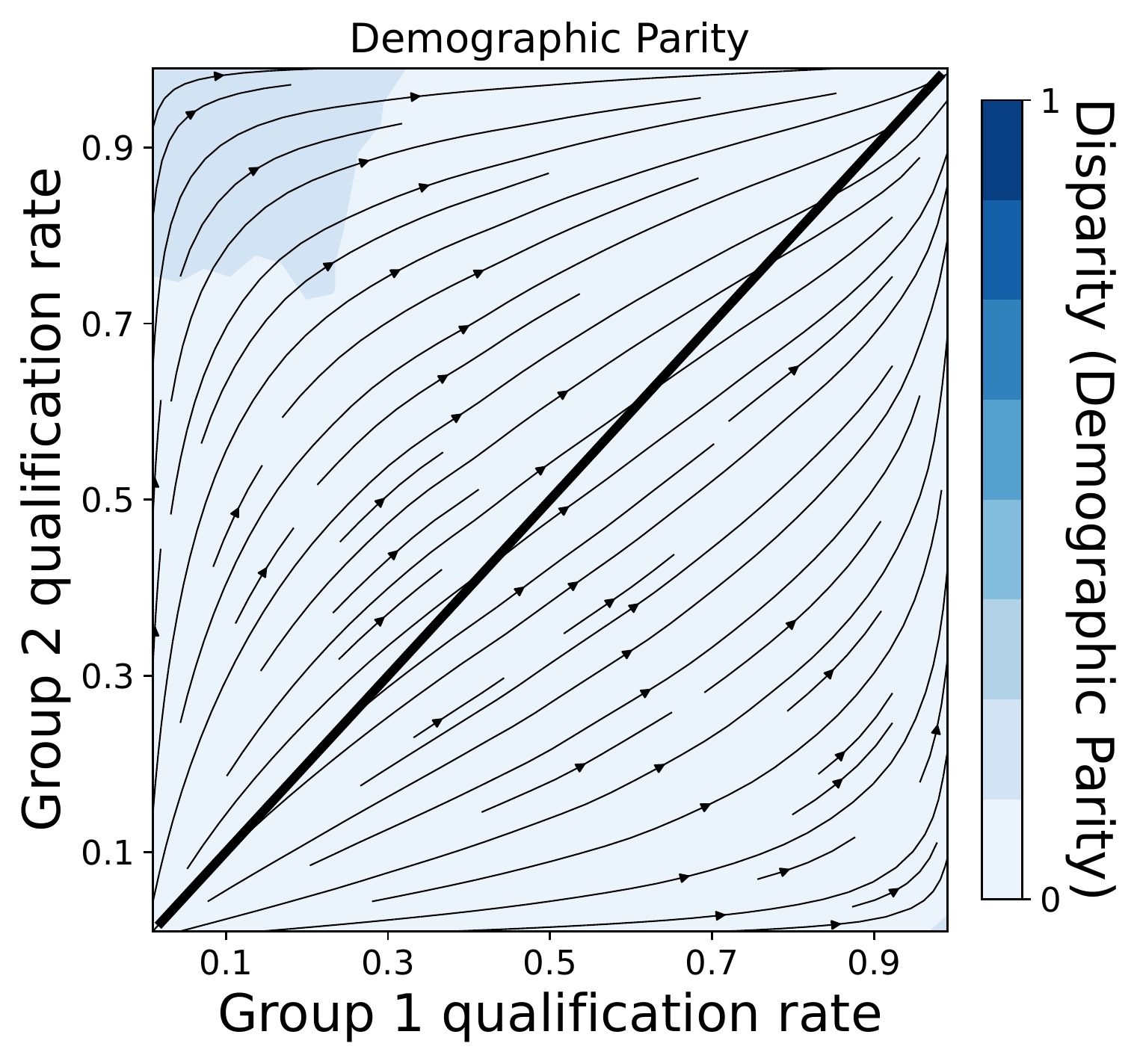}
  \includegraphics[width=0.49\linewidth]{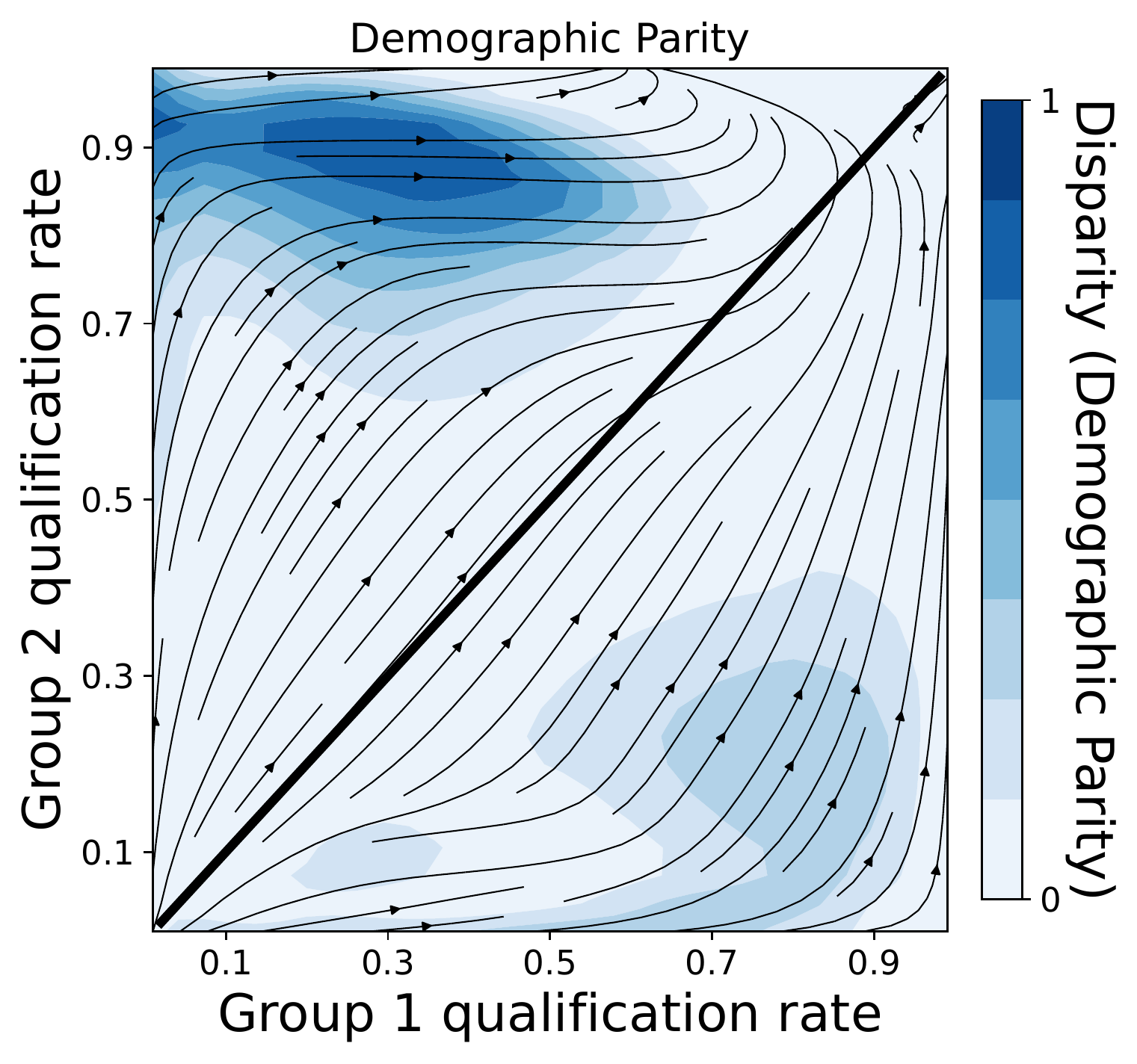}
  \includegraphics[width=0.88\linewidth]{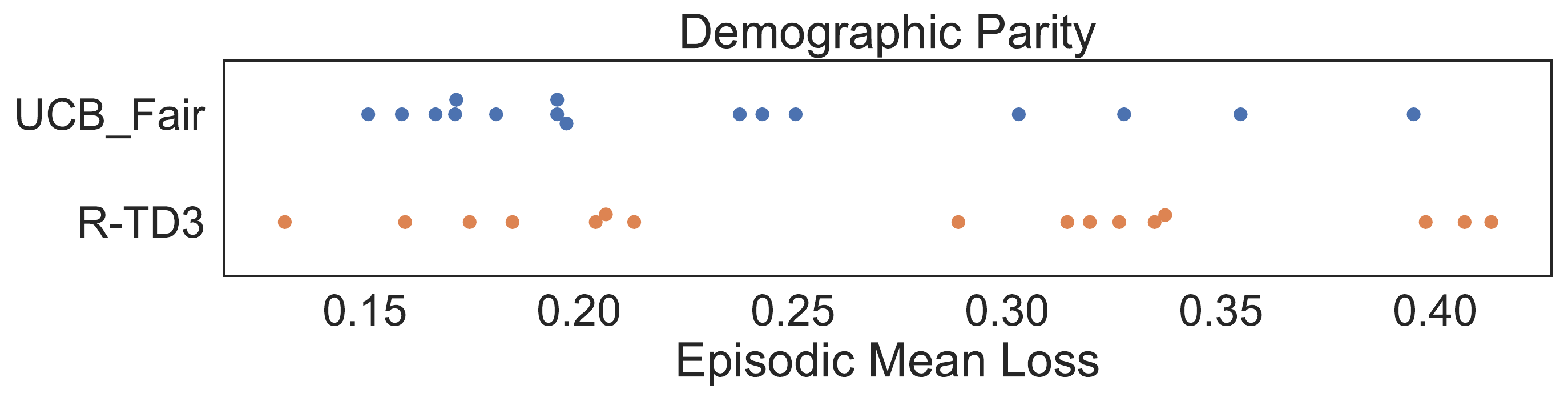}
  \caption{Using a modelled population initialized with the ``Adult'' dataset, reweighted for equal group representation (\cref{sec:synthesis}), \ucbfair (left) and \drl (right) are tasked, as in \cref{fig:1-inline}, to maximize the fraction of true-positive classifications (\(\mathscr{L}=1-\tp\), \cref{eq:loss_function}), subject to demographic parity (\(\mathscr{D}{=}\DP\), \cref{eq:DP-2}). \ucbfair performs almost indistinguishably from the experiment on the synthetic dataset (\cref{fig:1-inline}), while \drl learns qualitatively similar behavior with more aggressive short-term violations of the fairness constraint.}
  \label{fig:2-inline}
\end{figure}
\begin{figure}[htb]
  \centering
    \includegraphics[width=0.49\linewidth]{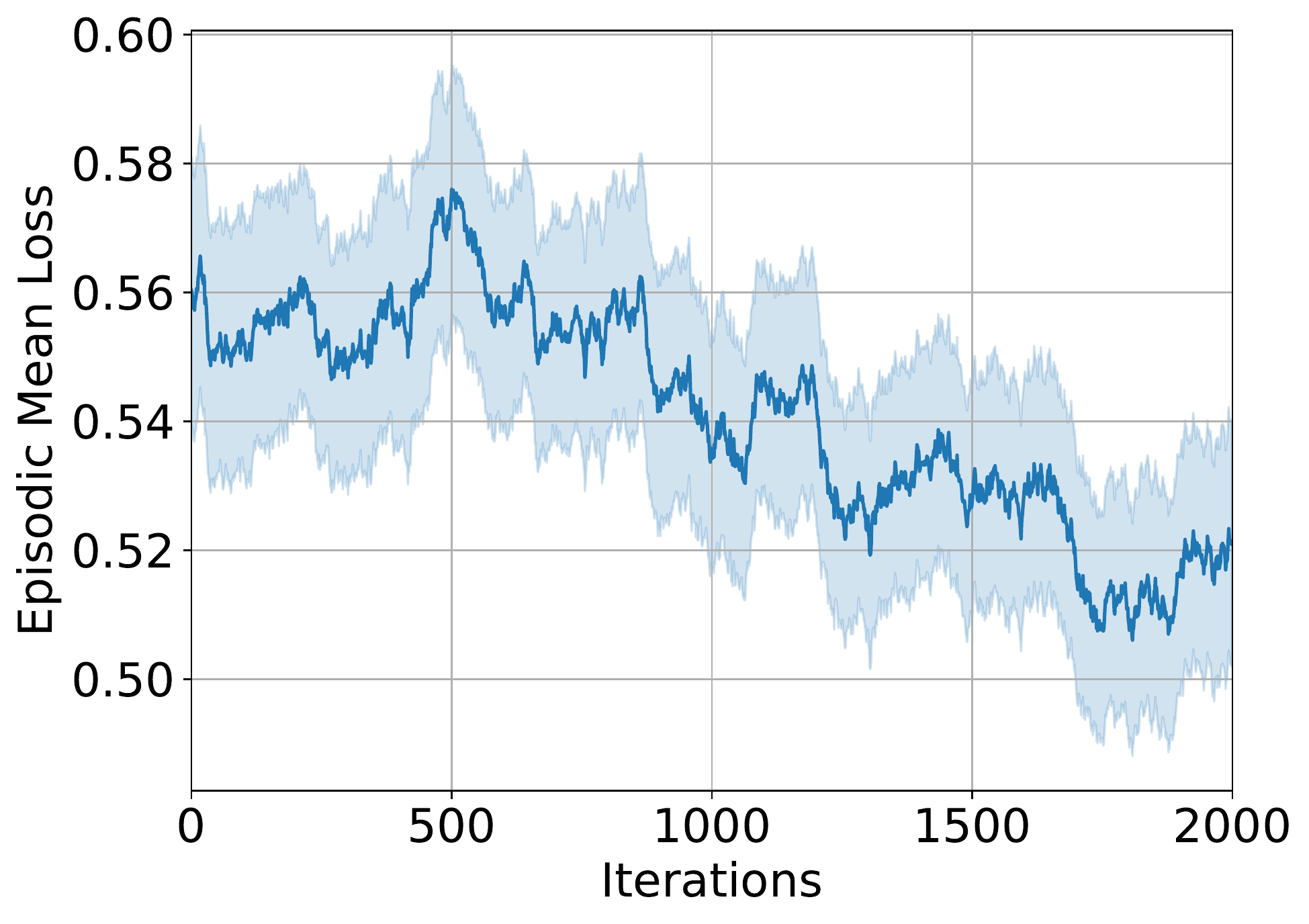}
    \includegraphics[width=0.49\linewidth]{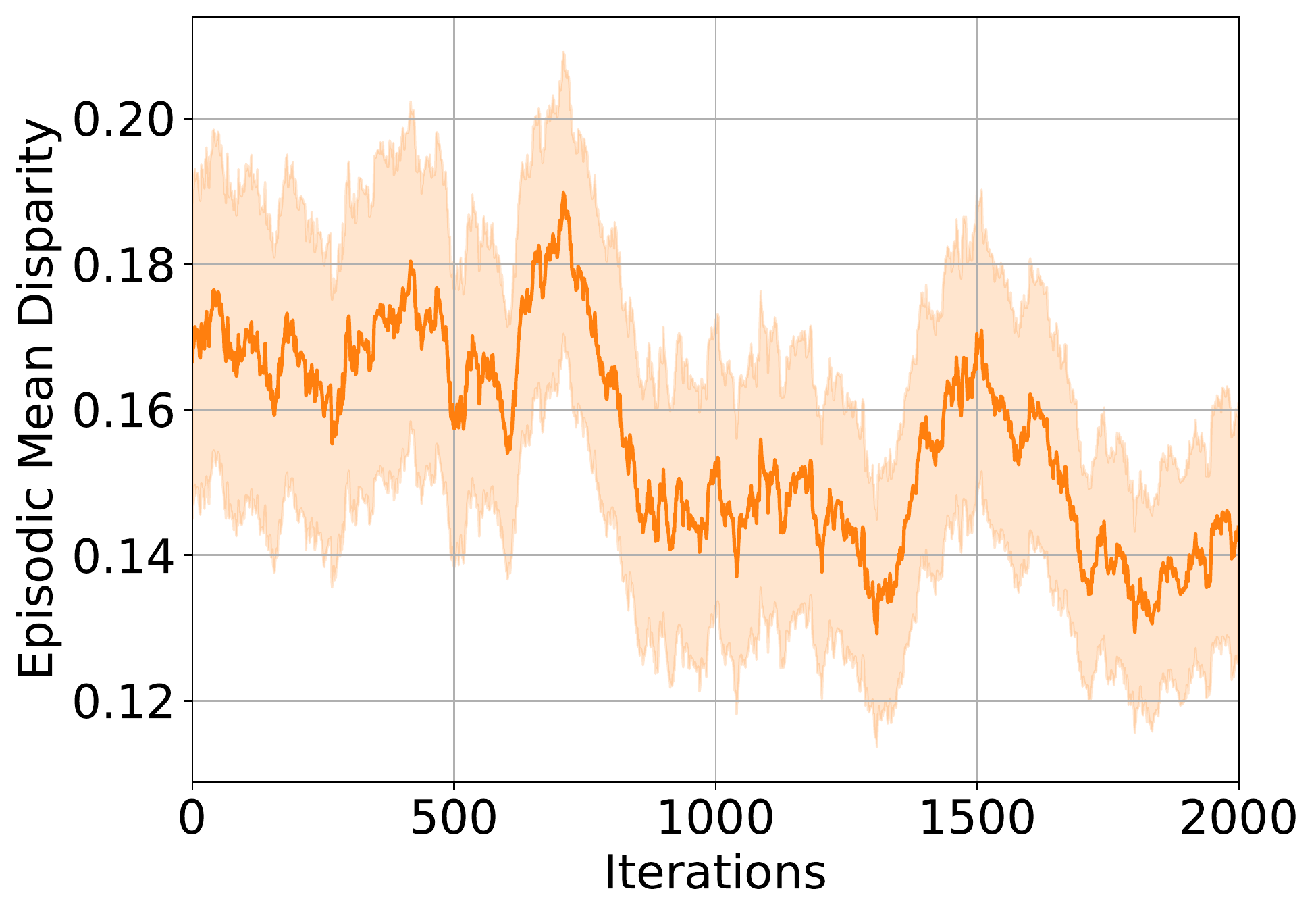}
    \caption{\ucbfair 20-step sliding mean \& std training loss (left) and disparity (right) for the \cref{fig:1-inline} setting.}
  \label{fig:loss-curve-inline}
  \vspace{-0.1in}
\end{figure}

While both algorithms
achieve no or little violation of demographic parity, the myopic algorithm
eventually precludes future true-positive classifications (arrows in \cref{fig:1-inline} approach a low qualification state), while \ucbfair
maintains stochastic thresholds at equilibrium (mean [0.49, 0.38], by group)
with a non-trivial fraction of true-positives. The episodic mean loss and
disparity training curves for \ucbfair are depicted in
\cref{fig:loss-curve-inline}. 

We show that RL algorithms that are not limited by the same restrictive assumptions as \ucbfair are applicable to long-term fairness. In \cref{fig:2-inline}, \drl achieves similar qualitative behavior (\ie, driving near-universal qualification at the expense of short-term utility) when optimizing a loss subject to scheduled disparity regularization. This figure also highlights the lack of guarantees of \drl in incurring prominent violations of the fairness constraint and failing to convincingly asymptote
to the global optimum.

\begin{figure}[htb]
    \centering
      \includegraphics[width=0.49\linewidth]{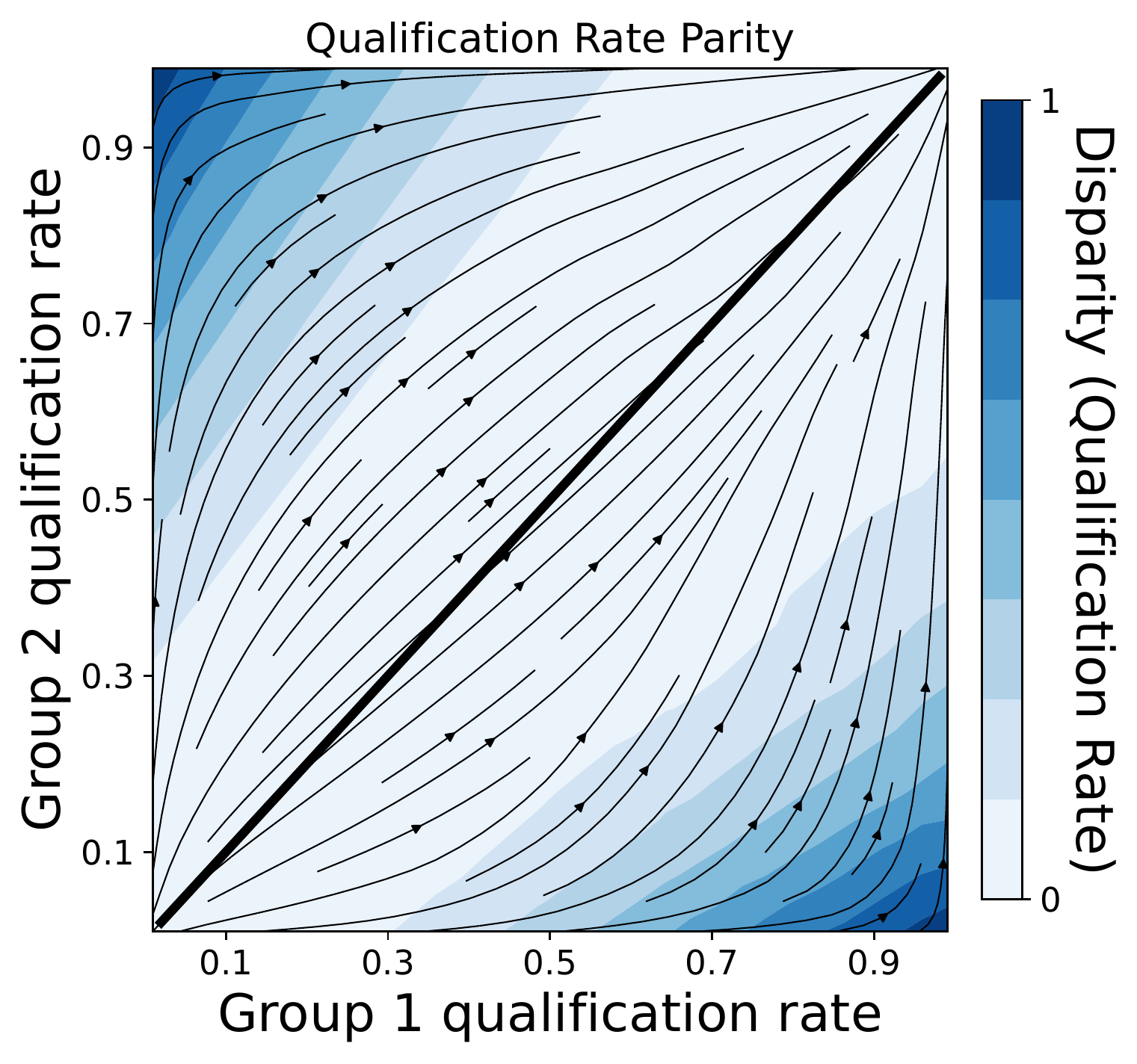}
      \includegraphics[width=0.49\linewidth]{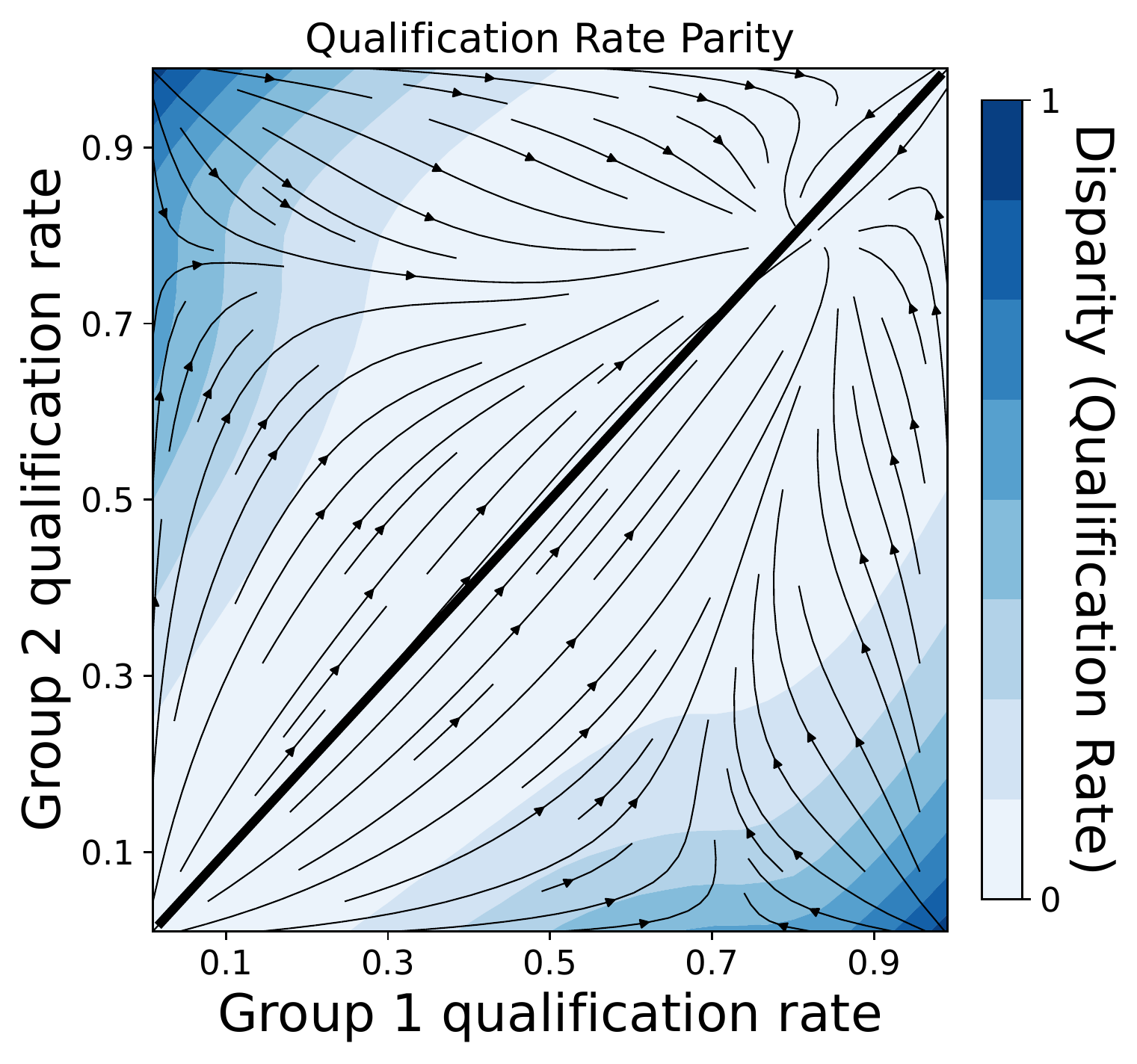}
      \includegraphics[width=0.88\linewidth]{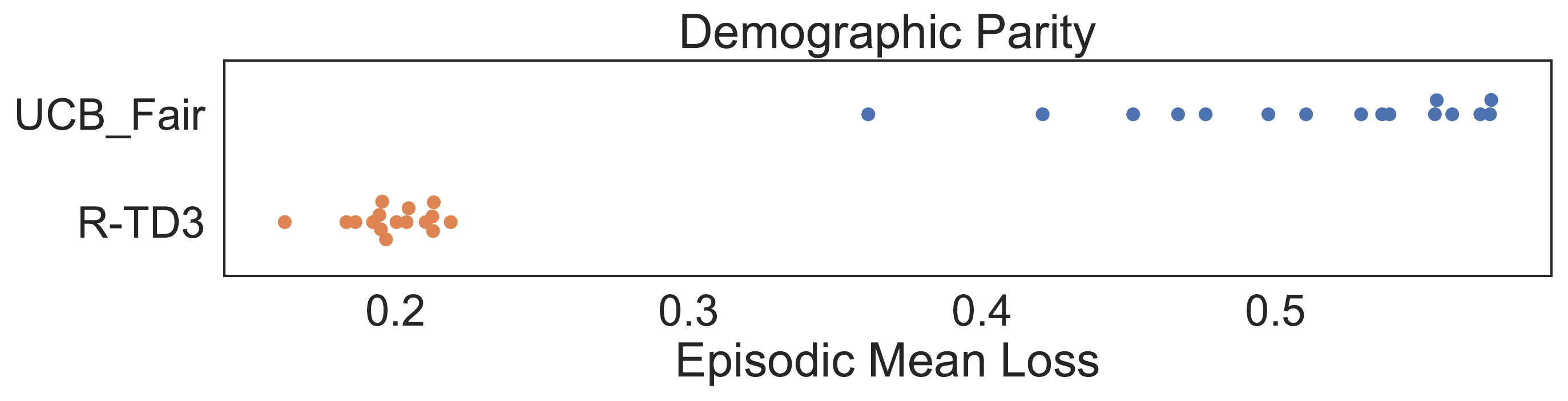}
    \caption{Phase portraits for \ucbfair (left), and \drl (right) interacting on the synthetic distribution \(X \sim \mathcal{N}(Y, 1)\) with groups of equal size. Both algorithms use \(\mathscr{L}=1-\tp-\tn\) (\ie, zero-one loss) and \(\mathscr{D}=\QR\). Shading: qualification rate disparity for the \emph{next} time-step.}
    \label{fig:qr}
    \vspace{-0.1in}
\end{figure}

Finally, we demonstrate the capability of RL to utilize notions of fairness that are impossible to treat in the myopic setting, \viz qualification rate parity, in \cref{fig:qr}. In this example, while both RL agents achieve qualification rate parity, we note that \ucbfair fails to realize the optimal equilibrium discovered by \drl.

\paragraph{Concluding remarks}
We have demonstrated that 
desirable social outcomes that are in conflict with myopic optimization may be realized as solutions to long-term fairness formalized through the lens of reinforcement learning. Moreover, we have shown that this problem both admits solutions with theoretical guarantees and can be relaxed to accommodate a wider class of recent advances in RL. We hope our contributions spur interest in long-term mechanisms and incentive structures for machine learning to be a driver of positive social change.

\paragraph{Acknowledgement}
This work is partially supported by the National Science Foundation (NSF) under grants IIS-2143895 and IIS-2040800, and CCF-2023495.

\newpage
\bibliographystyle{plainnat}


\newpage
\appendix
\onecolumn
\newpage
\section{Proof}
Without loss of generality, we assume $\|\phi(s, a)\| \leq 1$ for all
$(s, a) \in \mathcal{S} \times \mathcal{A}$, and
$\max \left\{\left\|\mu_h(\mathcal{S})\right\|,\left\|\theta_h\right\|\right\} \leq$
$\sqrt{d}$ for all $h \in[H]$.

\subsection{Proof of Theorem \ref{thm:bound}} \label{proof:thmbound}
\bound*
\textbf{Outline}
The outline of this proof simulates the proof in \cite{ghosh2022provably}. For brevity, denote $\mathbb{P}_h V_{j, h+1}^\pi(s, a)=\mathbb{E}_{s^{\prime} \sim \mathbb{P}_h(\cdot \mid s, a)} V_{j, h+1}^\pi\left(s^{\prime}\right) \text { for } j=r, g$. Then 
\begin{align}
    &Q_{j, h}^\pi(s, a)=\left(r_h+\mathbb{P}_h V_{j, h+1}^\pi\right)(s, a)\\
    &V_{j, h}^\pi(s)=\left\langle\pi_h(\cdot \mid s) Q_{j, h}^\pi(s, \cdot)\right\rangle_{\mathcal{A}}\\
    &\left\langle\pi_h(\cdot \mid s), Q_{j, h}^\pi(s, \cdot)\right\rangle_{\mathcal{A}}= \sum_{a \in \mathcal{A}} \pi_h(a \mid s) Q_{j, h}^\pi(s, a)
\end{align}
Similar to \cite{efroni2020exploration}, we establish
\begin{align}
& \text{Regret}(K) + \nu \text{Distortion}(K) \nonumber\\
= &\sum_{k=1}^{K}\left(V_{r, 1}^{\pi^{*}}\left(s_{1}\right)-V_{r, 1}^{\pi_{k}}\left(s_{1}\right)\right)+\nu \sum_{k=1}^{K}\left(b-V_{g, 1}^{\pi_{k}}\left(s_{1}\right)\right) \nonumber\\
\leq & \underbrace{\sum_{k=1}^{K}\left(V_{r, 1}^{\pi^{*}}\left(s_{1}\right)+\nu_{k} V_{g, 1}^{\pi^{*}}\left(s_{1}\right)\right)-\left(V_{r, 1}^{k}\left(s_{1}\right)+\nu_{k} V_{g, 1}^{k}\left(s_{1}\right)\right)}_{\mathcal{T}_{1}} \nonumber \\
& + \underbrace{\sum_{k=1}^{K}\left(V_{r, 1}^{k}\left(s_{1}\right)-V_{r, 1}^{\pi_{k}}\left(s_{1}\right)\right)+\nu \sum_{k=1}^{K}\left(V_{g, 1}^{k}\left(s_{1}\right)-V_{g, 1}^{\pi_{k}}\left(s_{1}\right)\right)}_{\mathcal{T}_{2}} \nonumber\\
& + \underbrace{\frac{1}{2 \eta} \nu^{2}+\frac{\eta}{2} H^{2} K }_{\mathcal{T}_{3}} \nonumber \\
\end{align}
\(\mathcal{T}_{3}\) is easily bounded if \(\eta\). The major task remains bound \(\mathcal{T}_{1}\) and \(\mathcal{T}_{2}\). 

\textbf{Bound \(\mathcal{T}_{1}\) and \(\mathcal{T}_{2}\).} We have following two lemmas.
\begin{restatable}[Boundedness of \(\mathcal{T}_{1}\)]{lemma}{firstpart}\label{lem:firstpart}
    With probability $1-p / 2$, we have $\mathcal{T}_{1} \leq KH \left(\frac{\log (M)}{\alpha} + 2(1+\mathscr{V}) H \rho \epsilon_I\sqrt{\frac{dK}{\varsigma}}\right)$. Specifically, if $\alpha=\frac{\log (M) K}{2(1+\mathscr{V}+H)}$ and $\varsigma = 1$, we have $\mathcal{T}_{1} \leq 2 H(1+\mathscr{V}+H) + 2 K H^2 (1 + \mathscr{V}) \rho \epsilon_I \sqrt{dK}$ with probability $1-p / 2$.
\end{restatable}

\begin{restatable}[Boundedness of \(\mathcal{T}_{2}\)]{lemma}{secondpart}\cite{ghosh2022provably} \label{lem:secondpart}
With probability $1-p / 2, \mathcal{T}_{2} \leq \mathcal{O}\left((\nu+1) H^2 \zeta \sqrt{d^{3}  K }\right)$, where $\zeta=$ $\log [\log (M) 4 d H K / p]$.
\end{restatable}

\cref{lem:secondpart} follows the same logic in \citet{ghosh2022provably}, and we delay the proof of Lemma \ref{lem:firstpart} to Section \ref{sec:t1t2}. Now we are ready to proof Theorem \ref{thm:bound}.

\proof{
For any $\nu \in[0, \mathscr{V}]$, with prob. $1 - p$, 
\begin{align}
& \text{Regret}(K) + \nu \text{Distortion}(K) \nonumber \\
\leq & \mathcal{T}_1 + \mathcal{T}_2 + \mathcal{T}_3 \nonumber\\
\leq & \frac{1}{2 \eta} \nu^{2}+\frac{\eta}{2} H^{2} K + \frac{HK \log M}{\alpha} + 2 K H^2 (1 + \mathscr{V}) \rho \epsilon_I \sqrt{dK} + \mathcal{O}\left((\nu+1) H^2 \zeta \sqrt{d^{3}  K }\right)
\end{align}

Taking $\nu = 0, \eta = \frac{\mathscr{V}}{\sqrt{KH^2}}, \alpha = \frac{K\log M}{2(1 + \mathscr{V} + H)}, \epsilon_I = \frac{1}{2\rho (1 + \mathscr{V}) KH\sqrt{d}}$, there exist constant $b$,
\begin{align}
\text{Regret}(K) &\leq \frac{\mathscr{V}H}{2} \sqrt{K} + 2 H(1+\mathscr{V}+H) + 2 H^2K (1 + \mathscr{V}) \rho \epsilon_I \sqrt{dK} + \mathcal{O}\left( H^2 \zeta \sqrt{d^{3}  K }\right)\nonumber\\
& \leq \big( b \zeta H^2 \sqrt{d^{3}}+ (\mathscr{V} + 1) H \big) \sqrt{K} = \mathcal{\tilde{O}}(H^2 \sqrt{d^3 K}). \nonumber
\end{align}
Taking $\nu = \mathscr{V}, \eta = \frac{\mathscr{V}}{\sqrt{KH^2}}, \alpha = \frac{K\log M}{2(1 + \mathscr{V} + H)}, \epsilon_I = \frac{1}{2\rho (1 + \mathscr{V}) KH\sqrt{d}}$,
\begin{align}
\text{Regret}(K) + \mathscr{V} \text{Distortion}(K) &\leq (\mathscr{V} + 1) H \sqrt{K} + (1 + \mathscr{V}) \mathcal{O}\left( H^2 \zeta \sqrt{d^{3}  K }\right)\nonumber
\end{align}
Following the idea of \citet{efroni2020exploration}, there exists a policy $\pi^{\prime}$ such that $V_{r, 1}^{\pi^{\prime}}=\frac{1}{K} \sum_{k=1}^{K} V_{r, 1}^{\pi_{k}}, V_{g, 1}^{\pi^{\prime}}=\frac{1}{K} \sum_{k=1}^{K} V_{g, 1}^{\pi_{k}}$. By the occupancy measure, $V_{r, 1}^{\pi}$ and $V_{g, 1}^{\pi}$ are linear in occupancy measure induced by $\pi$. Thus, the average of $K$ occupancy measure also produces an occupancy measure which induces policy $\pi^{\prime}$ and $V_{r, 1}^{\pi^{\prime}}$, and $V_{g, 1}^{\pi^{\prime}}$. We take $\nu=0$ when $\sum_{k=1}^{K}\left(b-V_{g, 1}^{\pi_{k}}\left(s_{1}^{k}\right)\right)<0$, otherwise $\nu=\mathscr{V}$. Hence, we have
\begin{align}
&V_{r, 1}^{\pi^{*}}\left(s_{1}\right)-\frac{1}{K} \sum_{k=1}^{K} V_{r, 1}^{\pi_{k}}\left(s_{1}\right)+\mathscr{V} \max\big((c-\frac{1}{K} \sum_{k=1}^{K} V_{g, 1}^{\pi_{k}}\left(s_{1}\right), 0\big) \nonumber\\
&= V_{r, 1}^{\pi^{*}}\left(s_{1}\right)-V_{r, 1}^{\pi^{\prime}}\left(s_{1}\right)+\mathscr{V} \max \big(c -V_{g, 1}^{\pi^{\prime}}\left(s_{1}\right), 0 \big) \nonumber\\
&\leq \frac{\mathscr{V} + 1}{K} H \sqrt{K} + \frac{\mathscr{V} + 1}{K} \mathcal{O}\left( H^2 \zeta \sqrt{d^{3}  K }\right)
\end{align}

Since $\mathscr{V}=2 H / \gamma$, and using the result of \cref{lem:duality}, we have
$$
\max \big( c -\frac{1}{K} \sum_{k=1}^{K} V_{g, 1}^{\pi_{k}}\left(s_{1}^{k}\right), 0\big) \leq \frac{\mathscr{V} + 1}{K\mathscr{V}} \mathcal{O}\left( H^2 \zeta \sqrt{d^{3}  K }\right)
$$

In this section we proof \cref{lem:firstpart} and \cref{lem:secondpart}.  

\subsection{Prepare for \cref{lem:firstpart}}
In order to bound \(\mathcal{T}_{1}\) and \(\mathcal{T}_{2}\), we introduce the following lemma.
\begin{restatable}[]{lemma}{fluctuation}\label{lem:fluctuation}
There exists a constant $B_{2}$ such that for any fixed $p \in(0,1)$, with probability at least $1-p / 2$, the following event holds
$$
\| \sum_{\tau=1}^{k-1} \phi_{j, h}^{\tau}\left[V_{j, h+1}^{k}\left(s_{h+1}^{\tau}\right)-\mathbb{P}_{h} V_{j, h+1}^{k}\left(s_{h}^{\tau}, a_{h}^{\tau}\right) \|_{\left(\varsigma_{h}^{k}\right)^{-1}} \leq B_{2} d H q \right.
$$
for $j \in\{r, g\}$, where $q = \sqrt{\log \left[4\left(B_{1}+1\right) \log (M) d T / p\right]}$ for some constant $B_{1}$.
\end{restatable}
We delay the proof of \cref{lem:fluctuation} to \cref{proof-of-fluctuation}.

\cref{lem:fluctuation} shows the bound of estimated value function $V^{k}_{j, h}$ and value function $V^{\pi}_{j, h}$ corresponding in a given policy at k.
We now introcuce the following lemma appeared in \citet{ghosh2022provably}. This lemma bounds the difference between the value function without
bonus in \ucbfair and the true value function of any policy $\pi$. This is bounded using their expected difference at next step, plus a error term.

\begin{lemma}\cite{ghosh2022provably}\label{lem:ucb1}
There exists an absolute constant $\beta=C_{1} d H \sqrt{\zeta}, \zeta=\log (\log (M) 4 d T / p)$, and for any fixed policy $\pi$, for the event defined in \cref{lem:fluctuation}, we have
$$
\left\langle\phi(s, a), w_{j, h}^{k}\right\rangle-Q_{j, h}^{\pi}(s, a)=\mathbb{P}_{h}\left(V_{j, h+1}^{k}-V_{j, h+1}^{\pi}\right)(s, a)+\Delta_{h}^{k}(s, a)
$$
for some $\Delta_{h}^{k}(s, a)$ that satisfies $\left|\Delta_{h}^{k}(s, a)\right| \leq \beta \sqrt{\phi(s, a)^{T}\left(\Lambda_{h}^{k}\right)^{-1} \phi(s, a)}$.
\end{lemma}

\begin{lemma} \cite{ghosh2022provably}\label{lem:ucb2}
With probability at least $1-p / 2$, (for the event defined in \cref{lem:fluctuation})
$$
Q_{r, h}^\pi(s, a)+\nu_k Q_{g, h}^\pi(s, a) \leq Q_{r, h}^k(s, a)+\nu_k Q_{g, h}^k(s, a)-\mathbb{P}_h\left(V_{h+1}^k-V_{h+1}^{\pi, \nu_k}\right)(s, a)
$$
\end{lemma}

We also introduce the following lemma. This lemma bound the value function by taking \ucbfair policy and greedy policy.

\begin{lemma}\label{lem:greedy}
Define \(\bar{V}_h^k(\cdot)=\max _a\left[Q_{r, h}^k(\cdot, a)+\nu_k Q_{g, h}^k(\cdot, a)\right] \) the value function corresponding to greedy policy, we have
\begin{equation}
\bar{V}_{h}^{k}(s)-V_{h}^{k}(s) \leq \frac{\log M}{\alpha} + 2(1+\mathscr{V}) H \rho \epsilon_I\sqrt{\frac{dk}{\varsigma}}.
\end{equation}
\end{lemma}
\proof{
Define $a_g$ the solution of greedy policy, 
    \begin{align}
        V_h^k(s)-V_h^k(s)=&\left[Q_{r, h}^k\left(s, a_g\right)+\nu_k Q_{g, h}^k\left(s, a_g\right)\right] \\
        & -\int_a \pi_{h, k}(a \mid s)\left[Q_{r, h}^k(s, a)+\nu_k Q_{g, h}^k(s, a)\right] da\\
        \leq &\left[Q_{r, h}^k\left(s, a_g\right)+\nu_k Q_{g, h}^k\left(s, a_g\right)\right] \\
        & -\sum_i \text{SM}_{\alpha}(I_i \mid x)\left[Q_{r, h}^k(x, I_i)+\nu_k Q_{g, h}^k(x, I_i)\right] + 2(1+\mathscr{V}) H \rho \epsilon_I\sqrt{\frac{dk}{\varsigma}}\\
         \leq&\left(\frac{\log \left(\sum_a \exp \left(\alpha\left(Q_{r, h}^k(s, I_i)+\nu_k Q_{g, h}^k(s, I_i)\right)\right)\right)}{\alpha}\right)\\
        & -\sum_i \text{SM}_{\alpha}(I_i \mid s)\left[Q_{r, h}^k(s, I_i)+\nu_k Q_{g, h}^k(s, I_i)\right]  + 2(1+\mathscr{V}) H \rho \epsilon_I\sqrt{\frac{dk}{\varsigma}}\\
         \leq & \frac{\log (M)}{\alpha} + 2(1+\mathscr{V}) H \rho \epsilon_I\sqrt{\frac{dk}{\varsigma}}.
\end{align}
The first inequality follows from \cref{lem:bound-of-q-l} and the second inequality holds because of Proposition 1 in \citet{pan2019reinforcement}.
}

\subsection{Proof of \cref{lem:firstpart}}
Now we're ready to proof \cref{lem:firstpart}.
\proof{This proof simulates Lemma 3 in \citet{ghosh2022provably}. 

We use induction to proof this lemma.
At step $H$, we have $Q_{j, H+1}^k=0=Q_{j, H+1}^\pi$ by definition. Under the event in \cref{lem:cover-for-value-function} and using \cref{lem:ucb1}, we have for $j=r, g$,
$$
\left|\left\langle\phi(s, a), w_{j, H}^k(s, a)\right\rangle-Q_{j, H}^\pi(s, a)\right| \leq \beta \sqrt{\phi(s, a)^T\left(\Lambda_H^k\right)^{-1} \phi(s, a)}
$$
Thus \(
Q_{j, H}^\pi(s, a) \leq \min \left\{\left\langle\phi(s, a), w_{j, H}^k\right\rangle+\beta \sqrt{\phi(s, a)^T\left(\Lambda_H^k\right)^{-1} \phi(s, a)}, H\right\} =Q_{j, H}^k(s, a)
\).

From the definition of $\bar{V}_h^k$,
$$
\bar{V}_H^k(s) =\max _a\left[Q_{r, H}^k(s, a)+\nu_k Q_{g, h}^k(s, a)\right] \geq \sum_a \pi(a \mid x)\left[Q_{r, H}^\pi(s, a)+\nu_k Q_{g, H}^\pi(s, a)\right] =V_H^{\pi, \nu_k}(s)
$$
for any policy $\pi$. Thus, it also holds for $\pi^*$, the optimal policy. Using \cref{lem:greedy} we can get 
$$
V_H^{\pi^*, \nu_k}(s)-V_H^k(s) \leq \frac{\log M}{\alpha} + 2(1+\mathscr{V}) H \rho \epsilon_I\sqrt{\frac{dk}{\varsigma}}
$$
Now, suppose that it is true till the step $h+1$ and consider the step $h$.
Since, it is true till step $h+1$, thus, for any policy $\pi$,
$$
\mathbb{P}_h\left(V_{h+1}^{\pi, \nu_k}-V_{h+1}^k\right)(s, a) \leq (H-h) \big( \frac{\log M}{\alpha} + 2(1+\mathscr{V}) H \rho \epsilon_I\sqrt{\frac{dk}{\varsigma}} \big)
$$
From (27) in Lemma 10 and the above result, we have for any $(s, a)$
$$
Q_{r, h}^\pi(s, a)+\nu_k Q_{g, h}^\pi(s, a) \leq Q_{r, h}^k(s, a)+\nu_k Q_{g, h}^k(s, a)+(H-h) \big( \frac{\log M}{\alpha} + 2(1+\mathscr{V}) H \rho \epsilon_I\sqrt{\frac{dk}{\varsigma}} \big)
$$
Hence,
$$
V_h^{\pi, \nu_k}(s) \leq \bar{V}_h^k(s)+(H-h) \big( \frac{\log M}{\alpha} + 2(1+\mathscr{V}) H \rho \epsilon_I\sqrt{\frac{dk}{\varsigma}} \big)
$$
Now, again from Lemma 11, we have $\bar{V}_h^k(s)-V_h^k(s) \leq \frac{\log (|\mathcal{A}|)}{\alpha}$. Thus,
$$
V_h^{\pi, \nu_k}(s)-V_h^k(s) \leq (H-h+1) \big( \frac{\log M}{\alpha} + 2(1+\mathscr{V}) H \rho \epsilon_I\sqrt{\frac{dk}{\varsigma}} \big)
$$
Now, since it is true for any policy $\pi$, it will be true for $\pi^*$. From the definition of $V^{\pi, \nu_k}$, we have
$$
\left(V_{r, h}^{\pi^*}(s)+\nu_k V_{g, h}^{\pi^*}(s)\right)-\left(V_{r, h}^k(s)+\nu_k V_{g, h}^k(s)\right) \leq (H-h+1) \big( \frac{\log M}{\alpha} + 2(1+\mathscr{V}) H \rho \epsilon_I\sqrt{\frac{dk}{\varsigma}} \big)
$$
Hence, the result follows by summing over $K$ and considering $h=1$.

}

\subsection{Proof of \cref{lem:fluctuation}}\label{proof-of-fluctuation}
We first define some useful sets. Let $\mathcal{Q}_{j}=\left\{Q \mid Q(\cdot, a)=\min \left\{w_{j}^{T} \phi(\cdot, a)+\beta \sqrt{\phi^{T}(\cdot, a)^{T} \Lambda^{-1} \phi(\cdot, a)}, H\right\},\ a\in \mathcal{A}\right\}$ be the set of Q functions, where \(j \in \{r, g\}\). Since the minimum eigen value of \(\Lambda\) is no smaller than one so the Frobenius norm of \(\Lambda^{-1}\) is bounded.

Let $\mathcal{V}_{j}=\left\{V_{j} \mid V_{j}(\cdot)=\int_{a} \pi(a \mid \cdot) Q_{j}(\cdot, a) da; Q_{r} \in \mathcal{Q}_{r}, Q_{g} \in \mathcal{Q}_{g}, \nu \in[0, \mathscr{V}]\right\}$ be the set of Q functions, where \(j \in \{r, g\}\). Define
$$
\Pi=\left\{\pi \mid \forall a \in \mathcal{A}, \pi(a \mid \cdot)=\frac{1}{\int_{b\in\mathcal{I}(a)} db}\text {SM}_{\alpha}\left(Q_{r}(\cdot, I(a))+\nu Q_{g}(\cdot, I(a))\right),\  Q_{r} \in \mathcal{Q}_{r}, Q_{g} \in \mathcal{Q}_{g}, \nu \in[0, \mathscr{V}]\right\}
$$
the set of policies. 

It's easy to verify \(V_j^k \in \mathcal{V}_j\).

Then we introduce the proof of \cref{lem:fluctuation}. To proof \cref{lem:fluctuation}, we need the $\epsilon$-covering number for the set of value functions(\cref{lem:cover-for-value-function}\cite{ghosh2022provably}). To achieve this, we need to show if two $Q$ functions and the dual variable $\nu$ are close, then the bound of policy and value function can be derived(\cref{lem:bound-of-policy}, \cref{lem:bound-of-v}). Though the proof of \cref{lem:bound-of-policy} and \cref{lem:bound-of-v} are different from \citet{ghosh2022provably}, we show the results remain the same, thus \cref{lem:cover-for-value-function} still holds. We'll only introduce \cref{lem:cover-for-value-function} and omit the proof.

We now proof \cref{lem:bound-of-policy}.
\begin{lemma}\label{lem:bound-of-policy}
Let \(\pi\) be the policy of \ucbfair corresponding to \(Q^k_r + \nu_k Q^k_g\), i.e., 
\begin{equation}
    \pi(a \mid \cdot)=\frac{1}{\int_{b\in\mathcal{I}(a)} db}\text {SM}_{\alpha}\left(Q_{r}(\cdot, I(a))+\nu Q_{g}(\cdot, I(a))\right)
\end{equation}

and 
\begin{equation}
    \tilde{\pi}(a \mid \cdot)=\frac{1}{\int_{b\in\mathcal{I}(a)} db}\text {SM}_{\alpha}\left(\tilde{Q}_r(\cdot, I(a))+\tilde{\nu} \tilde{Q}_g(\cdot, I(a))\right),
\end{equation}

if \(\left|Q_{j} - \tilde{Q}_j\right| \leq \epsilon^{\prime}\) and \(\left|\nu - \tilde{\nu}\right| \leq \epsilon^{\prime}\), then $\left|\int_{a} \left(\pi(a \mid x)  - \tilde{\pi}(a \mid x)\right) da \right| \leq 2 \alpha \epsilon^{\prime}(1+\mathscr{V}+H)$.
\end{lemma}
\proof{
\begin{align}
&\left|\int_{a} \left(\pi(a \mid x)  - \tilde{\pi}(a \mid x)\right) da \right| \\
=&\left|\sum_{i=1}^{M}\int_{a\in\mathcal{I}_i} \left(\pi(I(a) \mid x)  - \tilde{\pi}(I(a) \mid x)\right) da \right| \nonumber\\
=&\left|\sum_{i=1}^{M} \int_{b\in \mathcal{I}_i} db \left(\pi(I_i \mid x) - \tilde{\pi}(I_i \mid x)\right)\right| \nonumber\\
\leq &\sum_{i=1}^{M} \left| {SM}_{\alpha}\big(Q_{r}(s, I_i)+\nu Q_{g}(s, I_i)\big) - {SM}_{\alpha}\big(\tilde{Q}_{r}(s, I_i)+\tilde{\nu} \tilde{Q}_{g}(s, I_i)\big)\right| \nonumber\\
\leq & 2 \alpha \left| Q_r(\cdot, I(a))+\nu Q_g(\cdot, I(a)) - \tilde{Q}_r(\cdot, I(a))-\tilde{\nu} \tilde{Q}_g(\cdot, I(a)) \right|
\end{align}

The last inequaity holds because of Theorem 4.4 in \citet{epasto2020optimal}. Using \cref{cor:bound}, we have 
\begin{equation}
\left|\int_{a} \left(\pi(a \mid x)  - \tilde{\pi}(a \mid x)\right) da \right| \leq 2 \alpha \epsilon^{\prime}(1+\mathscr{V}+H)
\end{equation}
}

Now since we have \cref{lem:bound-of-policy}, we can further bound the value functions.

\begin{lemma}\label{lem:bound-of-v}
If $\left|\tilde{Q}_{j} - Q_{j}^{k}\right| \leq \epsilon^{\prime}$, where $\tilde{Q}_{j} \in \mathcal{Q}_{j}$, then there exists $\tilde{V}_{j} \in \mathcal{V}_{j}$ such that
$$
\left|V_{j}^{k} - \widetilde{V}_{j}\right| \leq H 2 \alpha \epsilon^{\prime}(1+\mathscr{V}+H)+\epsilon^{\prime},
$$
\end{lemma}

\proof{
For any $x$,
$$
\begin{aligned}
&V_{j}^{k}(s)-\widetilde{V}_{j}(s) \\
&=\left|\int_{a} \pi(a \mid s) Q_{j}^{k}(s, a) da -\int_{a} \tilde{\pi}(a \mid s) \tilde{Q}_{j}(s, a) da\right| \\
&=\left|\int_{a} \pi(a \mid s) Q_{j}^{k}(s, a) da -\int_{a} \pi(a \mid s) \tilde{Q}_{j}(s, a) da +\int_{a} \pi(a \mid s) \tilde{Q}_{j}(s, a) da -\int_{a} \tilde{\pi}(a \mid s) \tilde{Q}_{j}(s, a) da \right| \\
&\leq\left|\int_{a} \pi(a \mid s) \left( Q_{j}^{k}(s, a) - \tilde{Q}_{j}(s, a)\right) da \right|+\left|\int_{a} \pi(a \mid s) \tilde{Q}_{j}(s, a) da -\int_{a} \tilde{\pi}(a \mid s) \tilde{Q}_{j}(s, a) da\right| \\
&\leq \epsilon^{\prime}+ H\left|\int_{a} \left(\pi(a \mid s)  - \tilde{\pi}(a \mid s)\right) da \right|\\
&\leq \epsilon^{\prime}+H 2 \alpha \epsilon^{\prime}(1+\mathscr{V}+H)
\end{aligned}
$$
}

Using Lemmas above, we can have the same result presented in Lemma 13 of \citet{ghosh2022provably} as following.
\begin{lemma}\cite{ghosh2022provably}\label{lem:cover-for-value-function}
There exists a $\tilde{V}_{j} \in \mathcal{V}_{j}$ parameterized by $\left(\tilde{w}_{r}, \tilde{w}_{g}, \tilde{\beta}, \Lambda, \tilde{\mathscr{V}}\right)$ such that dist $\left(V_{j}, \tilde{V}_{j}\right) \leq \epsilon$ where
$$
\left| V_{j} - \tilde{V}_{j}\right|=\sup _{x}\left|V_{j}(s)-\tilde{V}_{r}(s)\right| .
$$
Let $N_{\epsilon}^{V_{j}}$ be the $\epsilon$-covering number for the set $\mathcal{V}_{j}$, then,
$$
\log N_{\epsilon}^{V_{j}} \leq d \log \left(1+8 H \frac{\sqrt{d k}}{\sqrt{\varsigma} \epsilon^{\prime}}\right)+d^{2} \log \left[1+8 d^{1 / 2} \beta^{2} /\left(\varsigma\left(\epsilon^{\prime}\right)^{2}\right)\right]+\log \left(1+\frac{\mathscr{V}}{\epsilon^{\prime}}\right)
$$
where $\epsilon^{\prime}=\frac{\epsilon}{H 2 \alpha(1+\mathscr{V}+H)+1}$
\end{lemma}

\begin{lemma}\label{lem:bound-of-q-l}
$|Q^{k}_{j,h}(s, a) - Q^{k}_{j,h}(s, I(a))| \leq 2H\rho \epsilon_{I} \sqrt{\frac{dK}{\varsigma}}$.
\end{lemma}
\proof{
\begin{align}
    & \Big|Q^{k}_{j,h}(s, a) - Q^{k}_{j,h}(s, I(a))\Big| \\
    = & \Big| w^{k}_{j,h}(s, a)^T (\phi(s, a) - \phi(s, I(a))) \Big| \\
    \leq & || w^{k}_{j,h}(s, a) \|_2 \| \phi(s, a) - \phi(s, I(a)) ||_2\\
\end{align}
From \cref{lem:w} and \cref{asm:lipschitz} we get the result.
}

\subsection{Preliminary Results}
\begin{lemma}\cite{ding2021provably}\label{lem:duality}
Let $\nu^*$ be the optimal dual variable, and $C \geq 2 \nu^*$, then, if
$$
V_{r, 1}^{\pi^*}\left(s_1\right)-V_{r, 1}^\pi\left(s_1\right)+C\left[c-V_{g, 1}^\pi\left(s_1\right)\right]_{+} \leq \delta,
$$
we have 
$$
\left[c-V_{g, 1}^{\tilde{\pi}}\left(x_1\right)\right]_{+} \leq \frac{2 \delta}{C} .
$$
\end{lemma}

\begin{lemma}\cite{jin2020provably} \label{lem:w} For any $(k, h)$, the weight $w_{j, h}^k$ satisfies
$$
\left\|w_{j, h}^k\right\| \leq 2 H \sqrt{d k / \varsigma}
$$
\end{lemma}

\begin{corollary}\label{cor:bound}
If $\operatorname{dist}\left(Q_{r}, \tilde{Q}_{r}\right) \leq \epsilon^{\prime}, \operatorname{dist}\left(Q_{g}, \tilde{Q}_{g}\right) \leq \epsilon^{\prime}$, and $\left|\tilde{\nu}_{k}-\nu_{k}\right| \leq \epsilon^{\prime}$, then, $\operatorname{dist}\left(Q_{r}^{k}+\right.$ $\left.\nu_{k} Q_{g}^{k}, \tilde{Q}_{r}+\tilde{\nu}_{k} \tilde{Q}_{g}\right) \leq \epsilon^{\prime}(1+\mathscr{V}+H)$.
\end{corollary}

\newpage
\section{Additional Figures} \label{app:morefigs}

\begin{figure}[h]
  \centering
      \includegraphics[width=0.8\linewidth, trim=1.2cm 0cm 0cm 0cm,
clip]{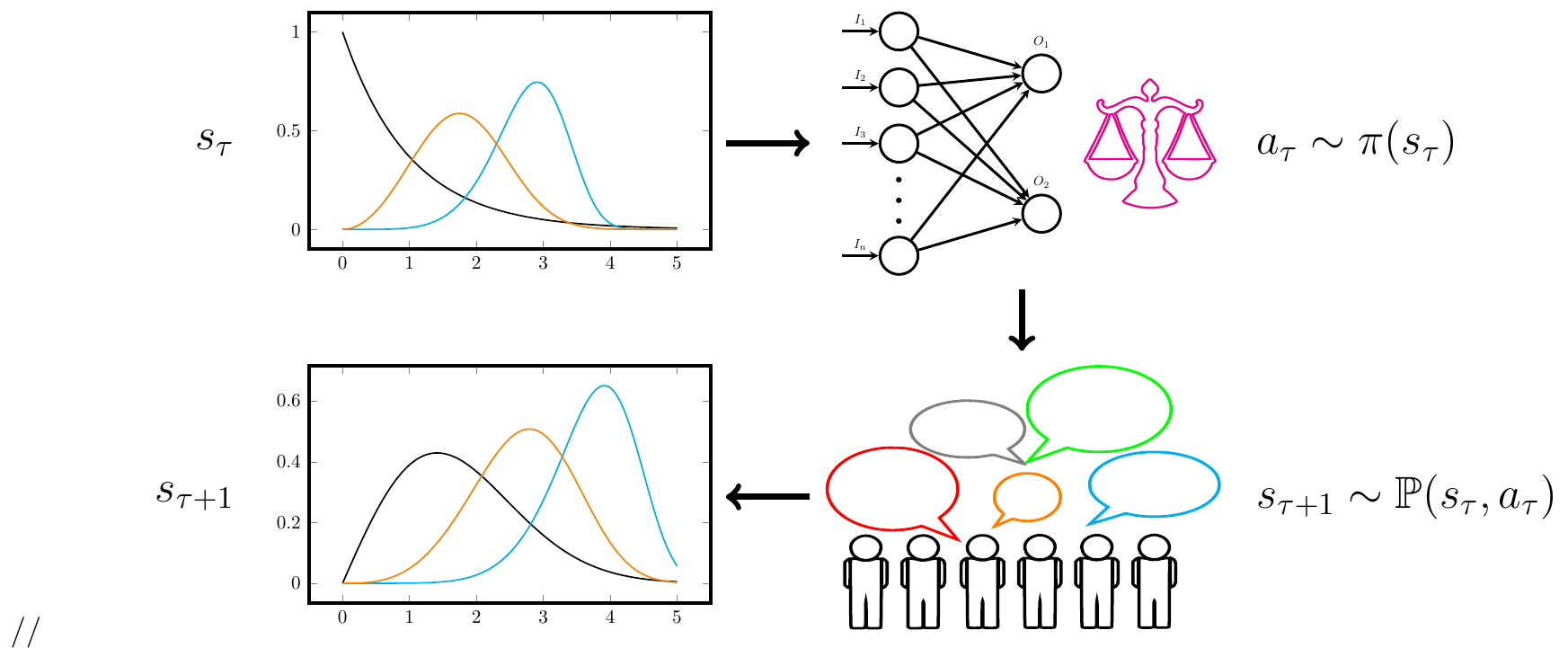}
      \caption{The interaction of an algorithmic classifier and a reactive
population. Given state \(\state_{\tau}\), the classifier uses policy \(\pi\) to
select action \(\action_{\tau}\). The population, in state \(\state_{\tau}\),
reacts to \(\action_{\tau}\) , transitioning state to \(\state_{\tau + 1}\),
then the process repeats.}
  \label{fig:overview}
\end{figure}

\begin{figure}[h]
\centering
\includegraphics[width=0.8\linewidth]{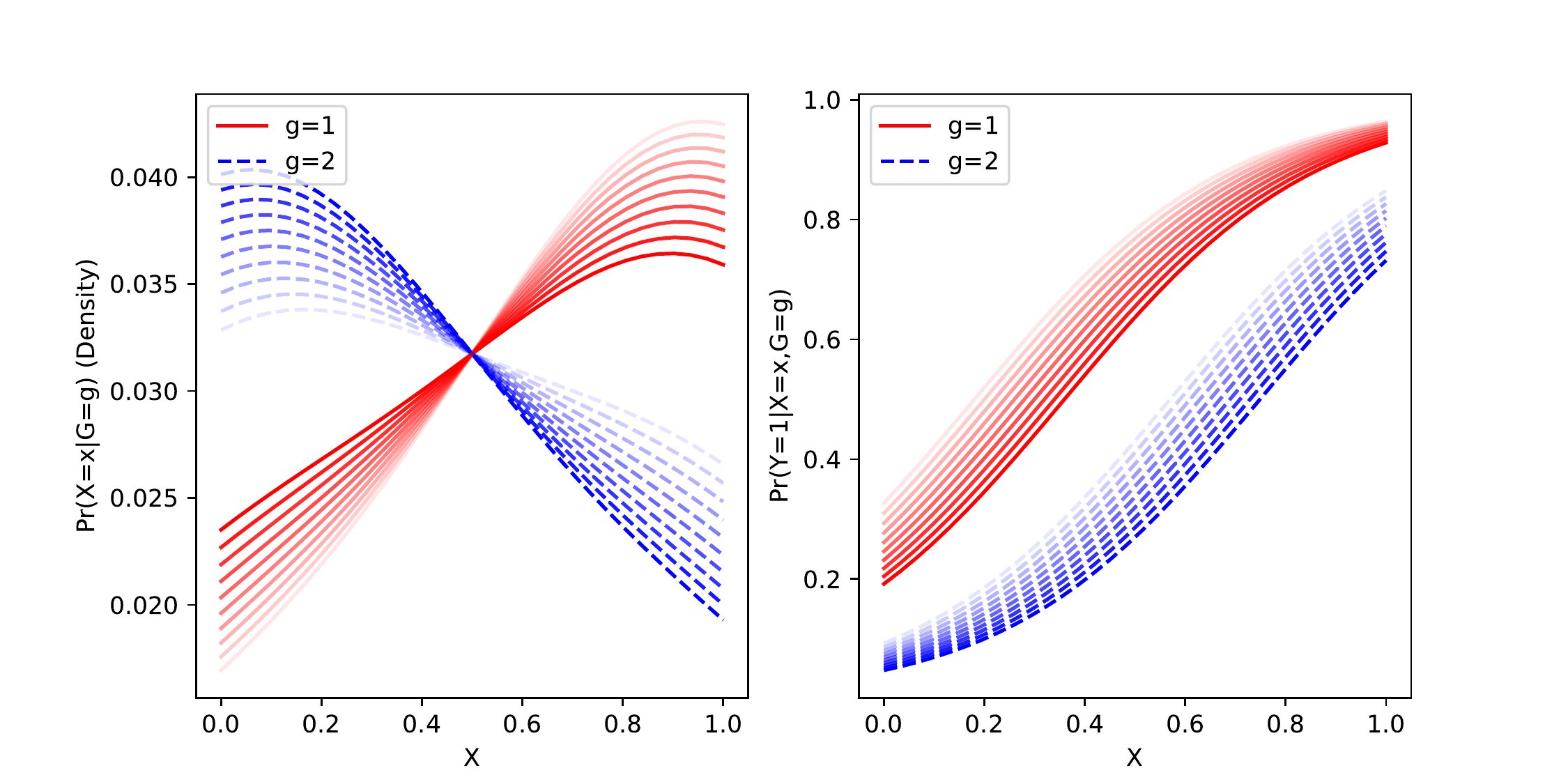}
\caption{A synthetic distribution is updated according to a dynamical kernel \(\PP\) based on evolutionary dynamics (\cref{sec:replicator}), when a classifier repeatedly predicts \(\hat{Y}{=}1\) iff \(X \geq 0.5\). We visualize how the distribution of \(X\) and conditional qualification rates \(\Pr(Y{=}1\mid X)\) change in each group \(g \in \{1 \text{ (red, solid)}, 2 \text{ (blue, dashed)}\}\), fading the plotted lines over 10 time steps. In this example, the feature values \(X\) in each group decrease with time, while the qualification rates of agents at any fixed value of \(X\) decrease.}
\label{fig:ex-dynamics}
\end{figure}

\newpage
\section{Experiment}\label{app:exp}
\subsection{Experiment Details}\label{app:exp_details}
\textbf{Device and Packages.} We run all the experiment on a single 1080Ti GPU. We implement the \drl agent using StableBaseline3\cite{stable-baselines3}. The neural network is implemented using Pytorch\cite{NEURIPS2019_9015}.

\textbf{Neural Network to learn $\phi$.} We use a multi-layer perceptron to learn $\phi$. Specifically, we sample 100000 data points using a random policy, storing $s$, $a$, $r$ and $g$. The inputs of the network are state and action, passing through fully connected (fc) layers with size 256, 128, 64, 64. ReLU is used as activation function between fc layers, while a SoftMax layer is applied after the last fc layer. We treat the outcome of this network as $\phi$. To learn $\phi$, we apply two separated fc layers (without bias) with size 1 to $\hat{\phi}$ and treat the outputs as predicted $r$ and predicted $g$. A combination of MSE losses of $r$ and $g$ are adopted. We use Adam as the optimizer. Weight decay is set to 1e-4 and learning rate is set to 1e-3, while batch size is 128.

Note that, $\hat{\phi}$ is linear regarding $r$ and $g$, but the linearity of transition kernel cannot be captured using such a schema. Therefore, equivalently we made an assumption that there always exists measure $\mu_h$ such that for given $\hat{\phi}$, the linearity of transition kernel holds. It's a stronger assumption than Assumption \ref{asm:linear-mdp}.

\begin{figure}[H]
    \centering
    \begin{subfigure}{\textwidth}
      \centering
      \includegraphics[width=0.26\textwidth]{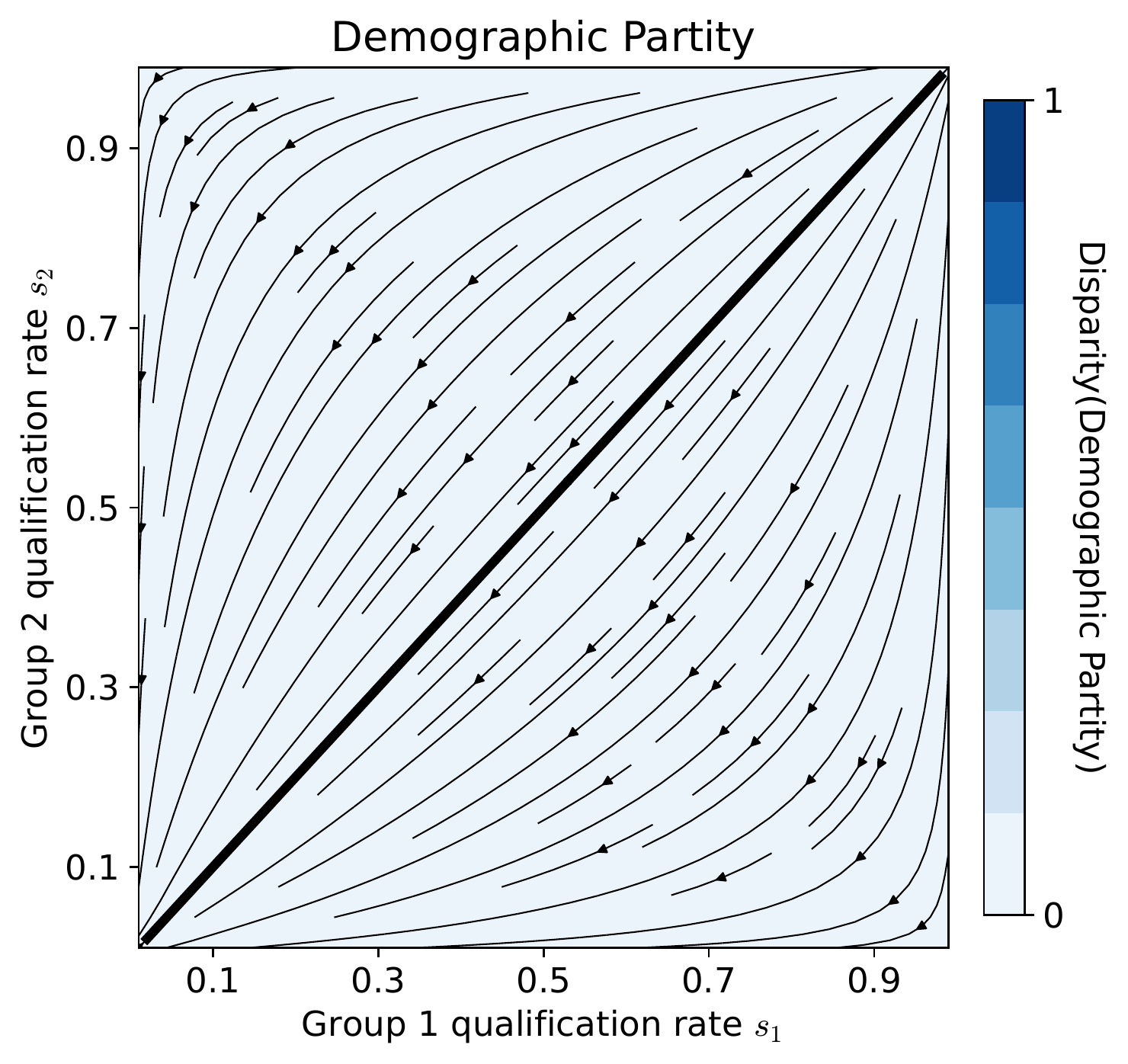}
      \includegraphics[width=0.26\textwidth]{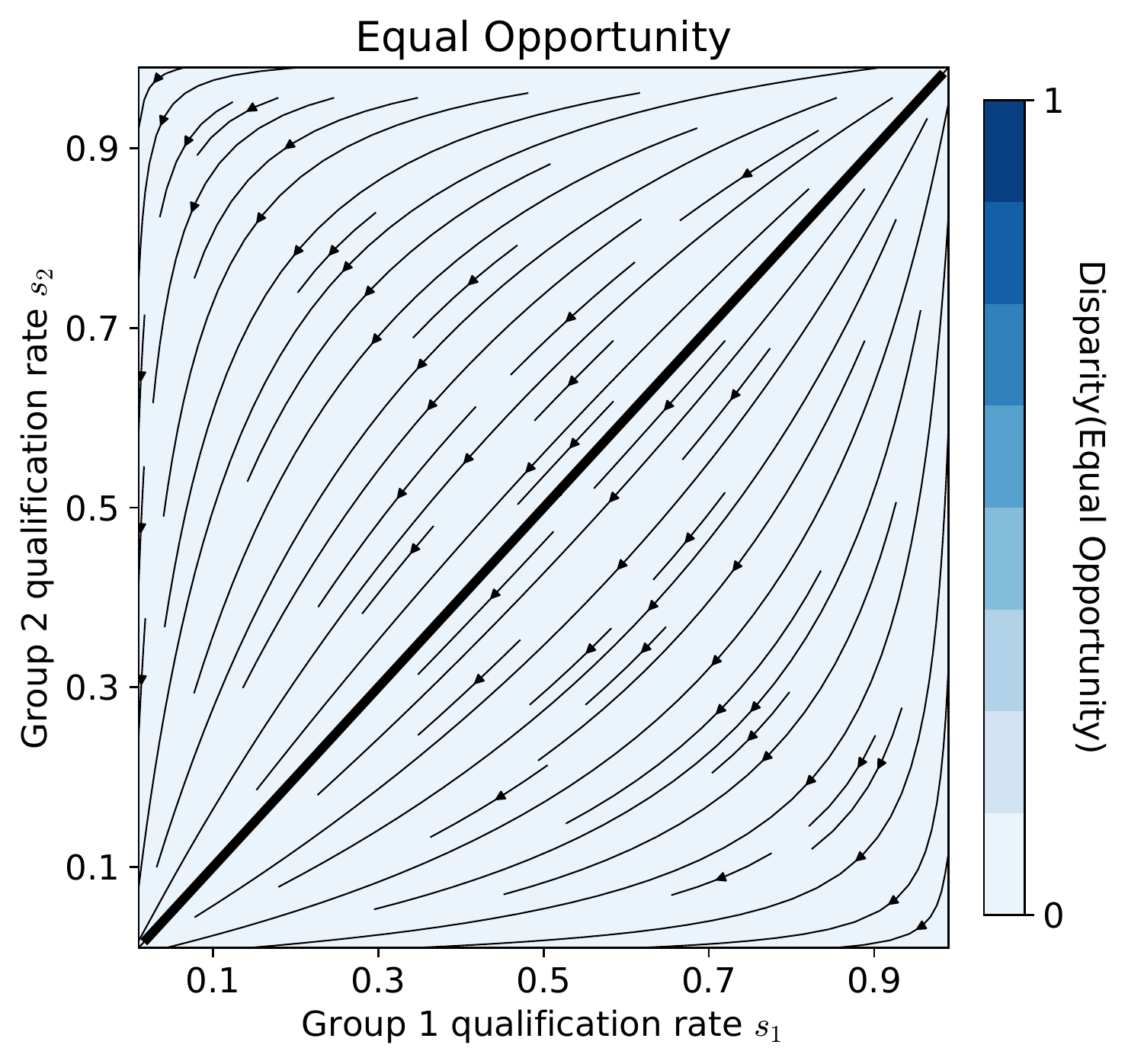}
      \includegraphics[width=0.26\textwidth]{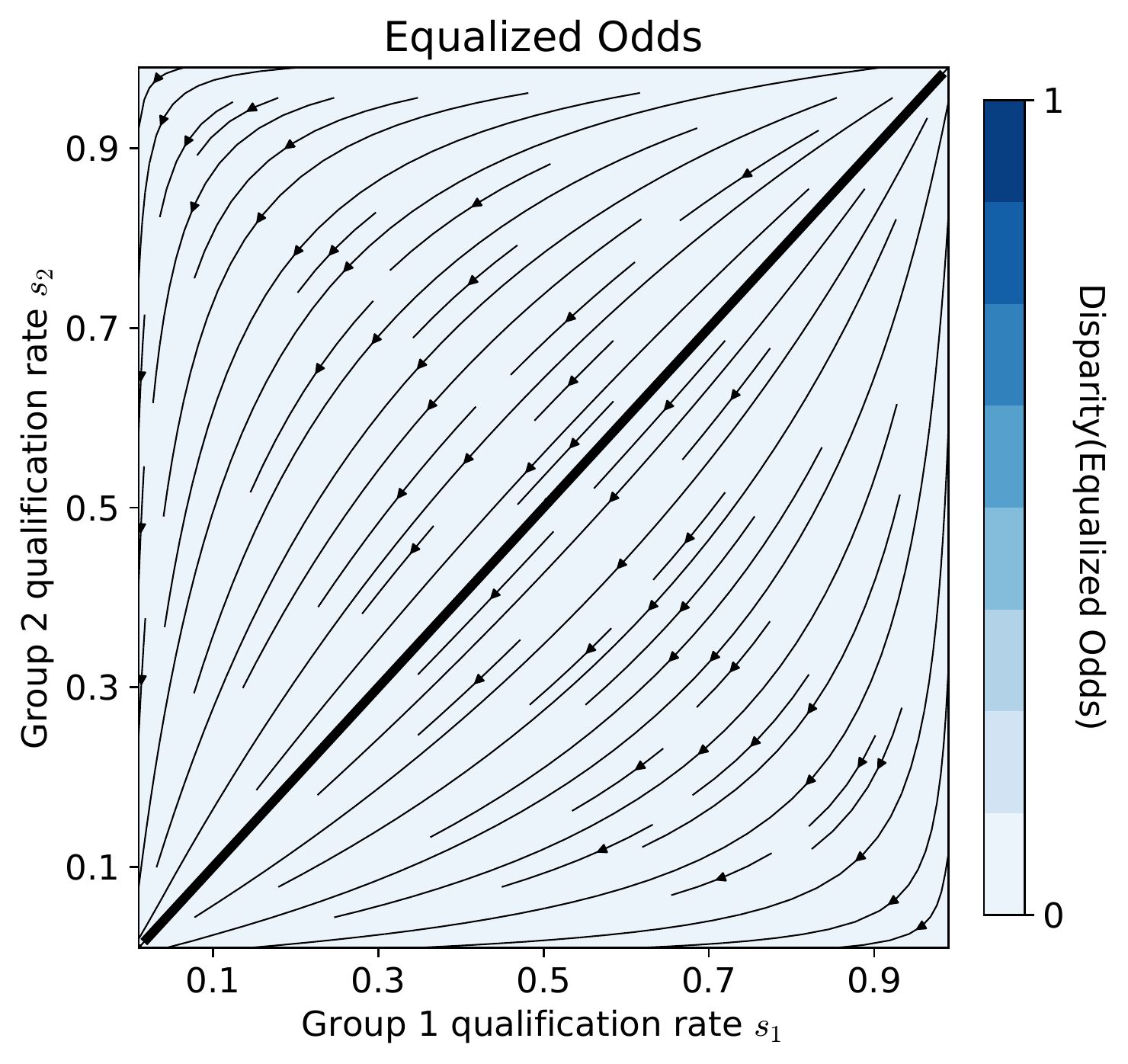}
      \caption{A baseline, greedy classifier performing gradient descent to (locally) maximize true-positives, with static fairness regularization (columns).}
      \label{fig:1a}
    \end{subfigure}
    \begin{subfigure}{\textwidth}
      \centering
      \includegraphics[width=0.26\textwidth]{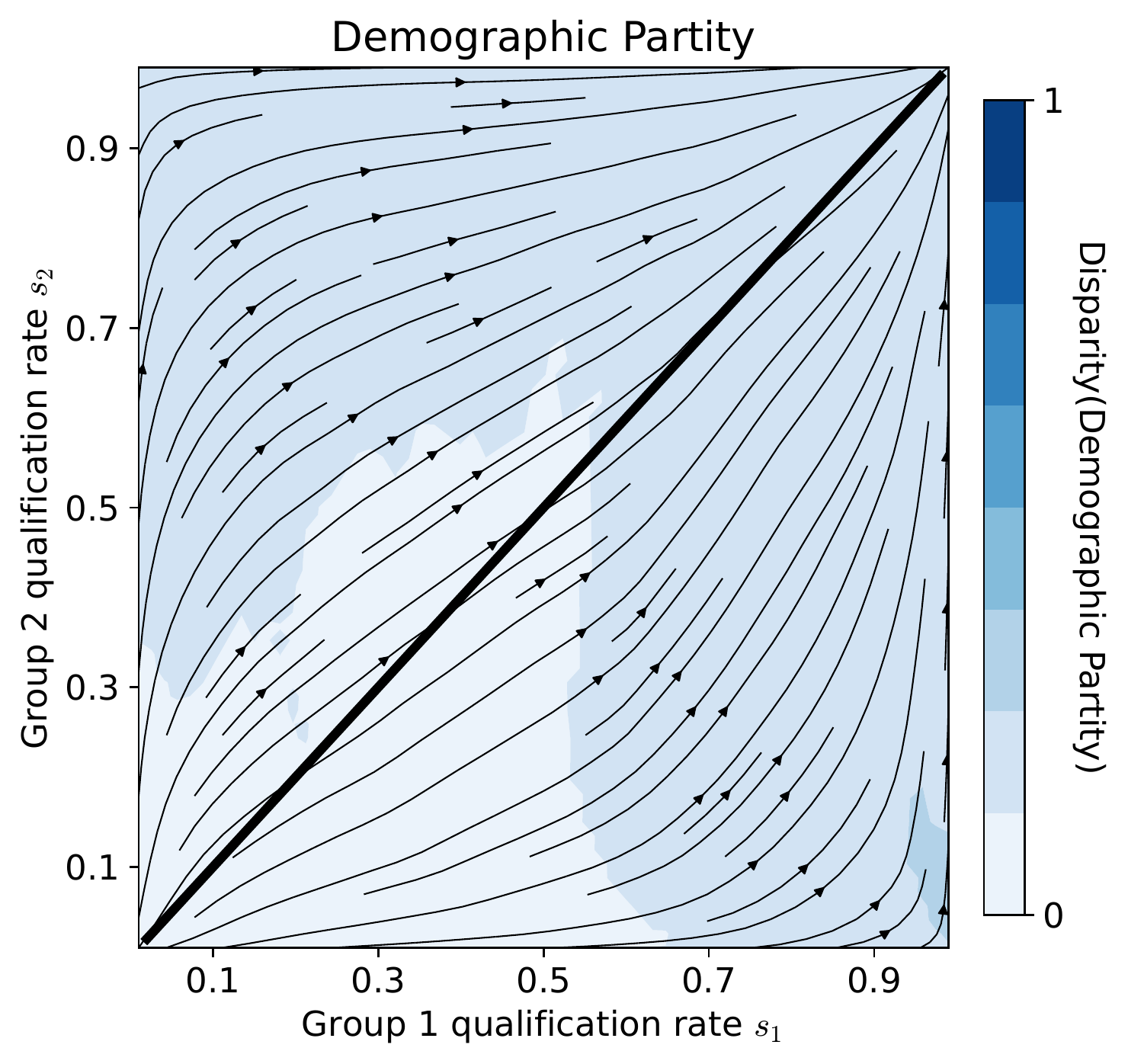}
      \includegraphics[width=0.26\textwidth]{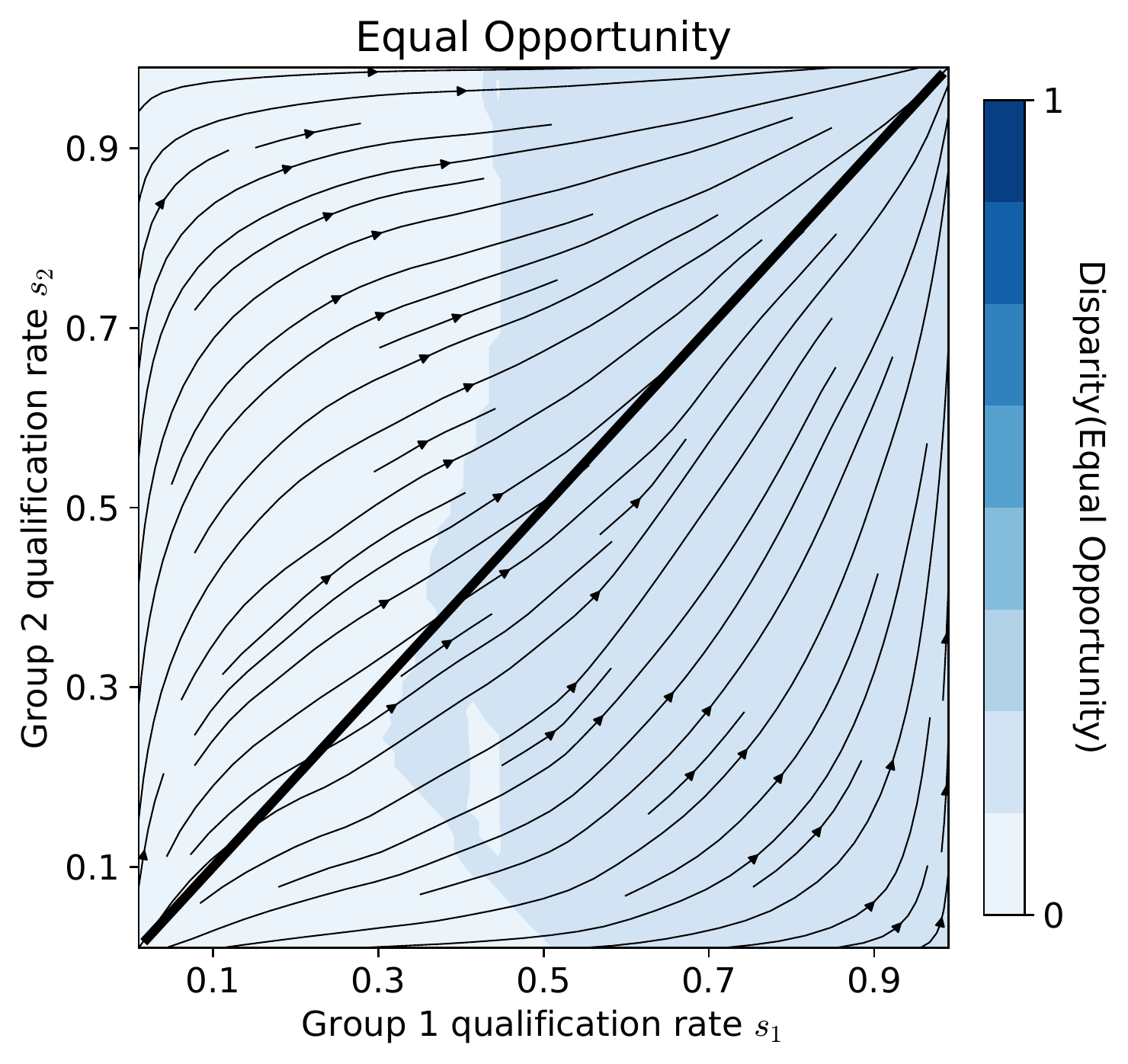}
      \includegraphics[width=0.26\textwidth]{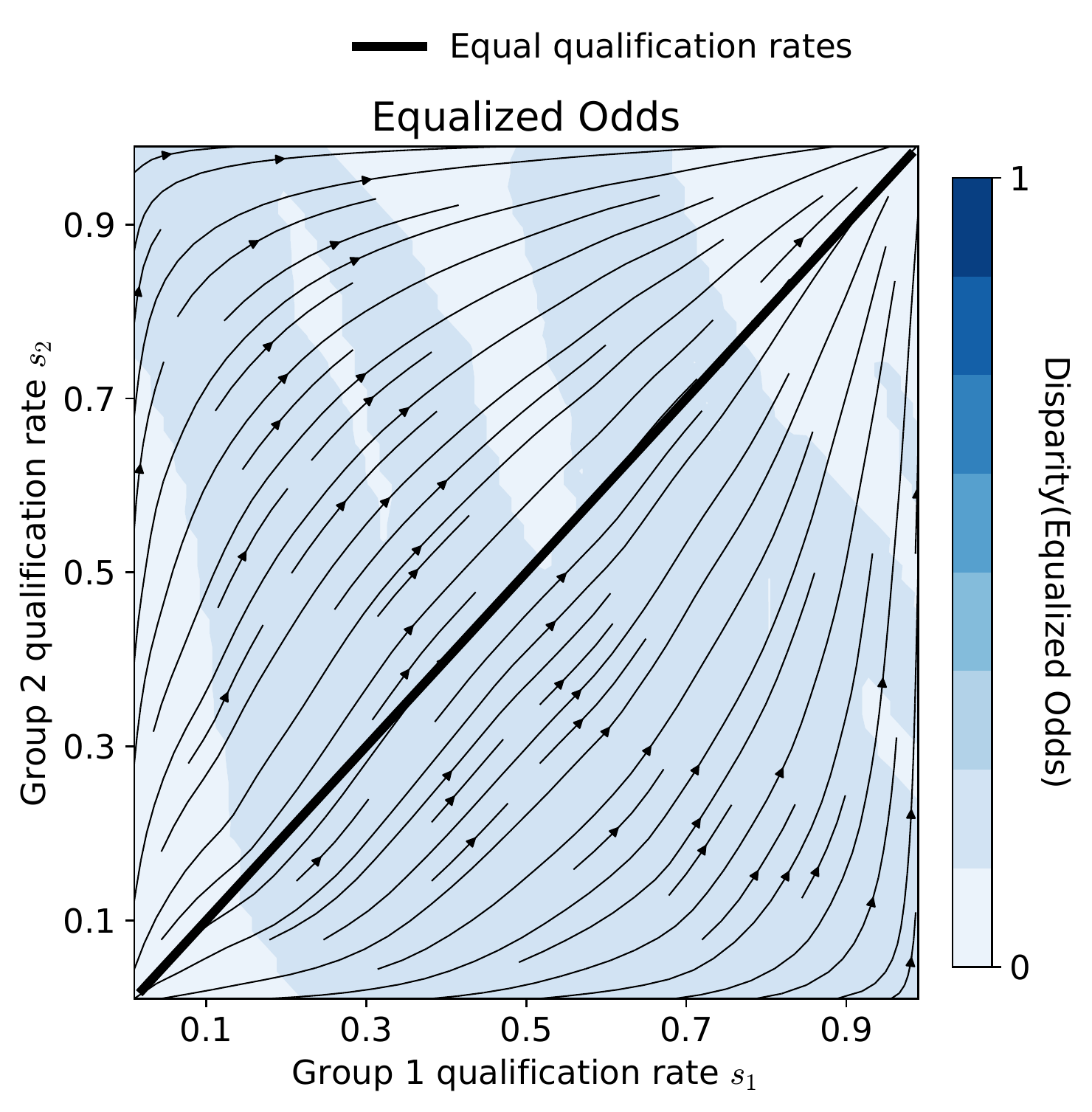}
      \caption{\ucbfair, trained for 2,000 steps on the same, cumulative utility functions.}
      \label{fig:2a}
    \end{subfigure}
    \begin{subfigure}{\textwidth}
      \centering
      \includegraphics[width=0.26\textwidth]{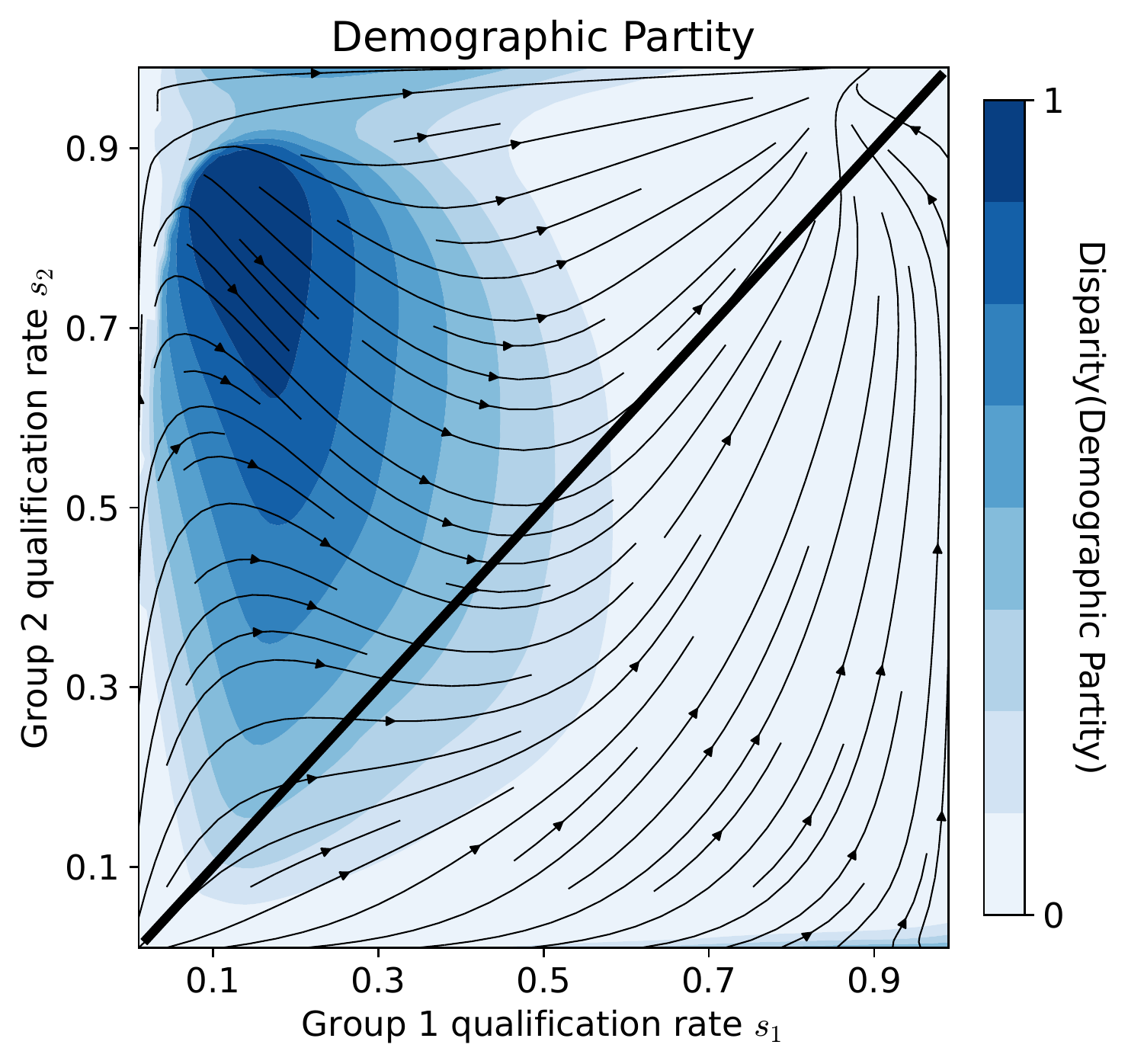}
      \includegraphics[width=0.26\textwidth]{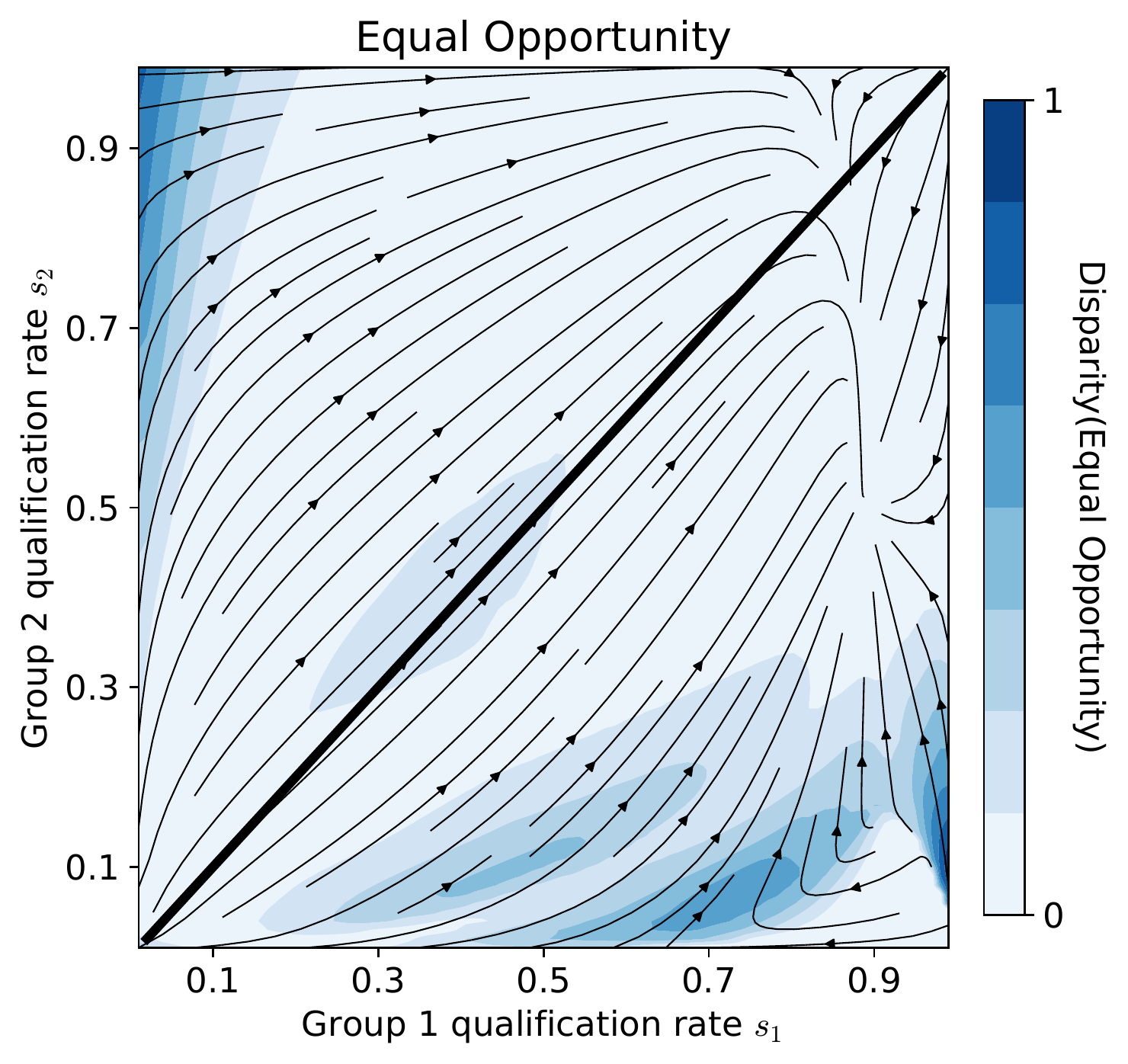}
      \includegraphics[width=0.26\textwidth]{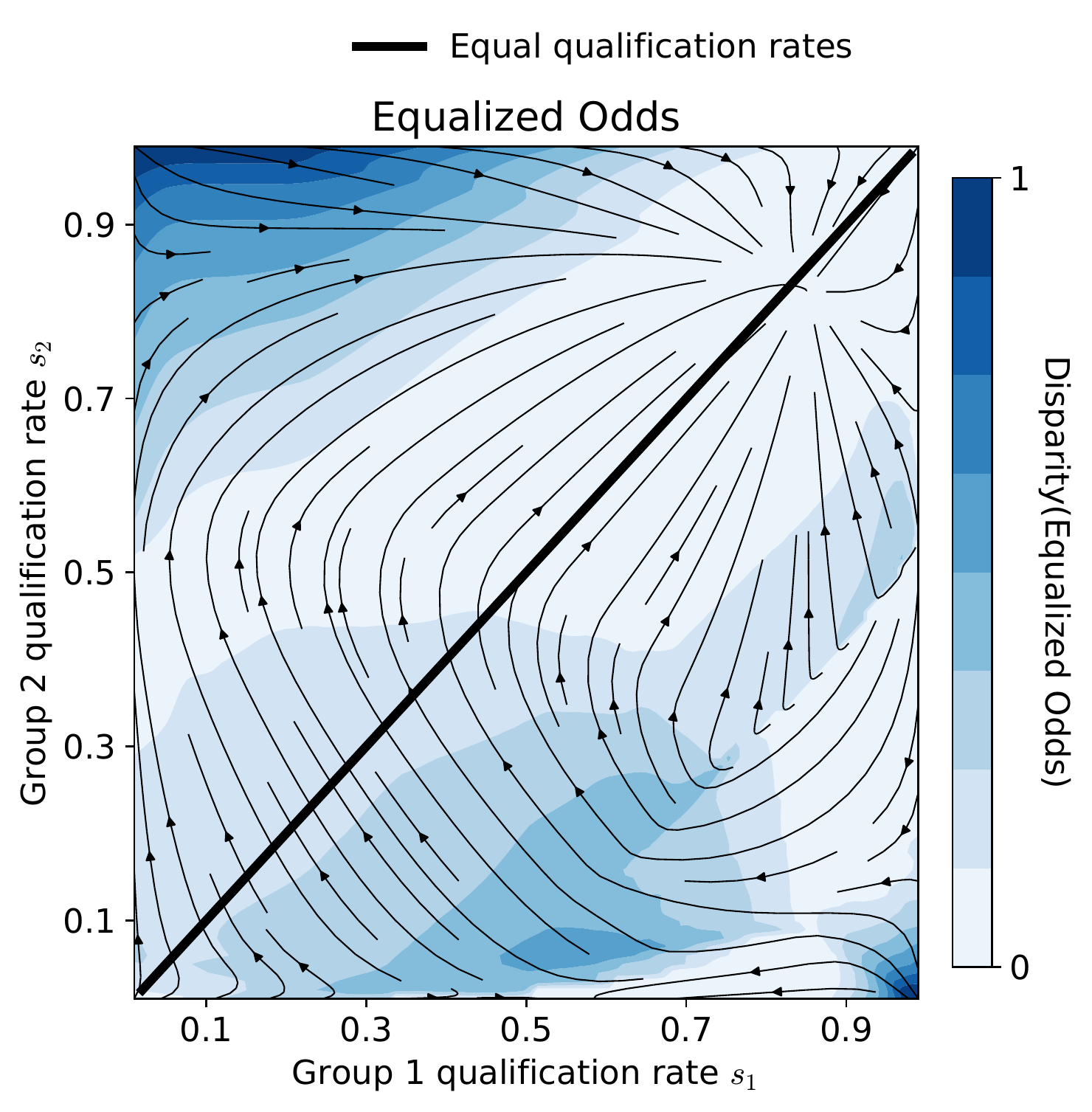}
      \caption{A \drl agent (\cref{sec:drl}) trained for 200,000 steps on the same, cumulative utility functions.}
      \label{fig:1b}
    \end{subfigure}
    \begin{subfigure}{\textwidth}
      \centering
      \includegraphics[width=0.26\textwidth]{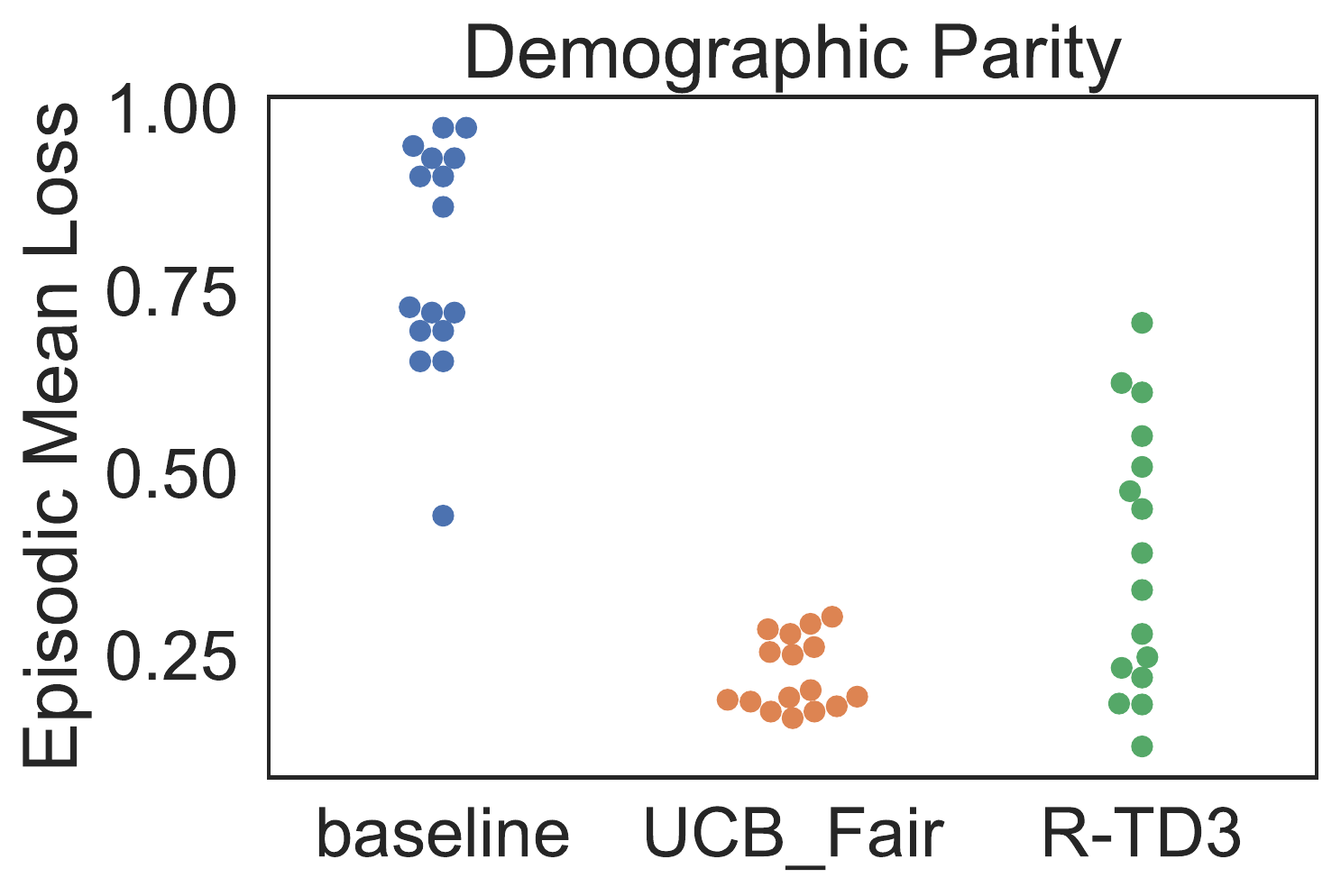}
      \includegraphics[width=0.26\textwidth]{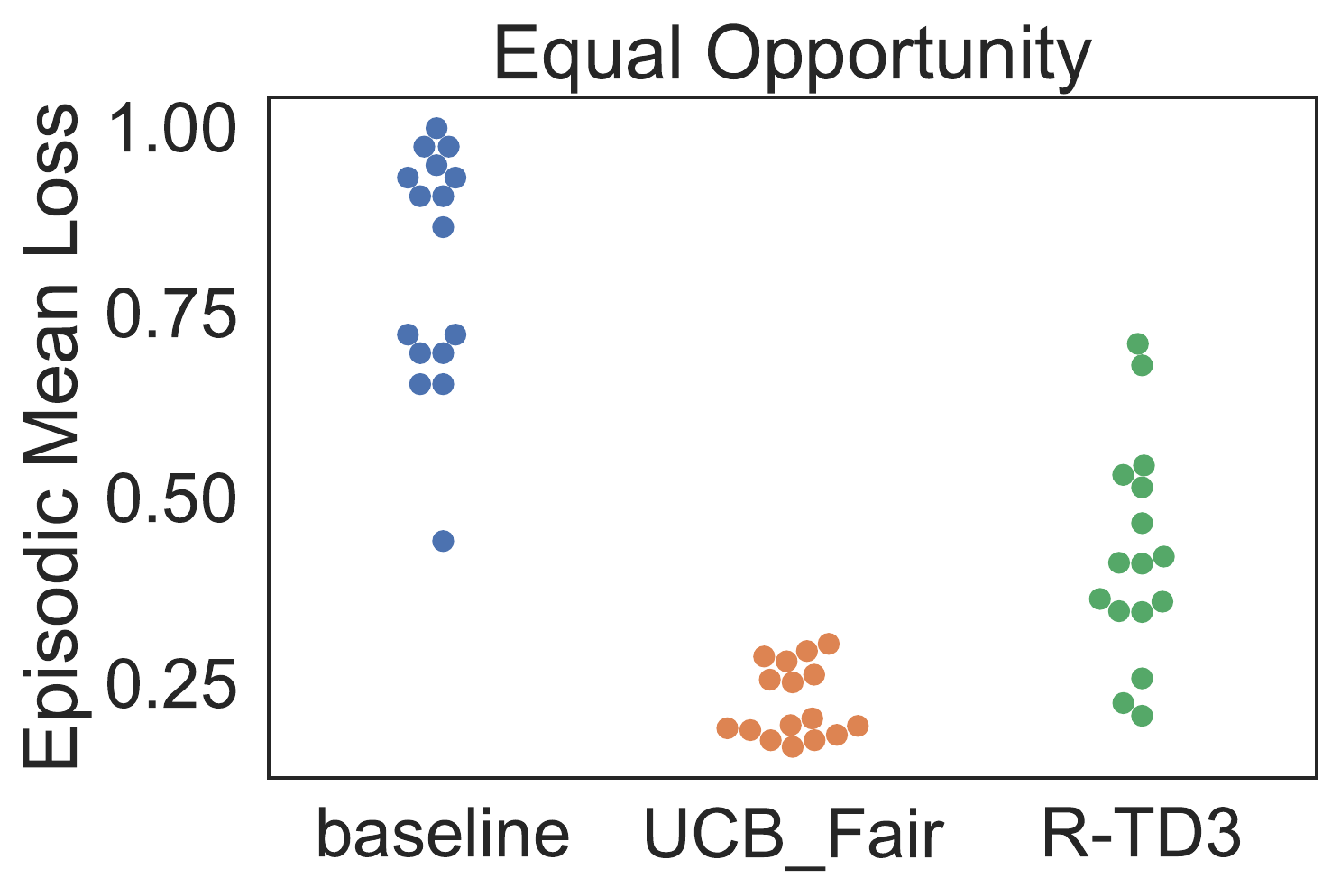}
      \includegraphics[width=0.26\textwidth]{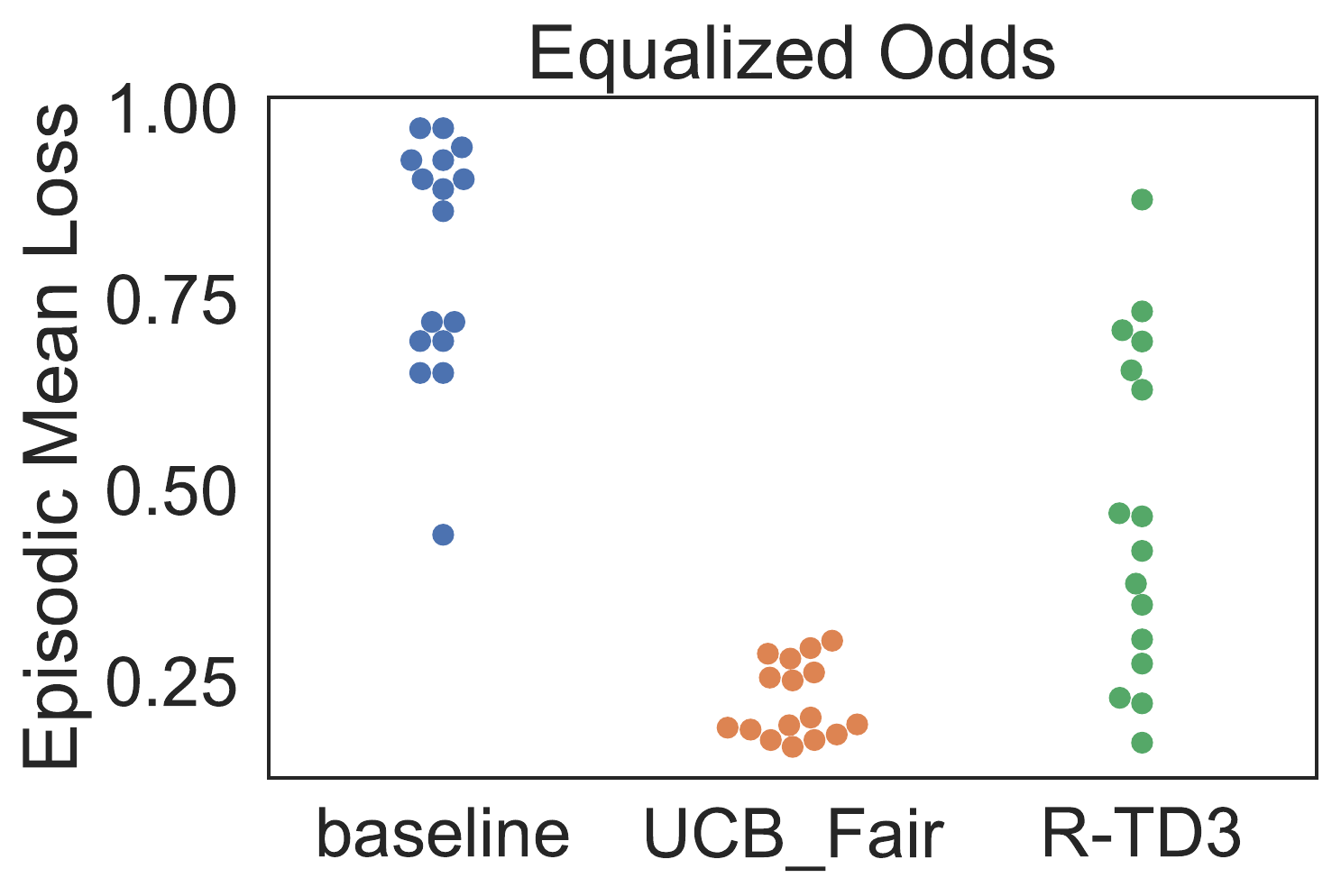}
      \caption{}
      \label{fig:1b}
    \end{subfigure}
    \caption{A comparison of learning policies trained to optimize \textbf{cumulative
true positive fraction} subject to three different regularized fairness
constraints (columns) with \(\nu=1\) (\ucbfair), \(\lambda=0.5\), (greedy agent), and the time-dependent regularization detailed in \cref{sec:drl} (\drl). The first policy (top row) is a
baseline, myopic policy that greedily seeks to optimize current utility in any
state by performing gradient decent. The second policy (bottom row) is trained
using deep reinforcement learning (\drl) as detailed in \cref{sec:drl} for
200,000 steps before we terminate learning and generate the phase portraits
depicted. This is on the synthetic distribution.
In all cases, the baseline, greedy policy drives the
system to promote unqualified individuals, with low qualification rates in each
group, while the \drl agent is able to drive the system to more favorable
equilibria characterized by higher qualification rates. The shading in the phase plots
depicts the violation of the regularizing fairness constraint within each
column, validating the claim that the \drl agent learns to sacrifice short-term
utility to drive towards preferable system states.
}
    \label{fig:1}
\end{figure}

\begin{figure}[H]
    \centering
    \begin{subfigure}{\textwidth}
      \centering
      \includegraphics[width=0.26\textwidth]{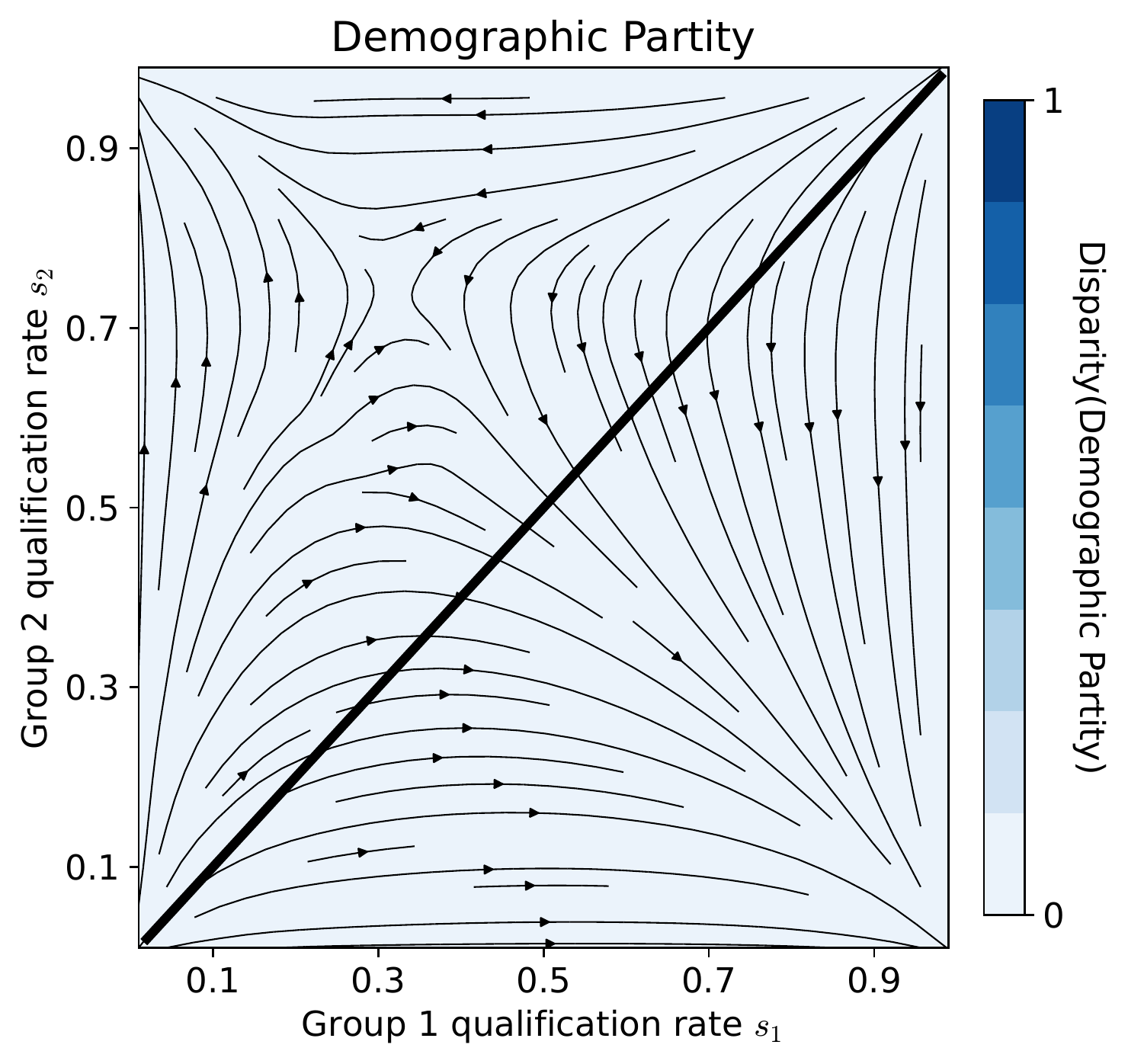}
      \includegraphics[width=0.26\textwidth]{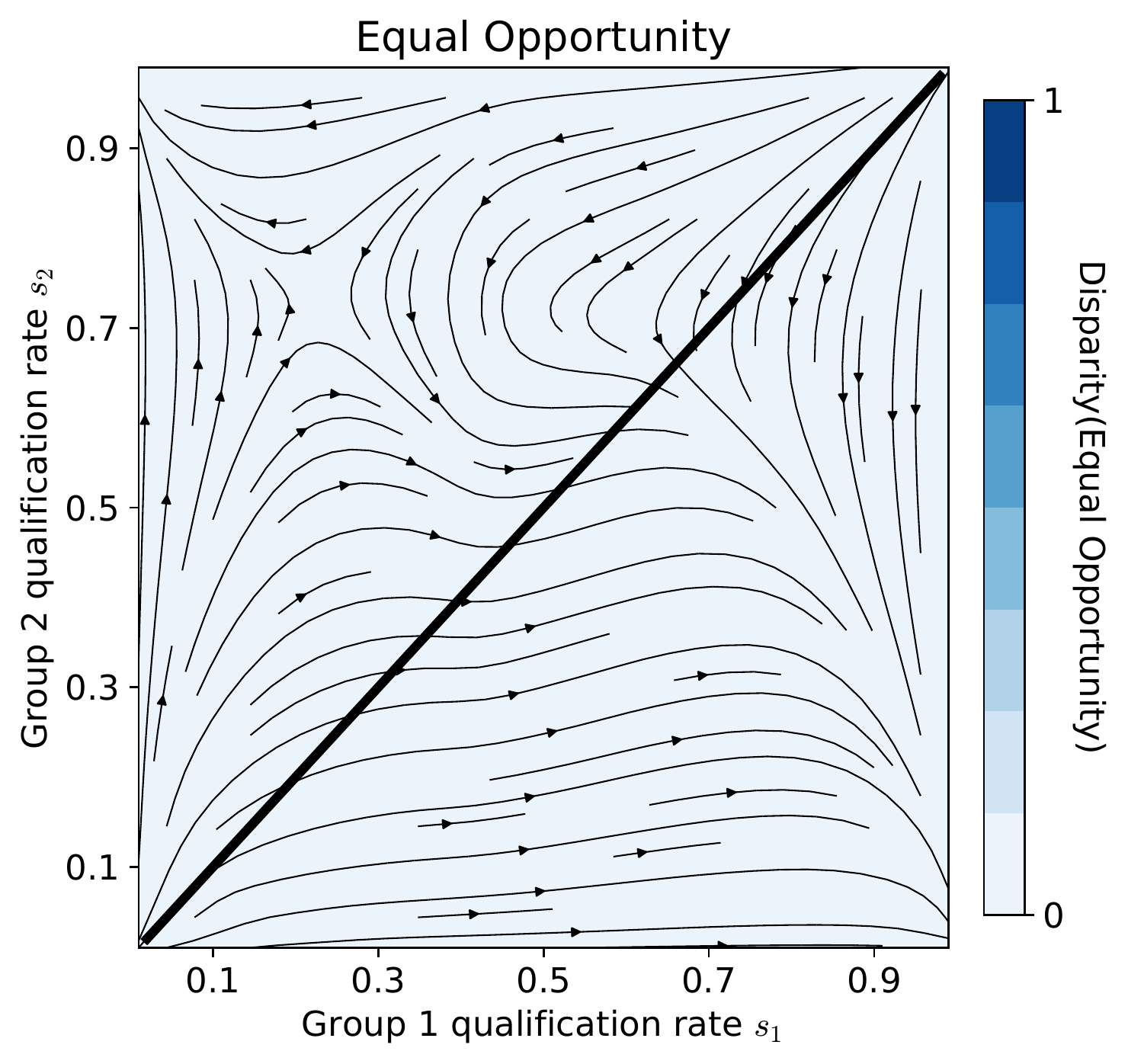}
      \includegraphics[width=0.26\textwidth]{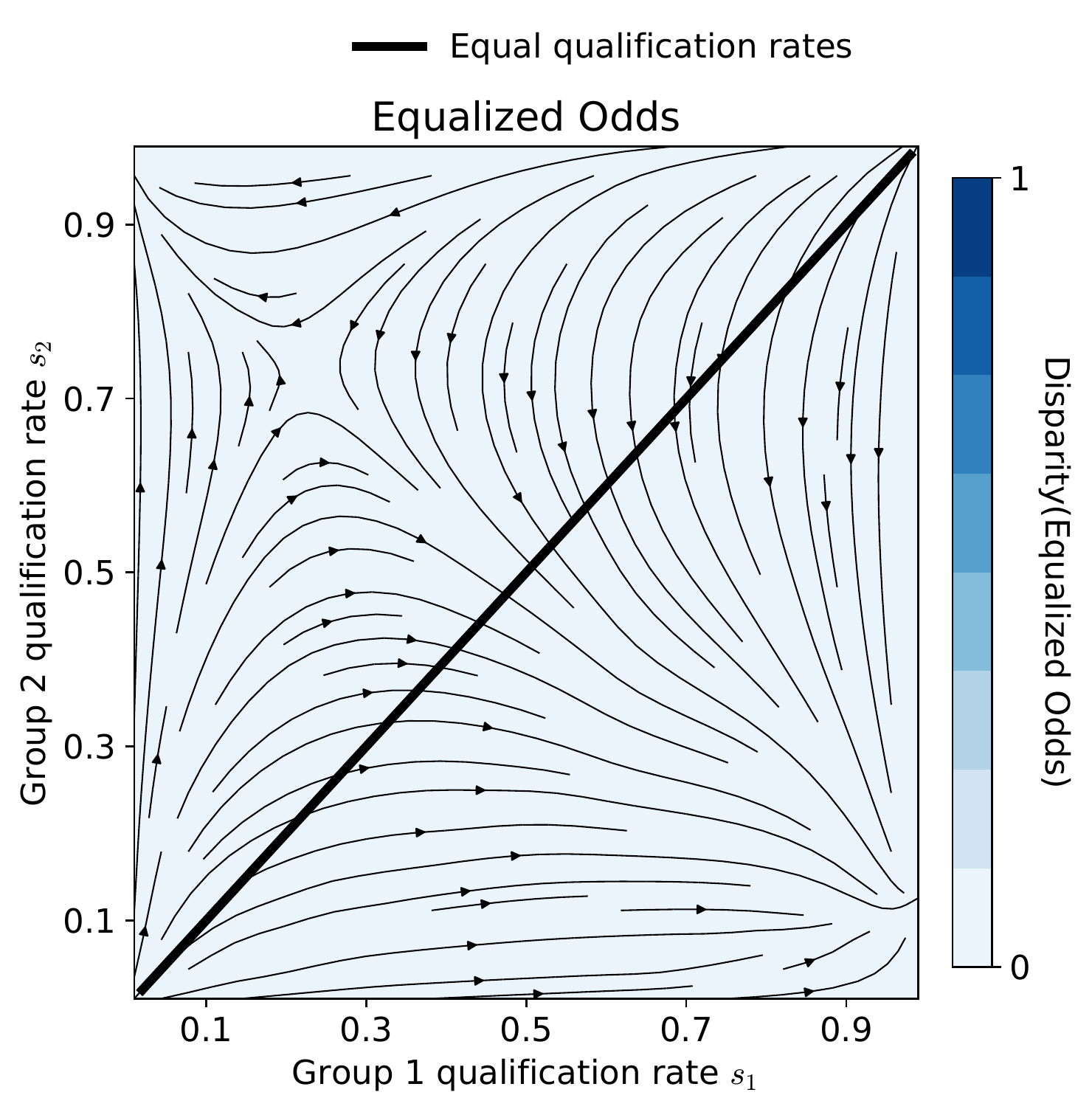}
      \caption{A baseline, greedy classifier locally maximizing true positive classifications, regularized by fairness (columns).}
      \label{fig:2a}
    \end{subfigure}
    \begin{subfigure}{\textwidth}
      \centering
      \includegraphics[width=0.26\textwidth]{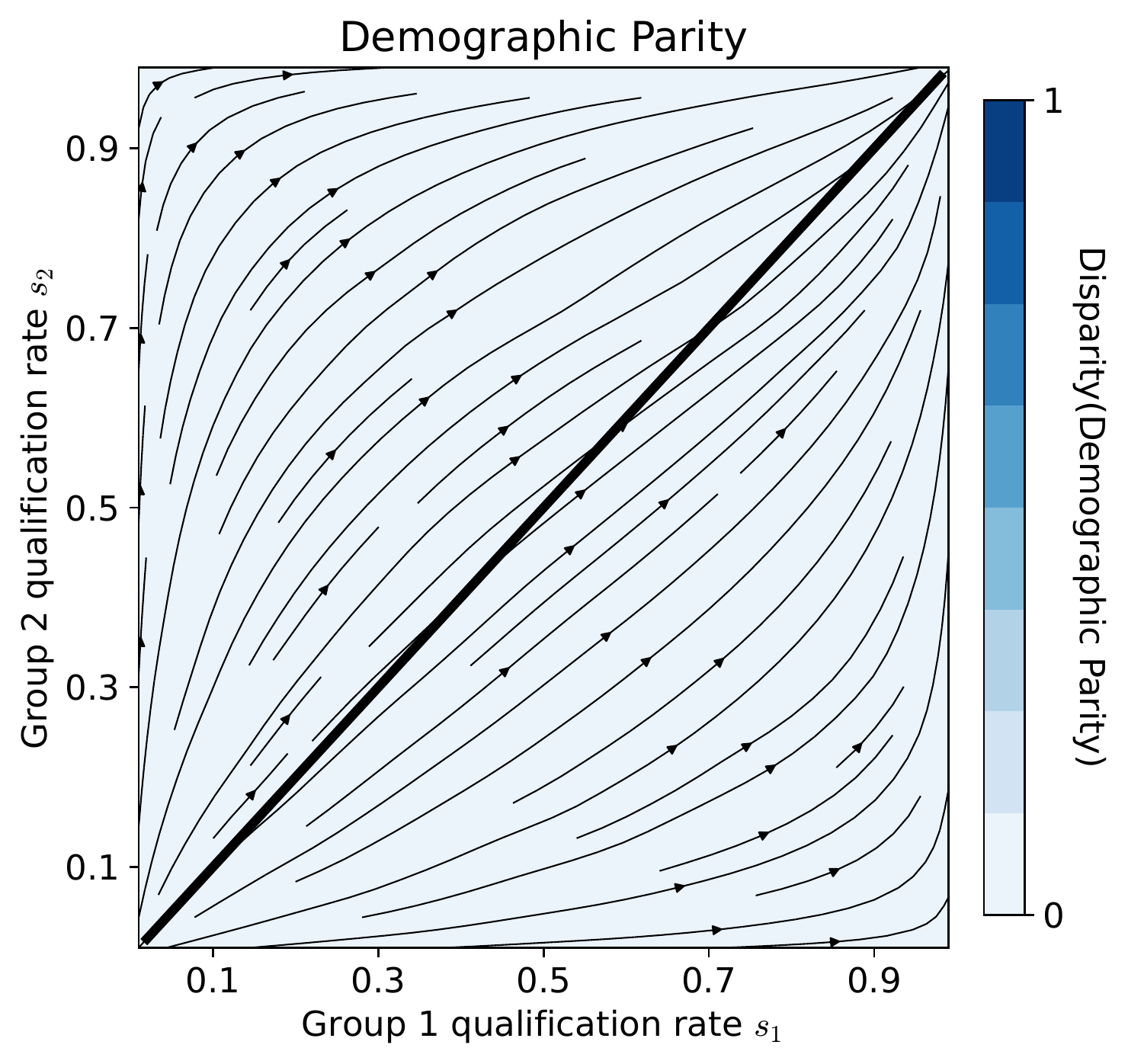}
      \includegraphics[width=0.26\textwidth]{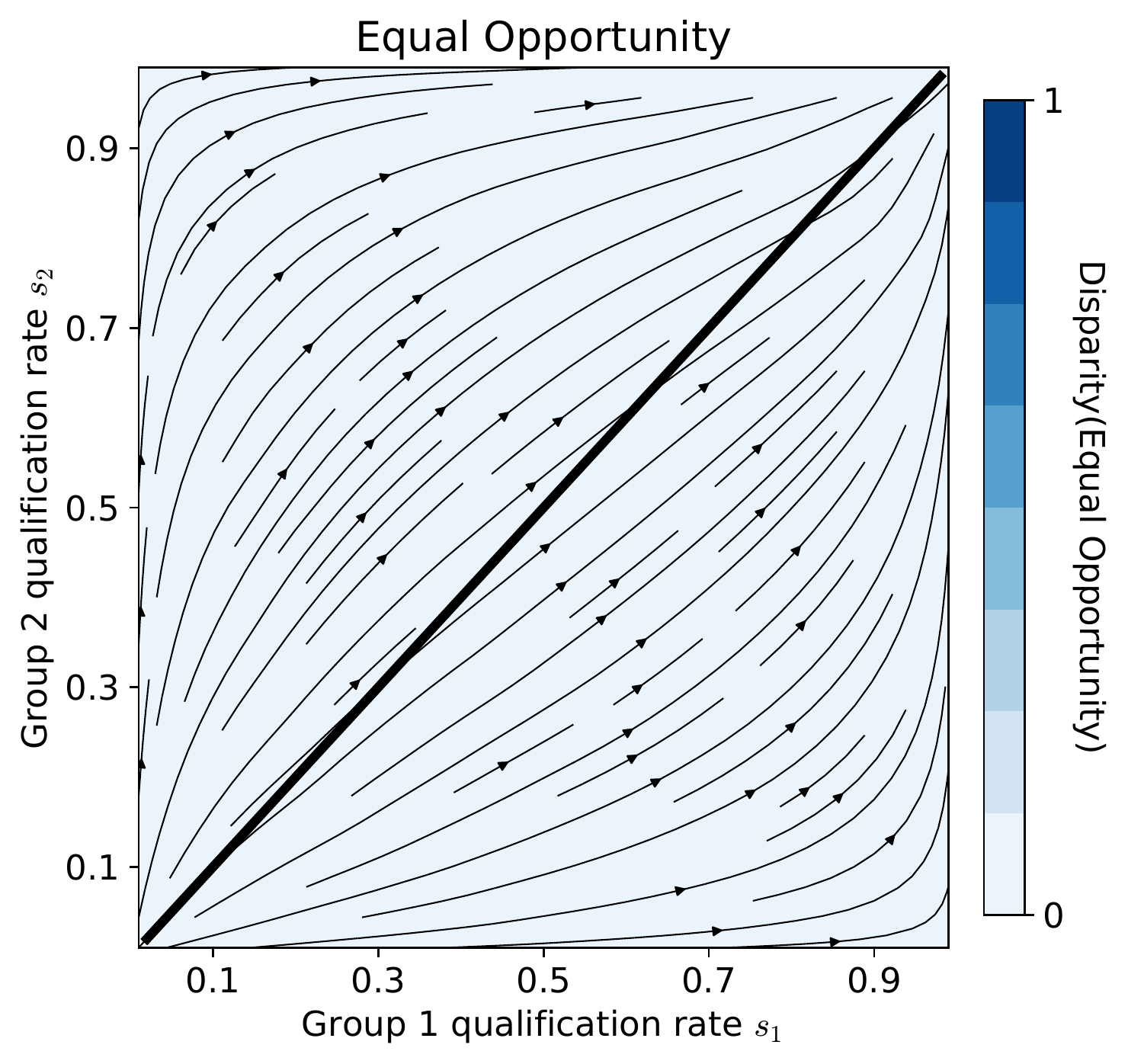}
      \includegraphics[width=0.26\textwidth]{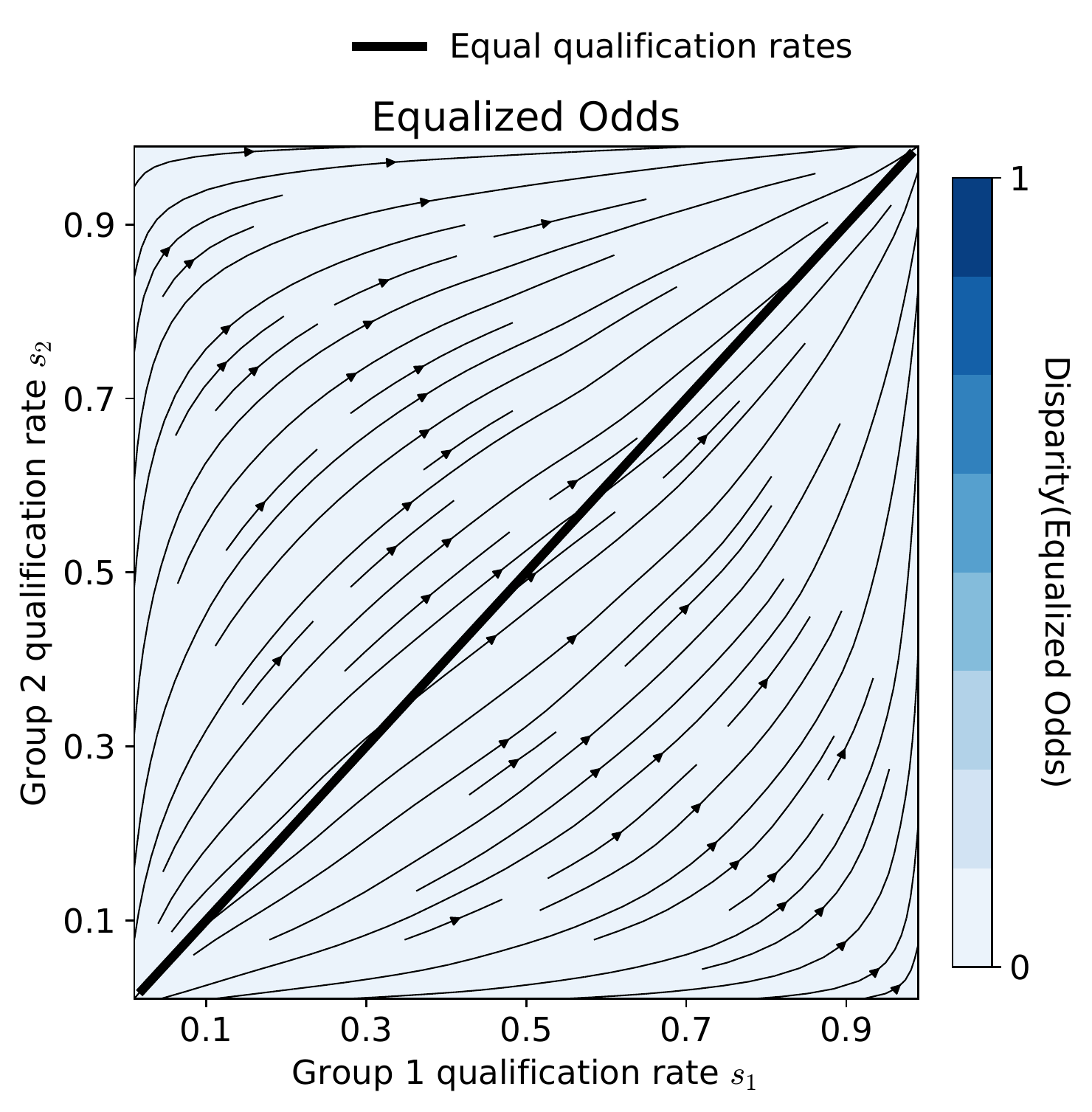}
      \caption{\ucbfair, trained for 2,000 steps on the same, cumulative utility functions.}
      \label{fig:2a}
    \end{subfigure}
    \begin{subfigure}{\textwidth}
      \centering
      \includegraphics[width=0.26\textwidth]{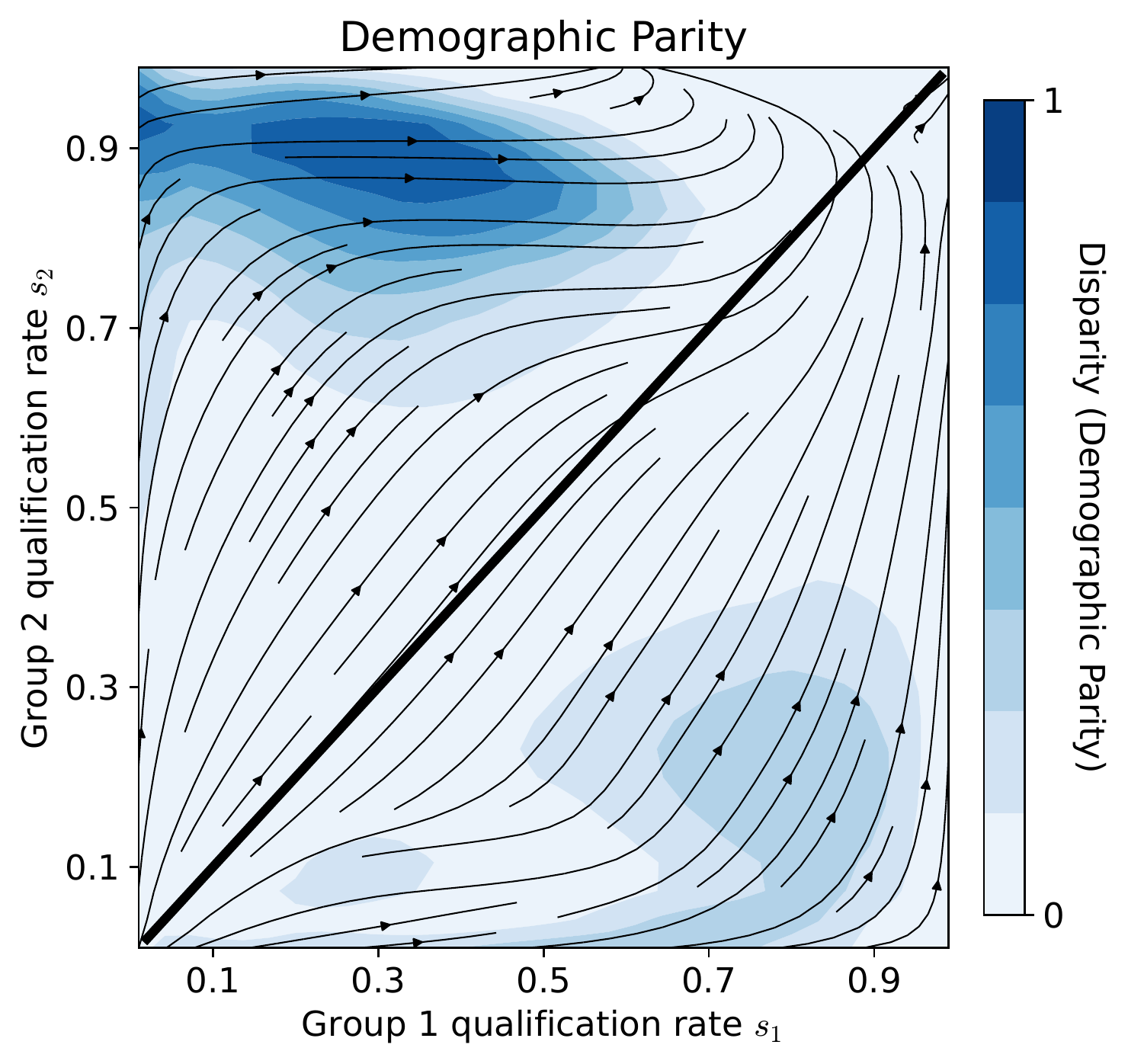}
      \includegraphics[width=0.26\textwidth]{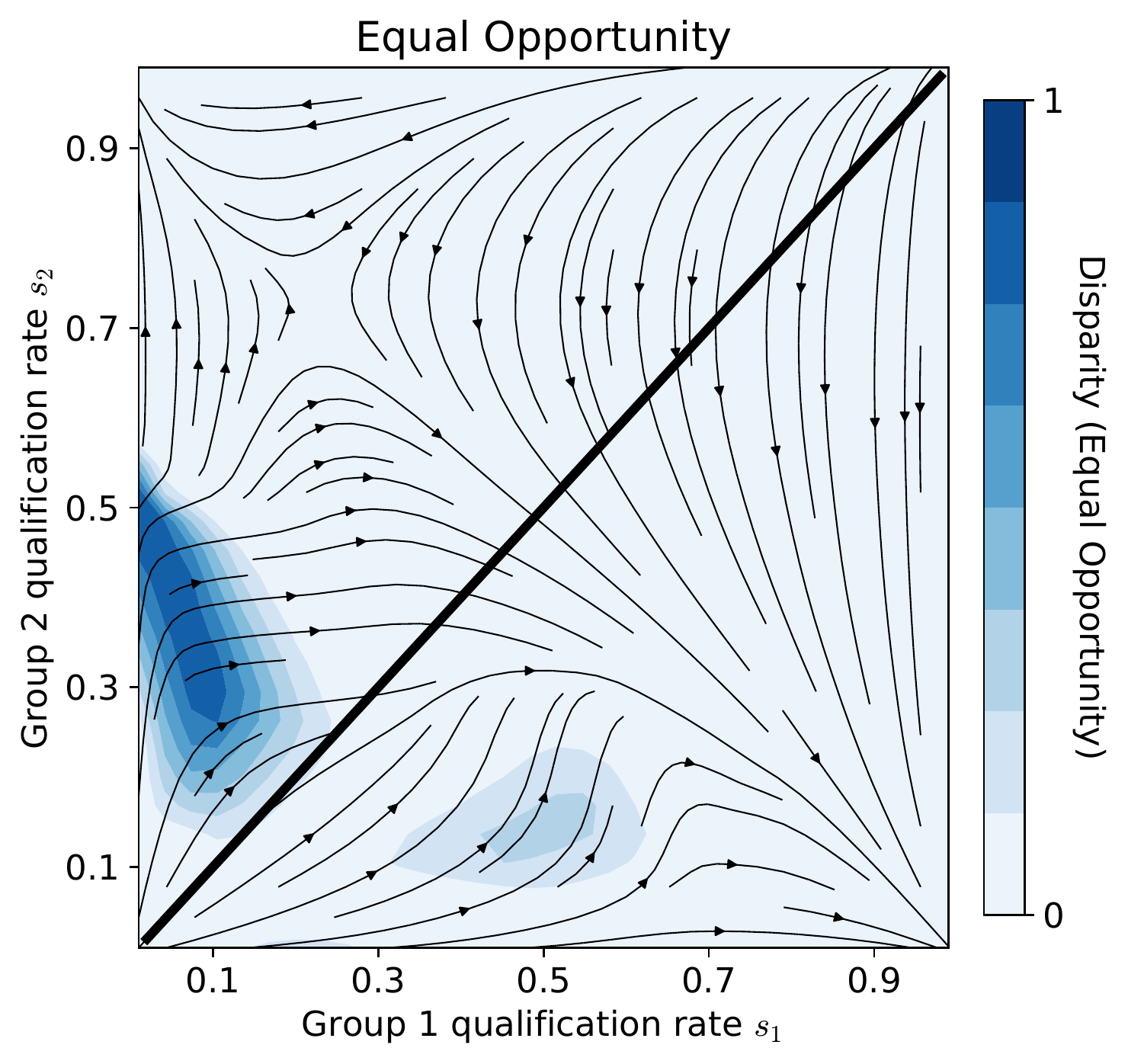}
      \includegraphics[width=0.26\textwidth]{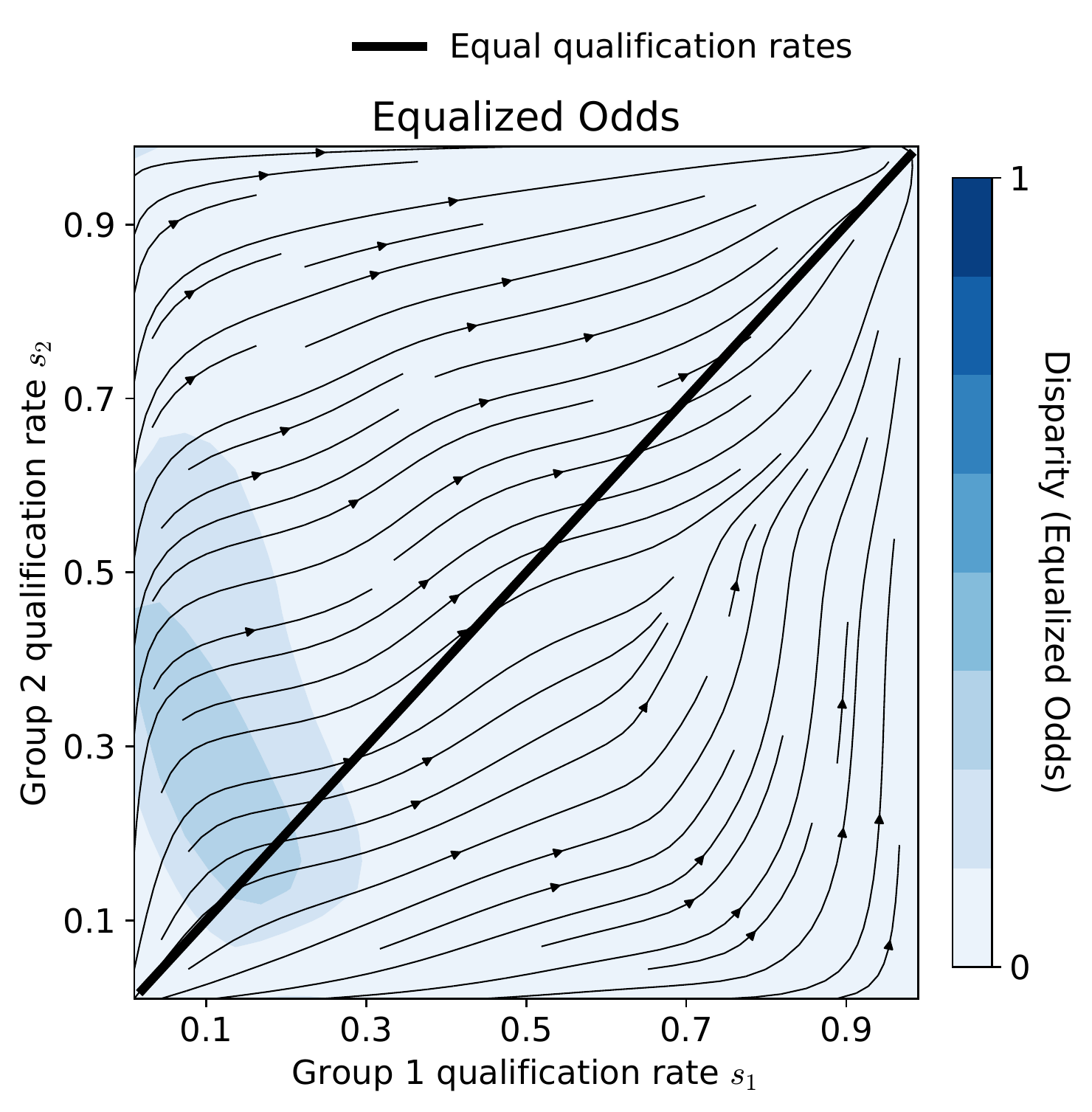}
      \caption{A \drl agent (\cref{sec:drl}) trained for 200,000 steps on the same, cumulative utility functions.}
      \label{fig:2b}
    \end{subfigure}
    \begin{subfigure}{\textwidth}
      \centering
      \includegraphics[width=0.26\textwidth]{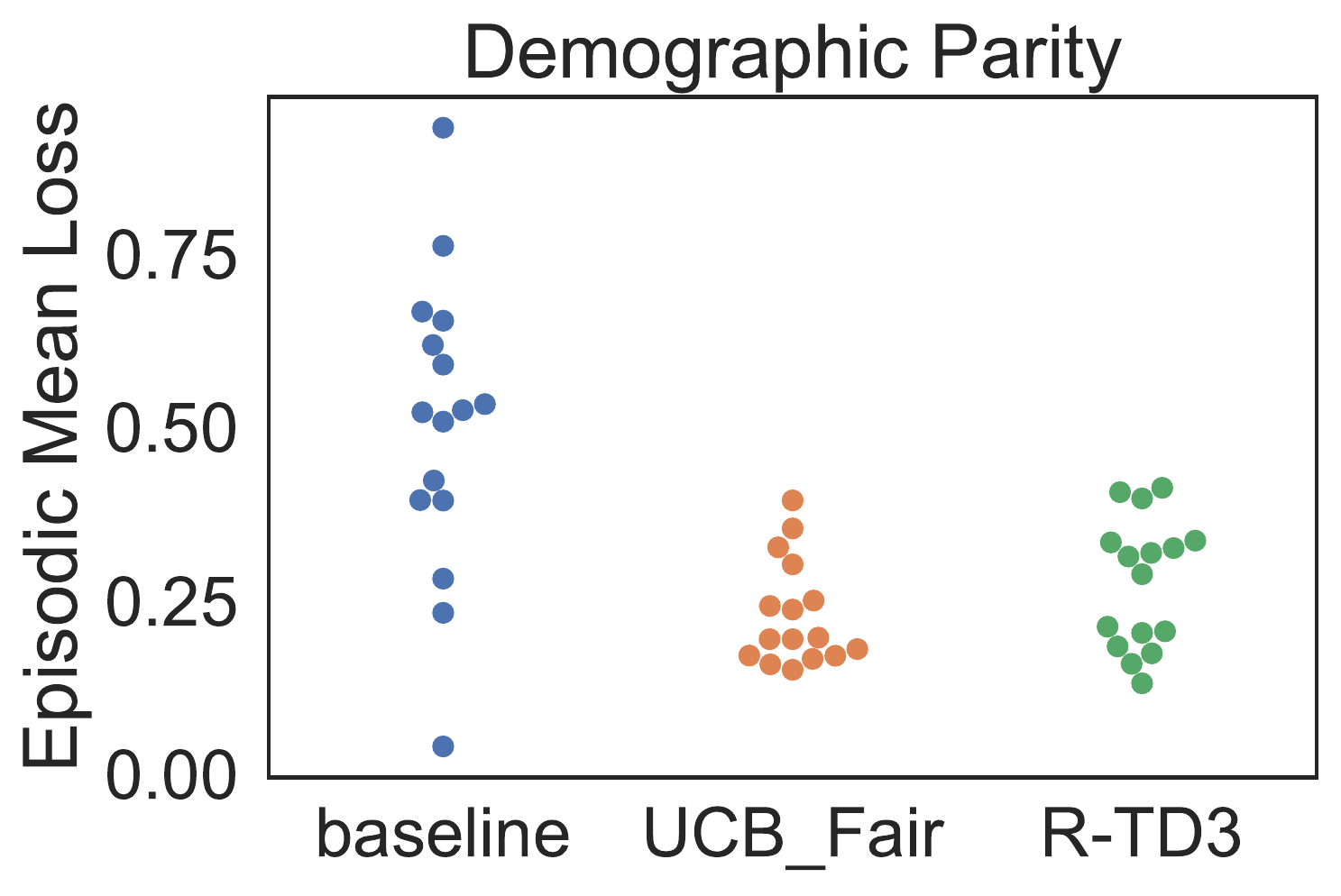}
      \includegraphics[width=0.26\textwidth]{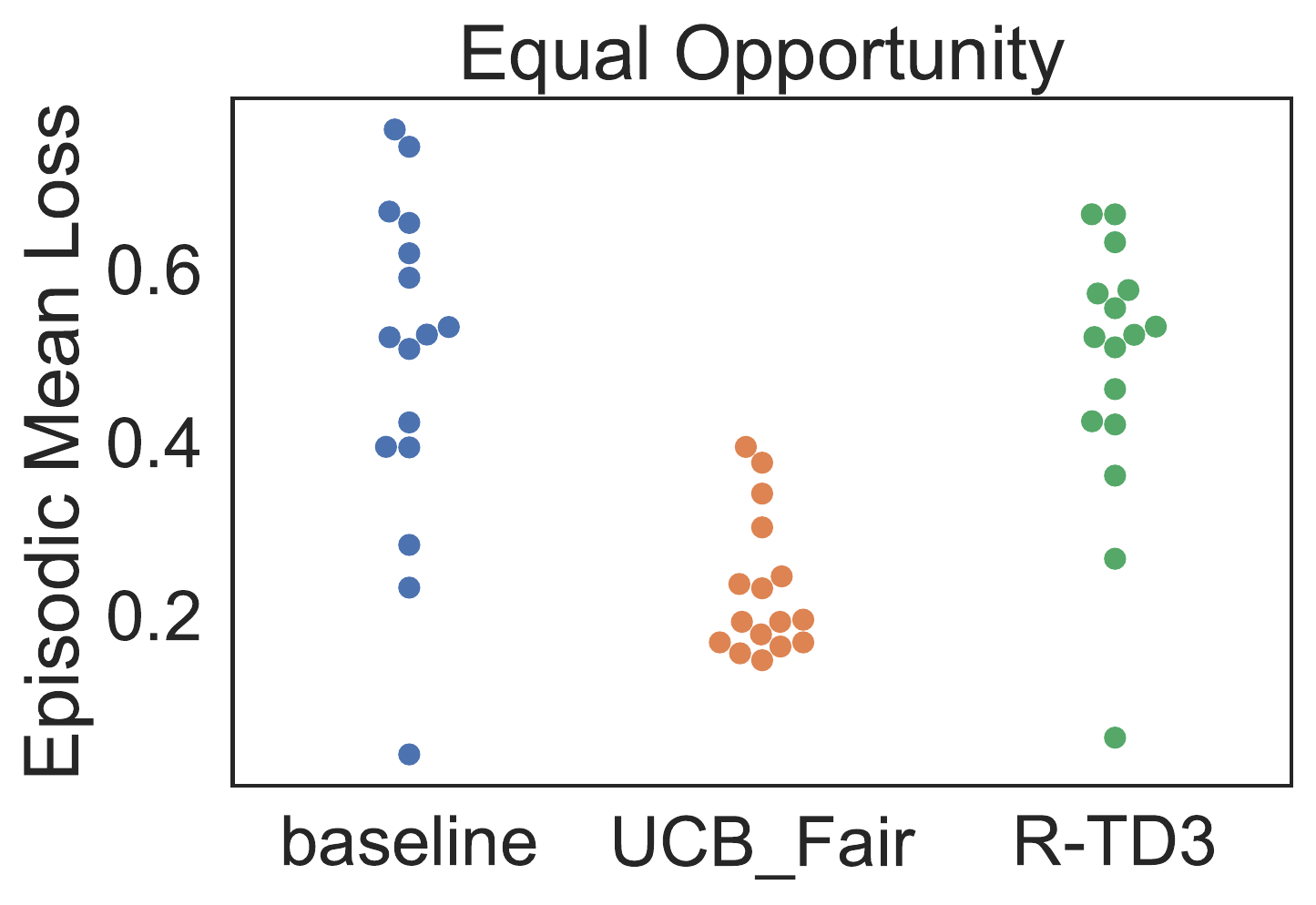}
      \includegraphics[width=0.26\textwidth]{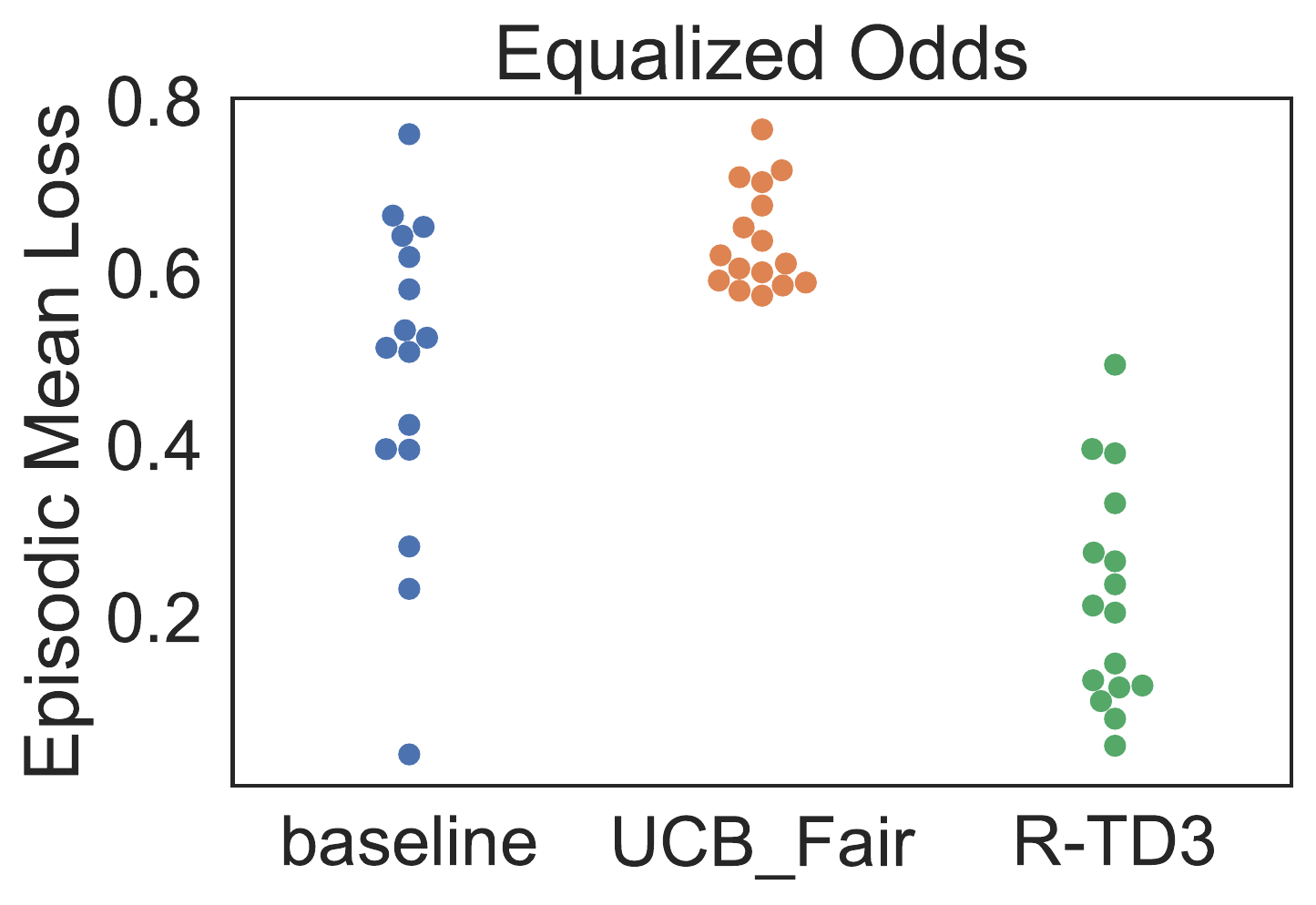}
    \end{subfigure}
    \caption{A repetition of the experiment performed in \cref{fig:1}, rewarding \textbf{true positive fraction} using data
synthesized from the \textbf{UCI ``Adult Data Set''}, as detailed in section
\cref{sec:synthesis} with equal group size reweighting. For this experiment, an individual's sex defined their
``group'' membership, which is an imbalanced label in the dataset
(\(\approx 67\)\% male, group 2, vertical axis) that we re-weight for equal
representation \cref{sec:numerical}. The stark difference between \cref{fig:1}
and this experiment in the qualitative behavior of the greedy agent can be
largely explained by the fact that \(\Pr(Y{=}1 | X{=}x))\) is not actually
monotonically increasing in \(x\), as stipulated by \cref{asm:well-behaved}.
Indeed, if \(Pr(Y{=}1 | X{=}x)\) is sufficiently rough, the threshold selected
by the baseline agent is liable to appear as if sampled uniformly at random,
which is how the initial threshold value is chosen for each of the 20 iterations
averaged over for each pair of group qualification rates used to generate the
phase portraits above. Despite this failure mode of the baseline agent, however,
the \drl agent is still largely able to drive the system towards equilibria with
more equal qualification rates in both groups. The line of equal qualification
rates in both groups is depicted in black, from the lower-left corner of each
phase plot to the upper-right.}
    \label{fig:2}
\end{figure}

\subsection{zero one accuracy with more weights on true positive}

\begin{figure}[H]
    \centering
    \begin{subfigure}{\textwidth}
      \centering
      \includegraphics[width=0.26\textwidth]{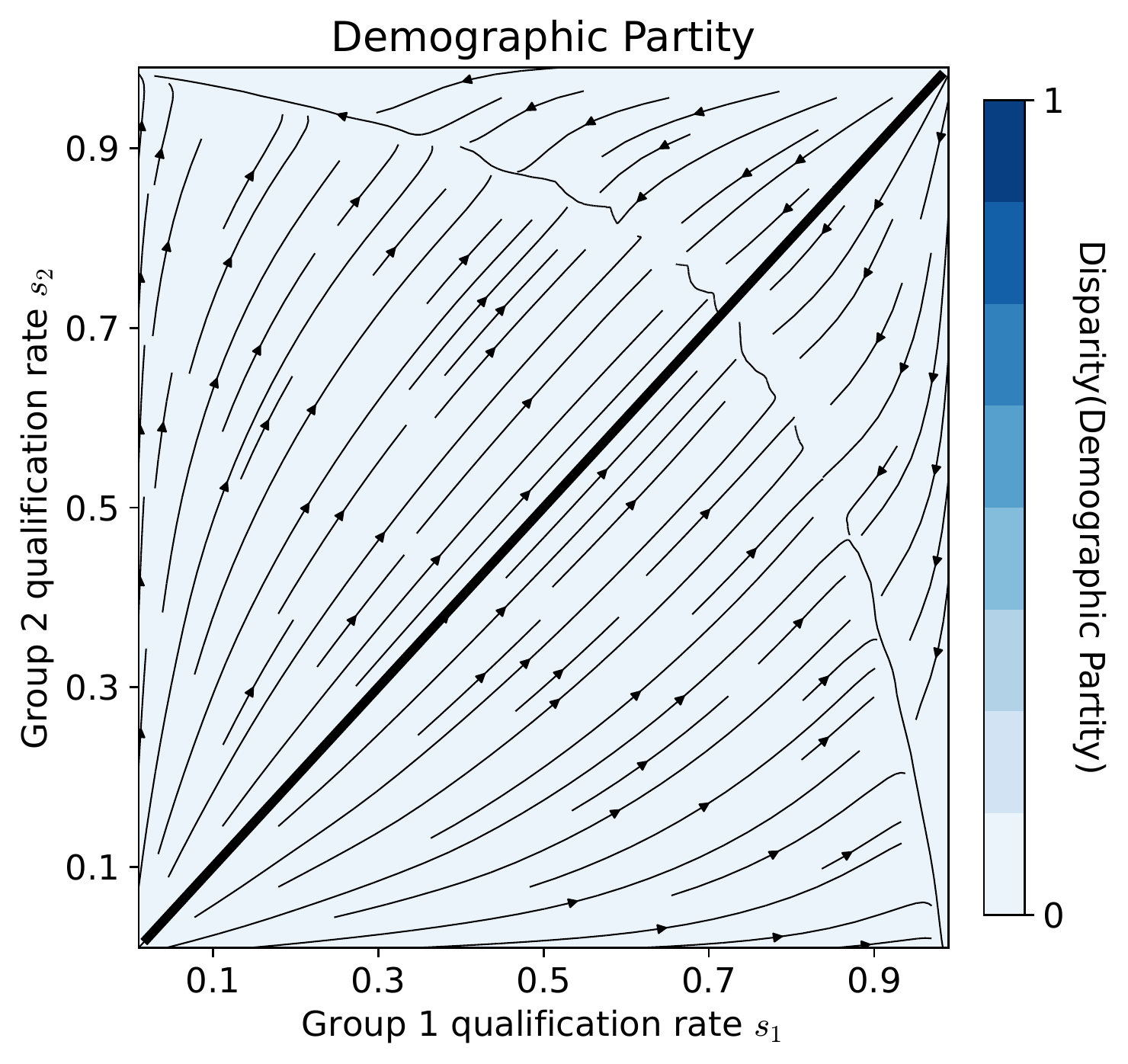}
      \includegraphics[width=0.26\textwidth]{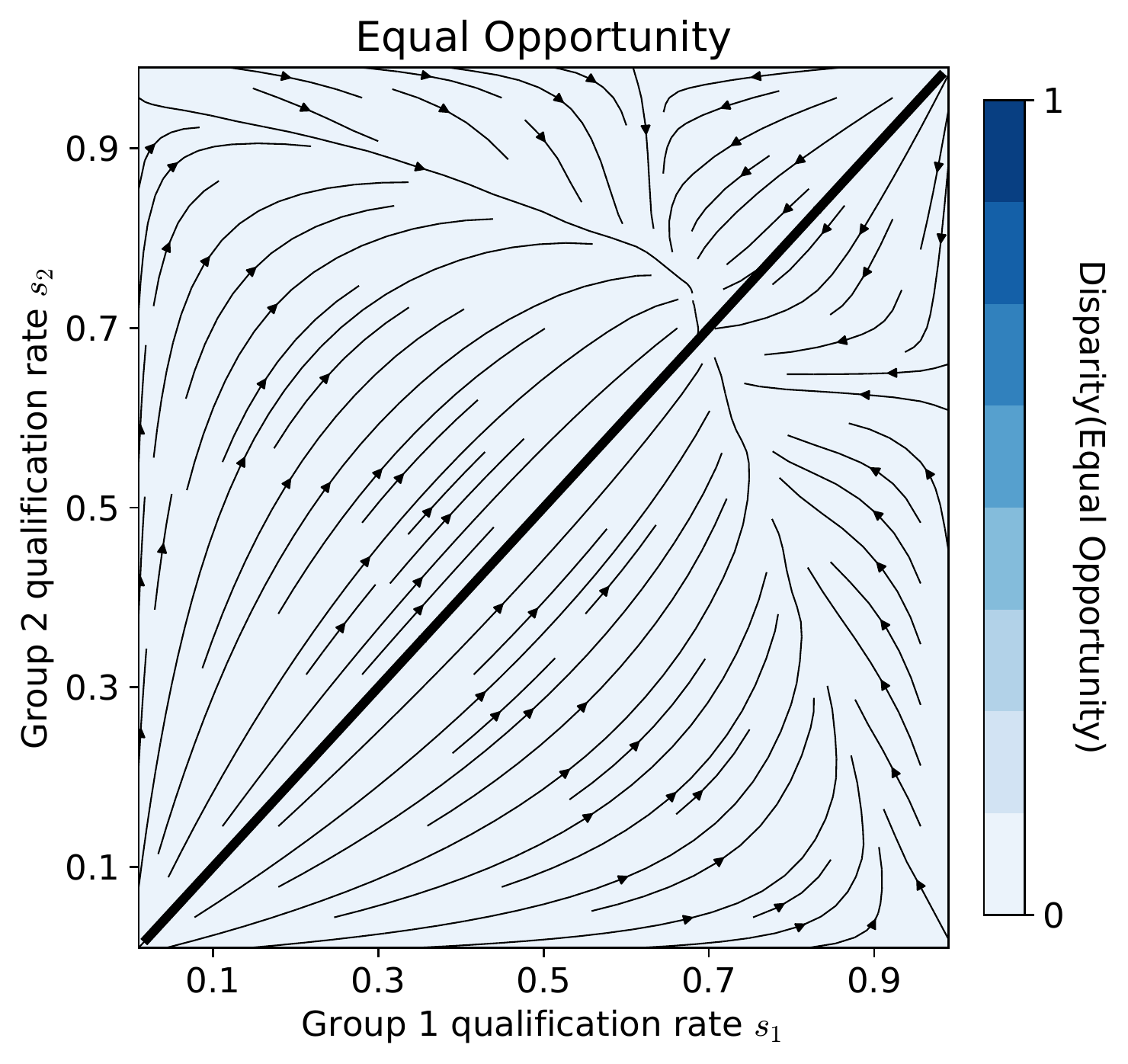}
      \includegraphics[width=0.26\textwidth]{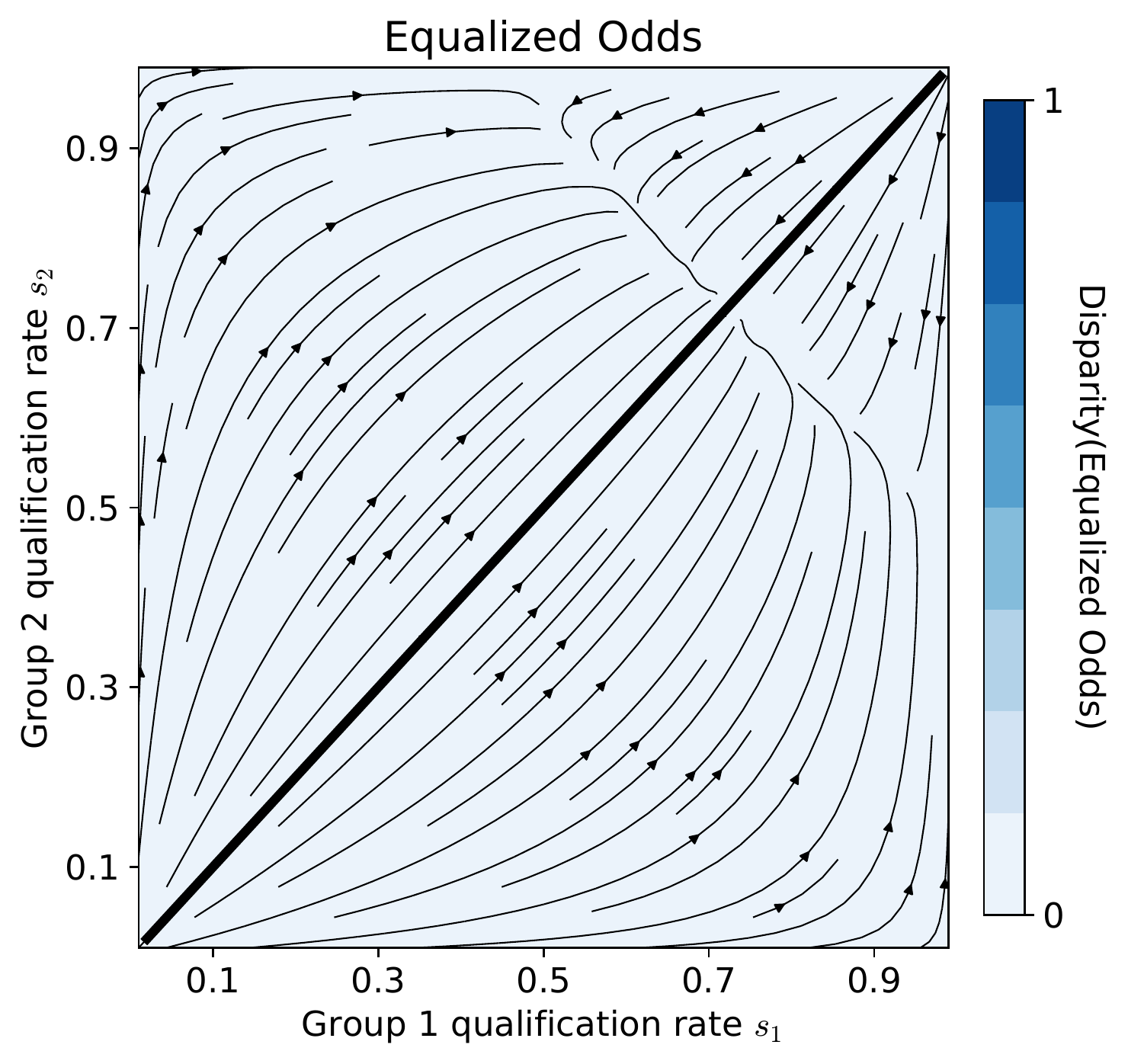}
      \caption{A baseline, greedy classifier locally maximizing utility, regularized by fairness (columns).}
      \label{fig:3a}
    \end{subfigure}
    \begin{subfigure}{\textwidth}
      \centering
      \includegraphics[width=0.26\textwidth]{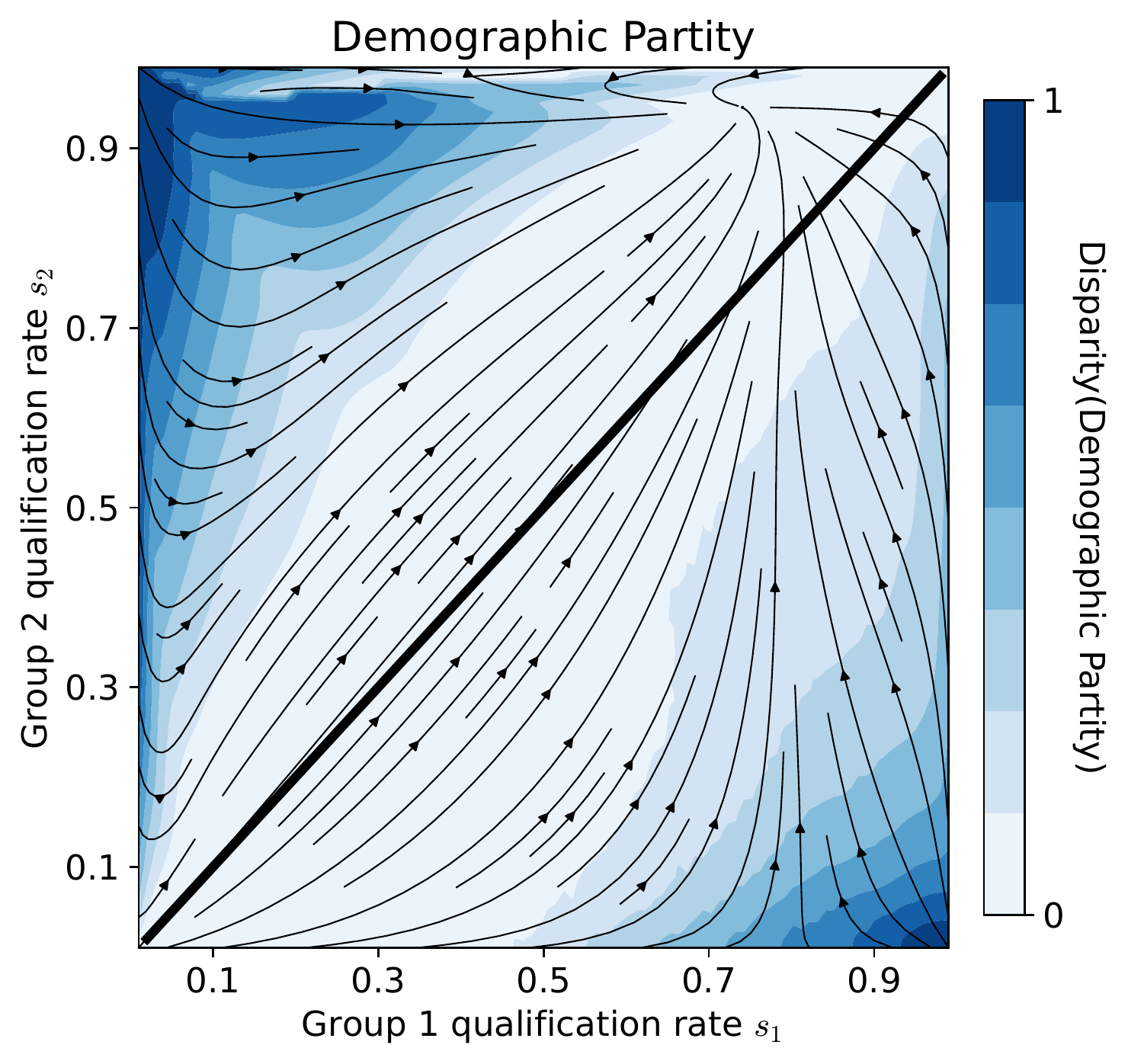}
      \includegraphics[width=0.26\textwidth]{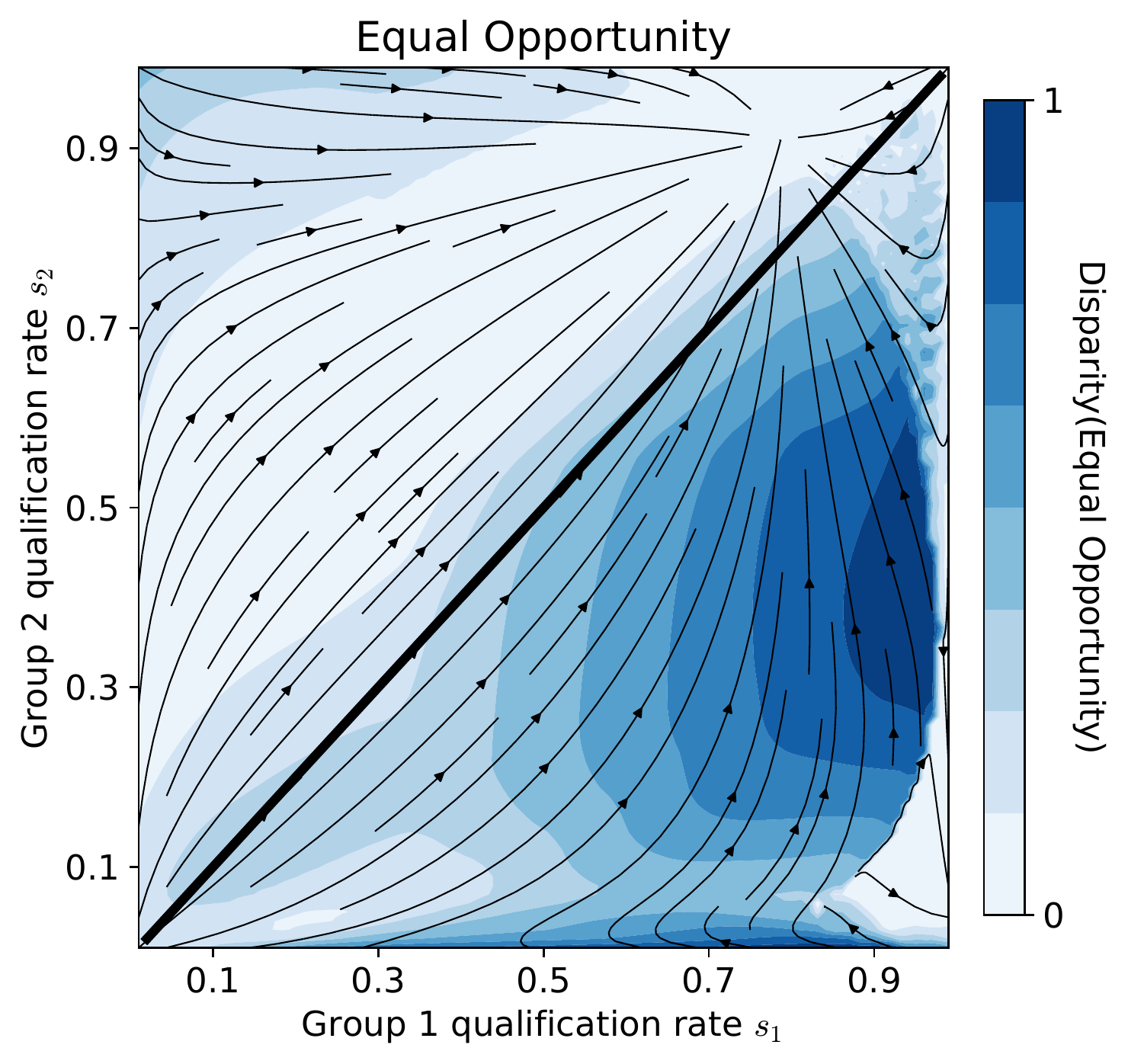}
      \includegraphics[width=0.26\textwidth]{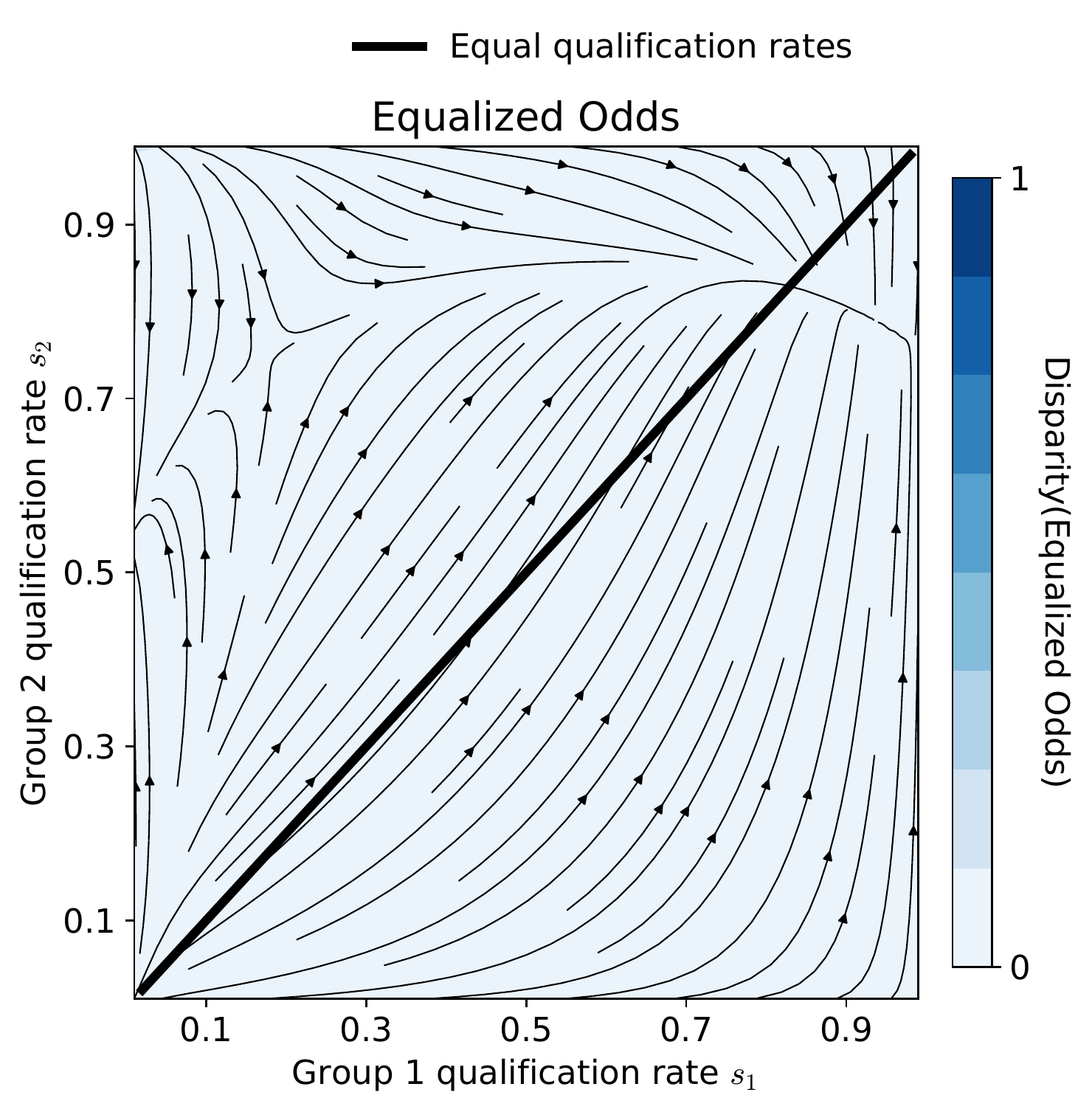}
      \caption{A \drl agent (\cref{sec:drl}) trained for 200,000 steps on the same, cumulative utility functions.}
      \label{fig:3b}
    \end{subfigure}
    \begin{subfigure}{\textwidth}
      \centering
      \includegraphics[width=0.26\textwidth]{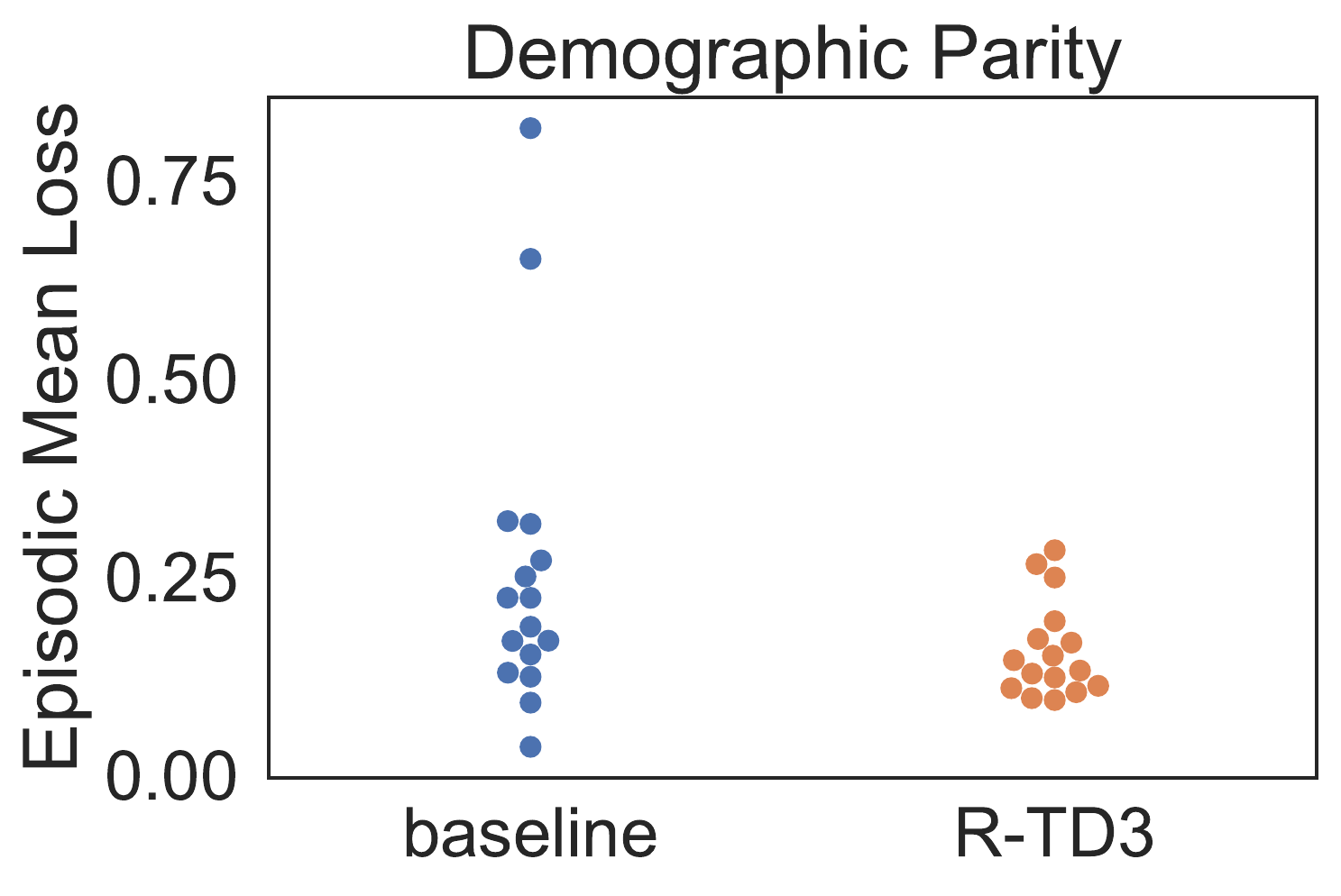}
      \includegraphics[width=0.26\textwidth]{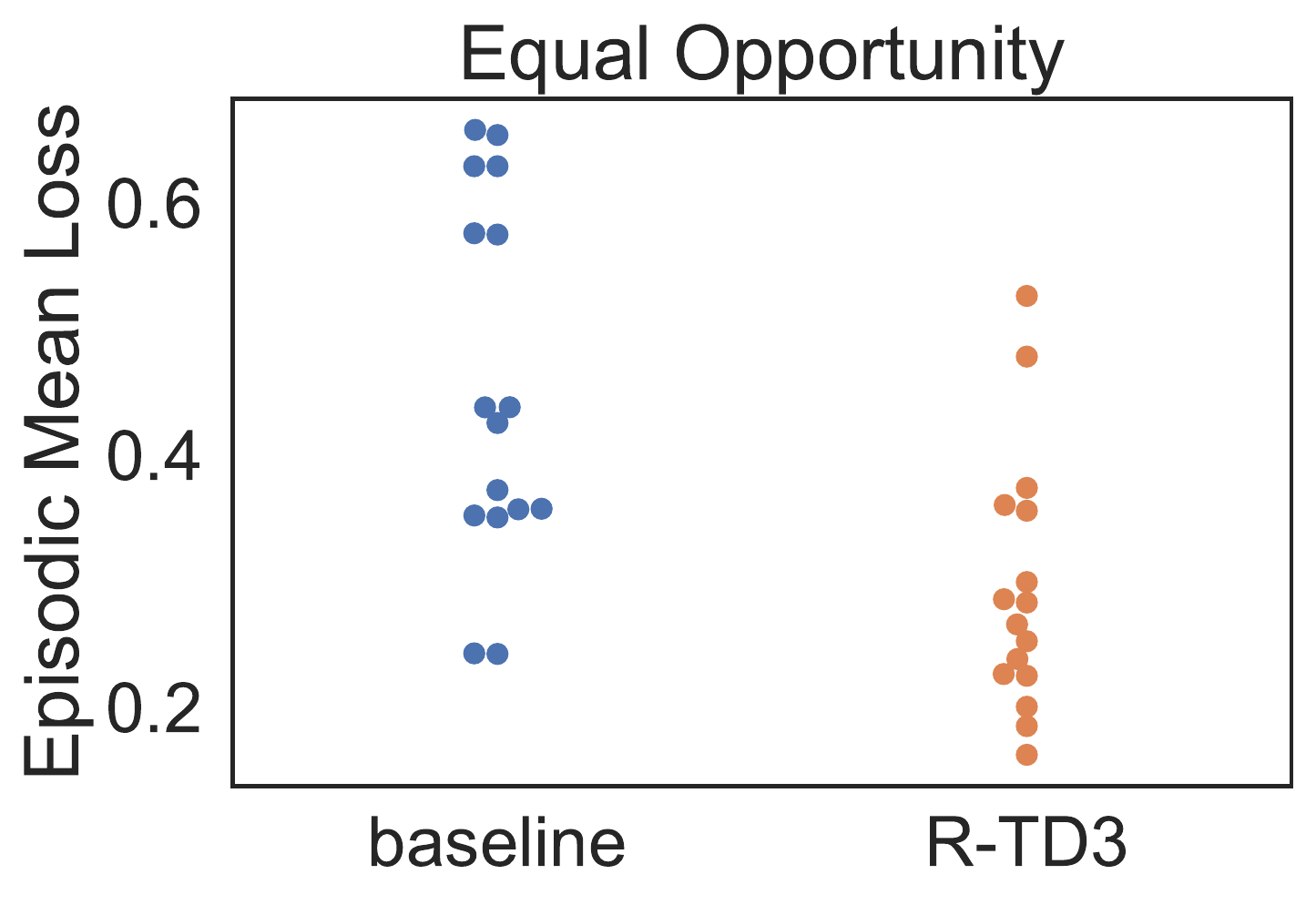}
      \includegraphics[width=0.26\textwidth]{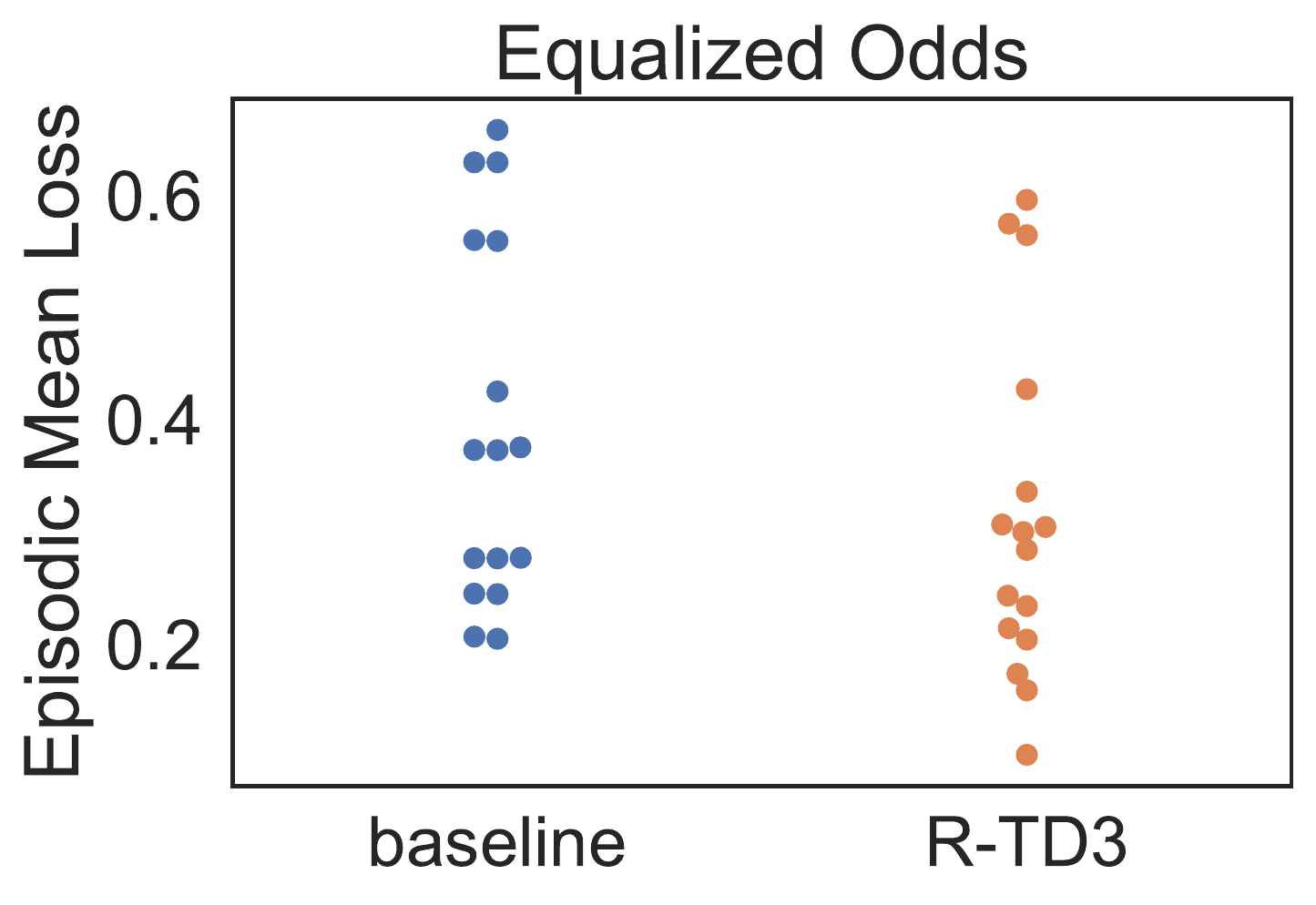}
    \end{subfigure}
    \caption{A repetition of the experiment performed in \cref{fig:1} with a
different base utility function \textbf{(true positive  fraction + 0.8 true
negative fraction)}, weighting each regularized disparity term with \(y\) = 1,
with the same \textbf{synthetic distribution}. While our observations are
largely consistent with \cref{fig:1}, we also note that the \drl agent drives a
subset of state-space in the third pane to an equilibrium less desired than the
one that the myopic agent reaches. 
}
    \label{fig:3}
\end{figure}

\begin{figure}[H]
    \centering
    \begin{subfigure}{\textwidth}
      \centering
      \includegraphics[width=0.26\textwidth]{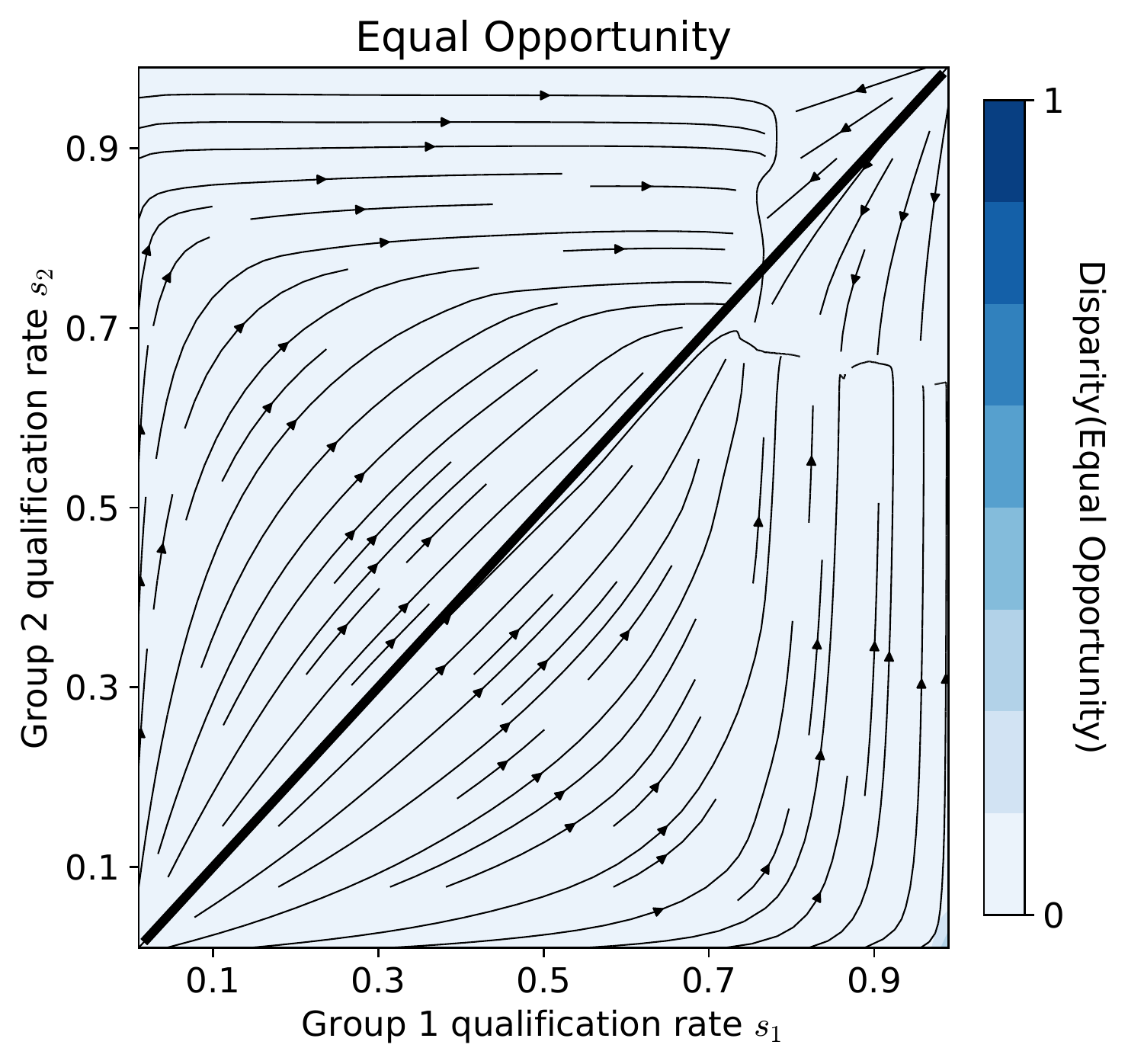}
      \includegraphics[width=0.26\textwidth]{figures/adult_baseline_loss_tp_tn+equal_opportunity_disparity0.pdf}
      \includegraphics[width=0.26\textwidth]{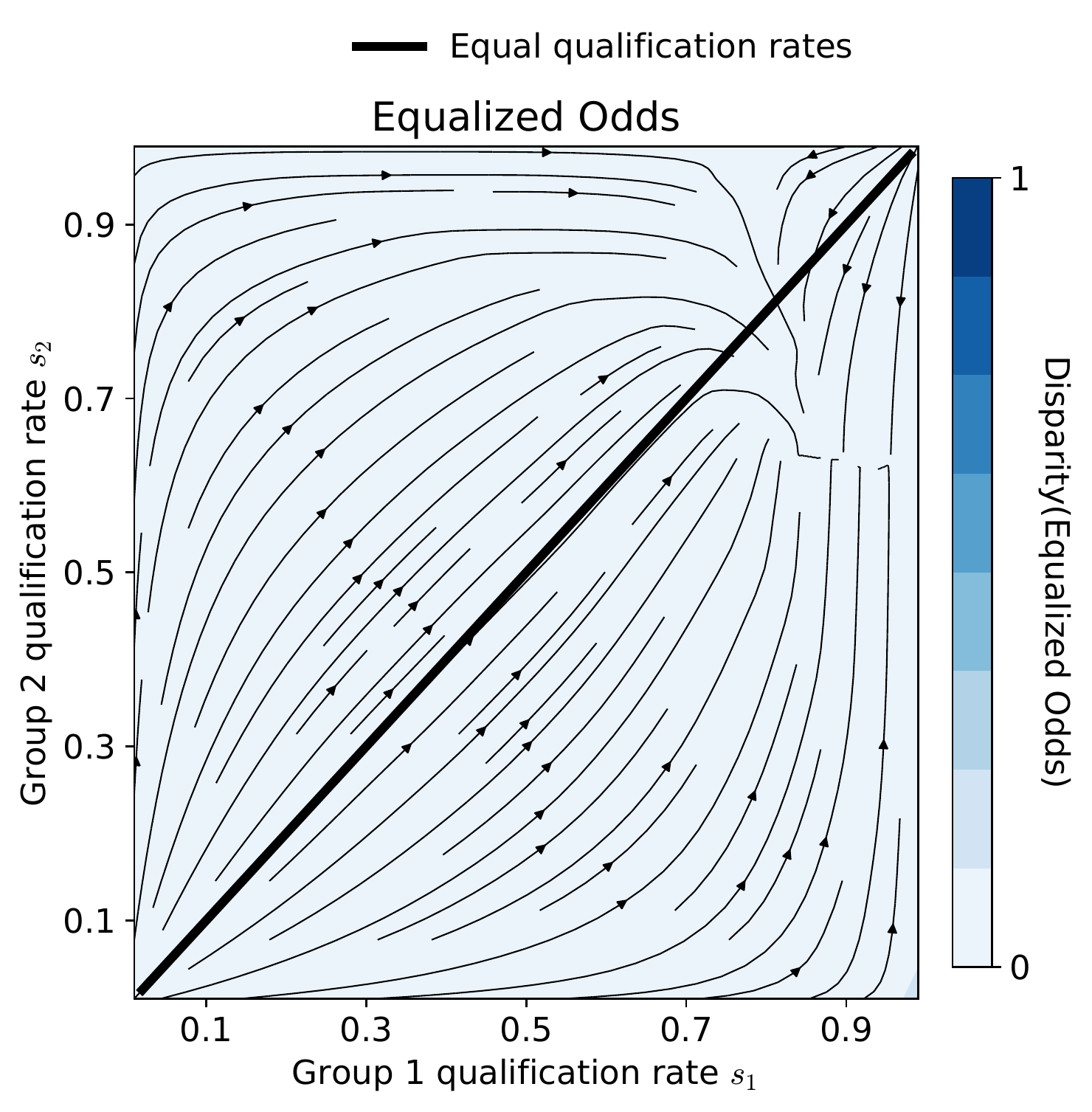}
      \caption{A baseline, greedy classifier locally maximizing utility, regularized by fairness (columns).}
      \label{fig:4a}
    \end{subfigure}
    \begin{subfigure}{\textwidth}
      \centering
      \includegraphics[width=0.26\textwidth]{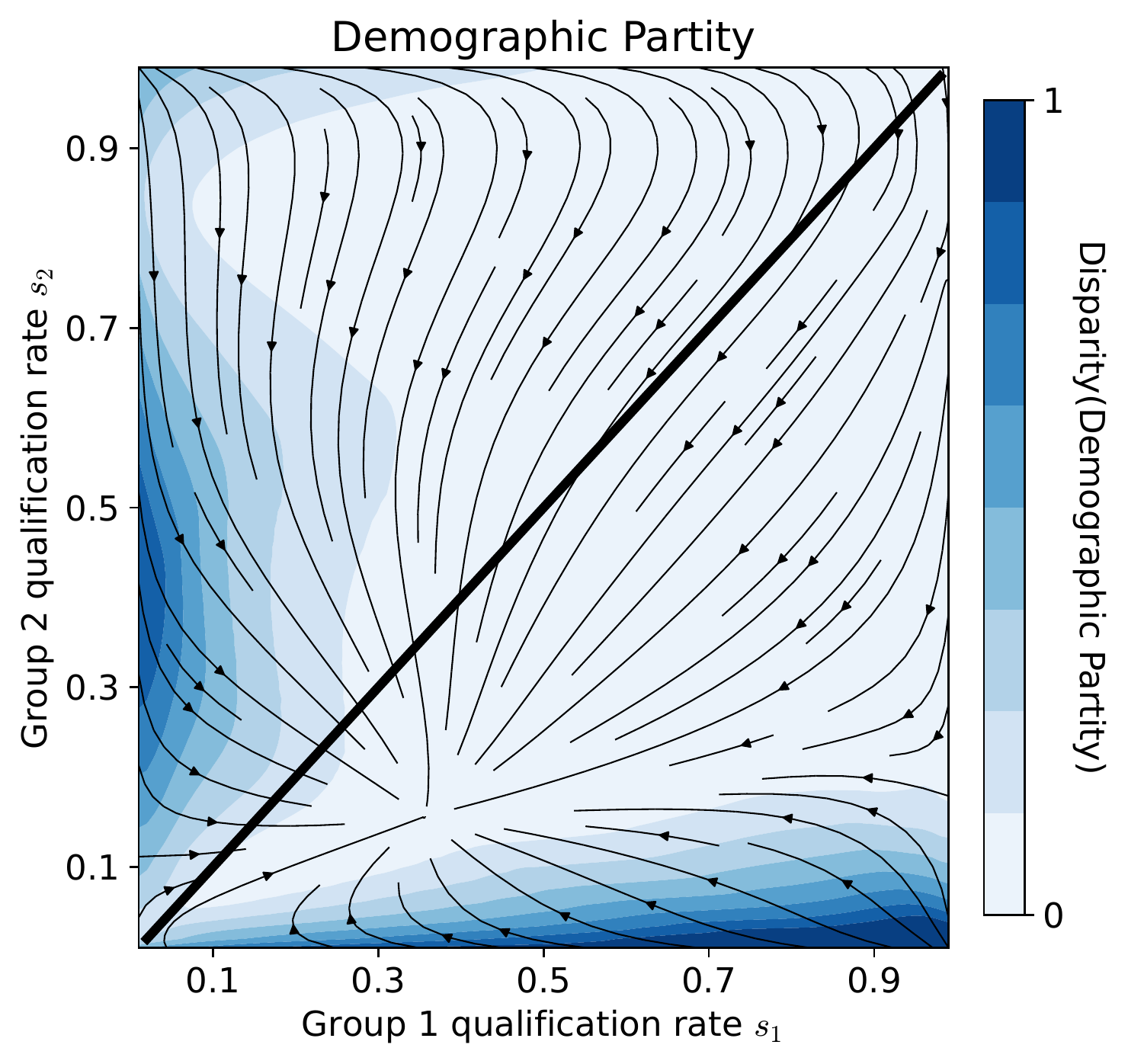}
      \includegraphics[width=0.26\textwidth]{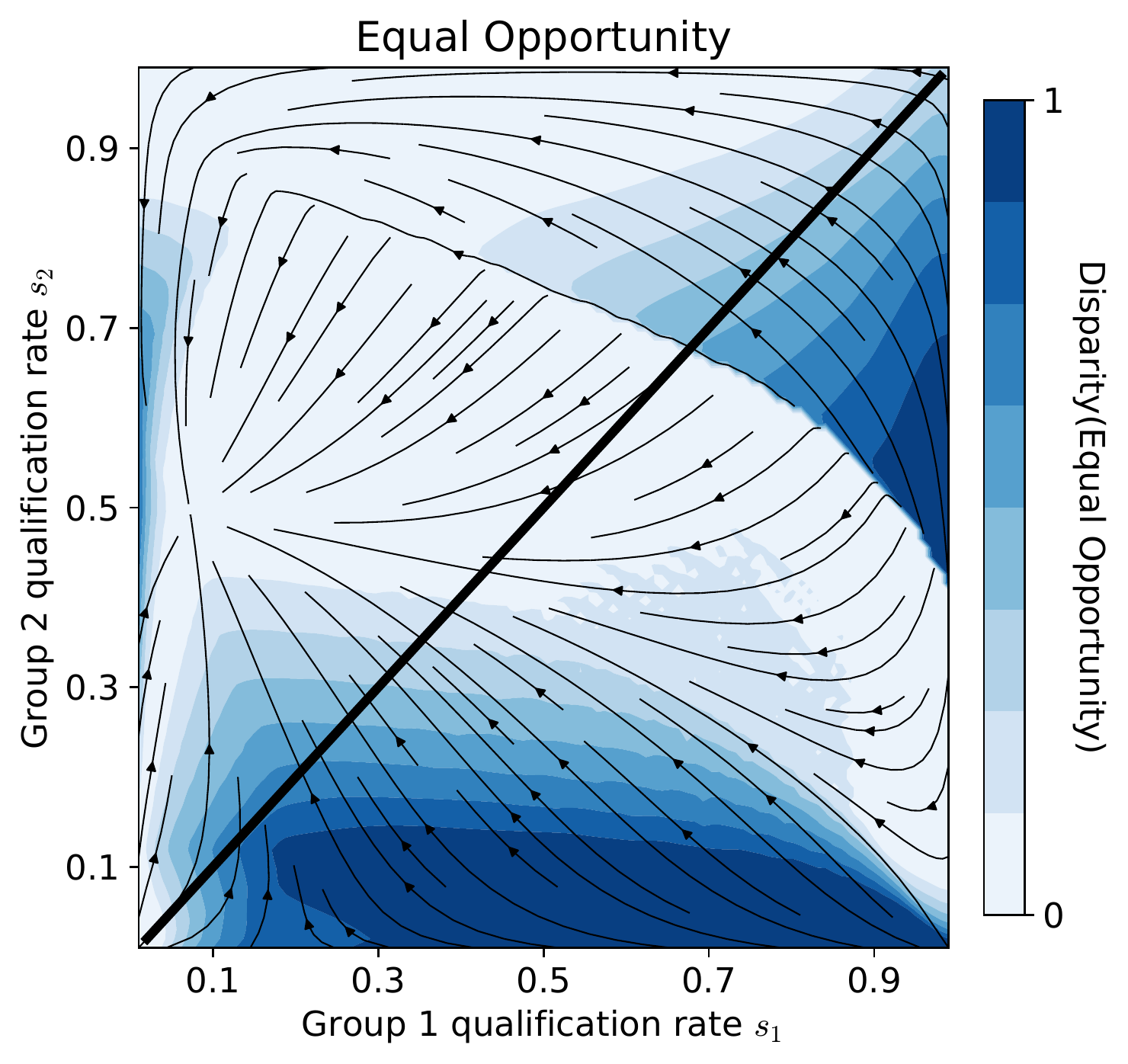}
      \includegraphics[width=0.26\textwidth]{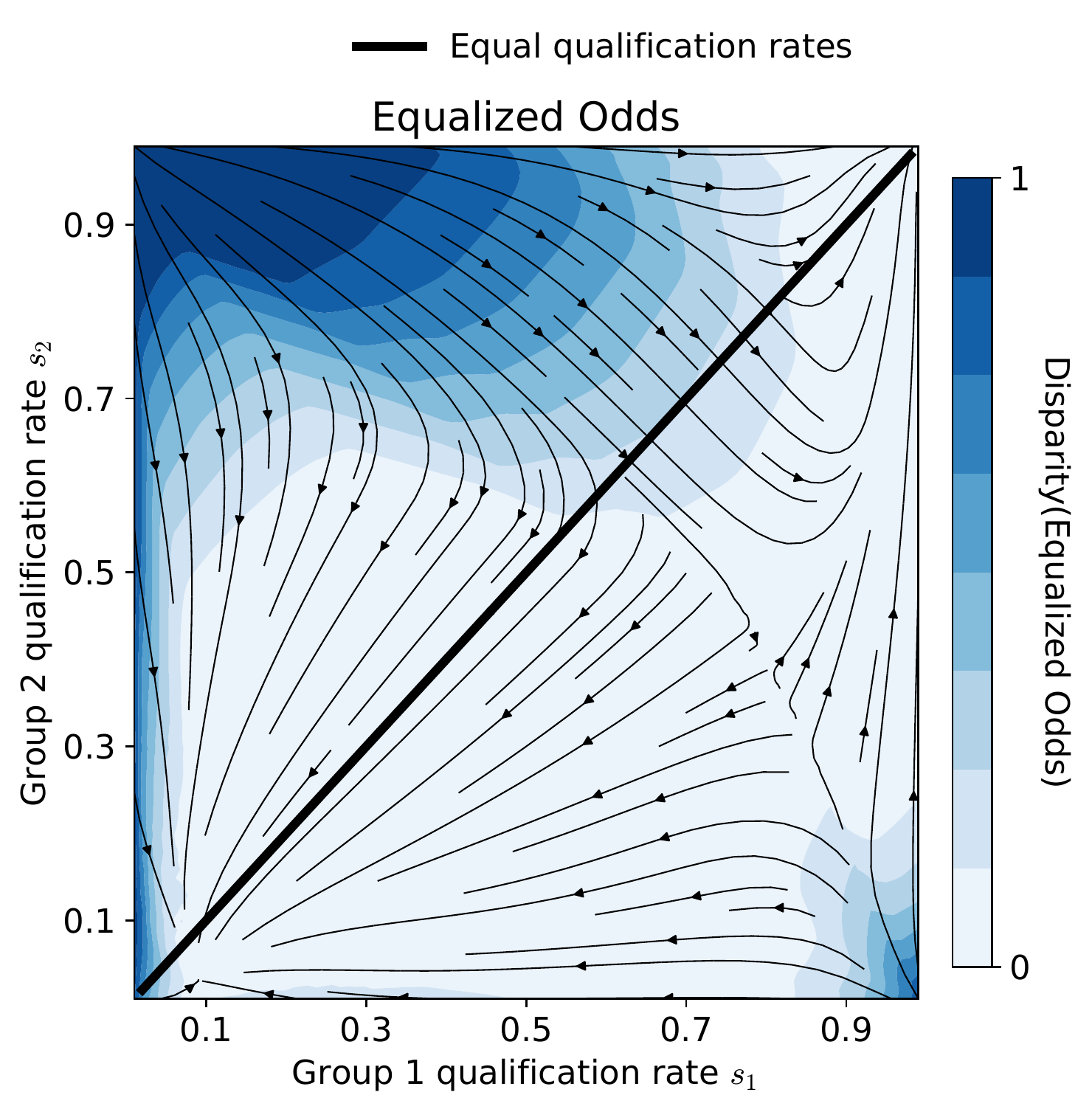}
      \caption{A \drl agent (\cref{sec:drl}) trained for 200,000 steps on the same, cumulative utility functions.}
      \label{fig:4b}
    \end{subfigure}
    \begin{subfigure}{\textwidth}
      \centering
      \includegraphics[width=0.26\textwidth]{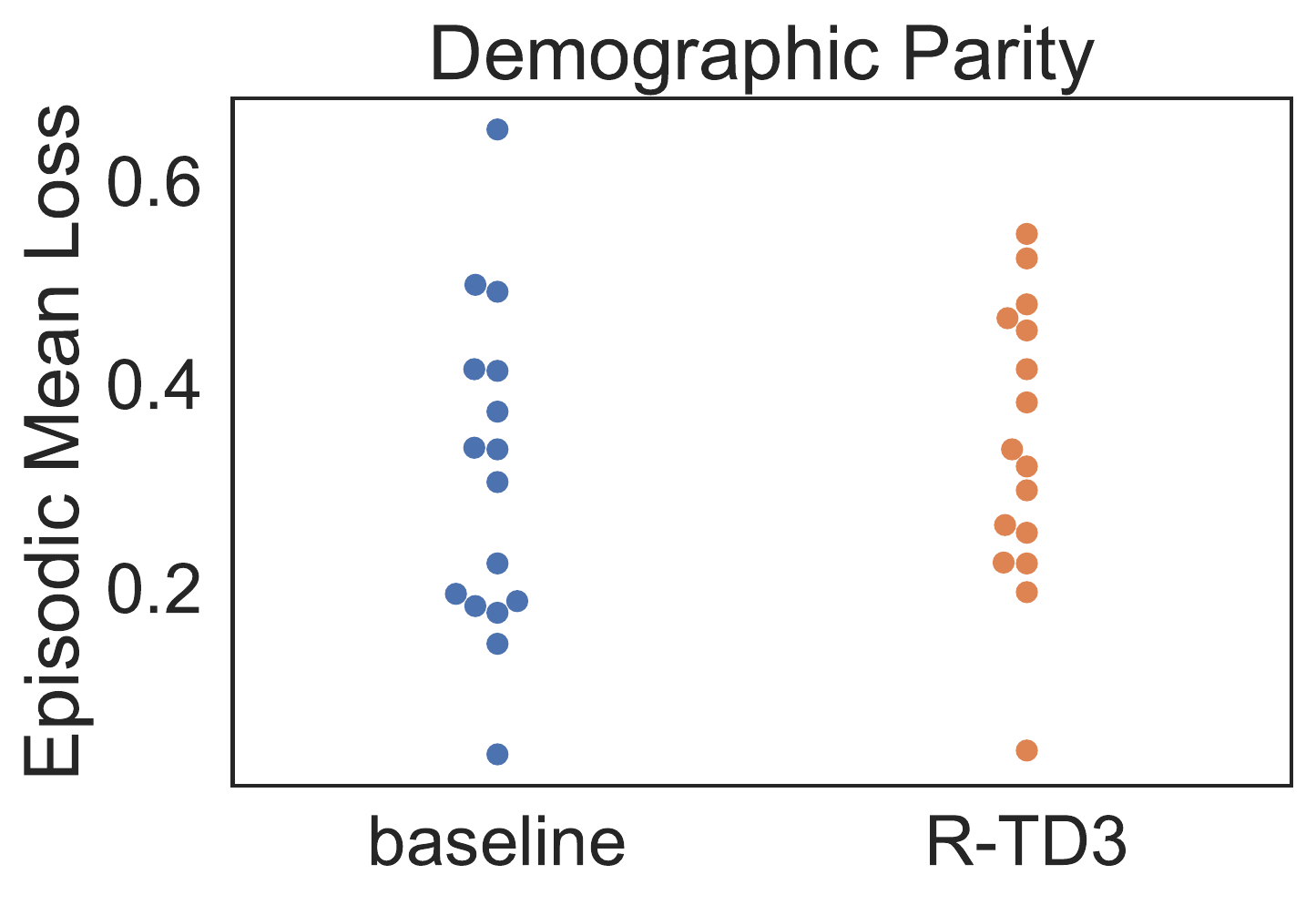}
      \includegraphics[width=0.26\textwidth]{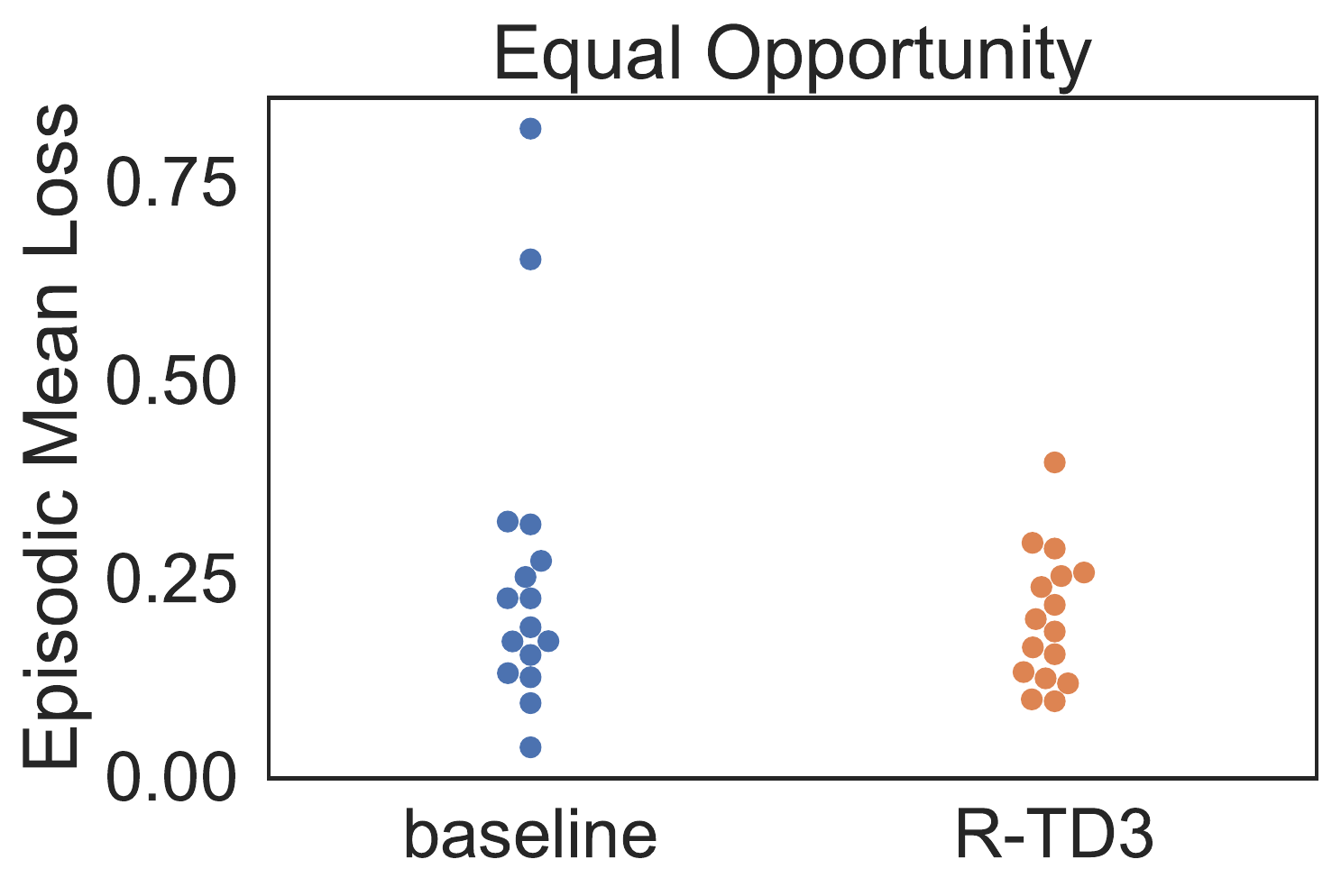}
      \includegraphics[width=0.26\textwidth]{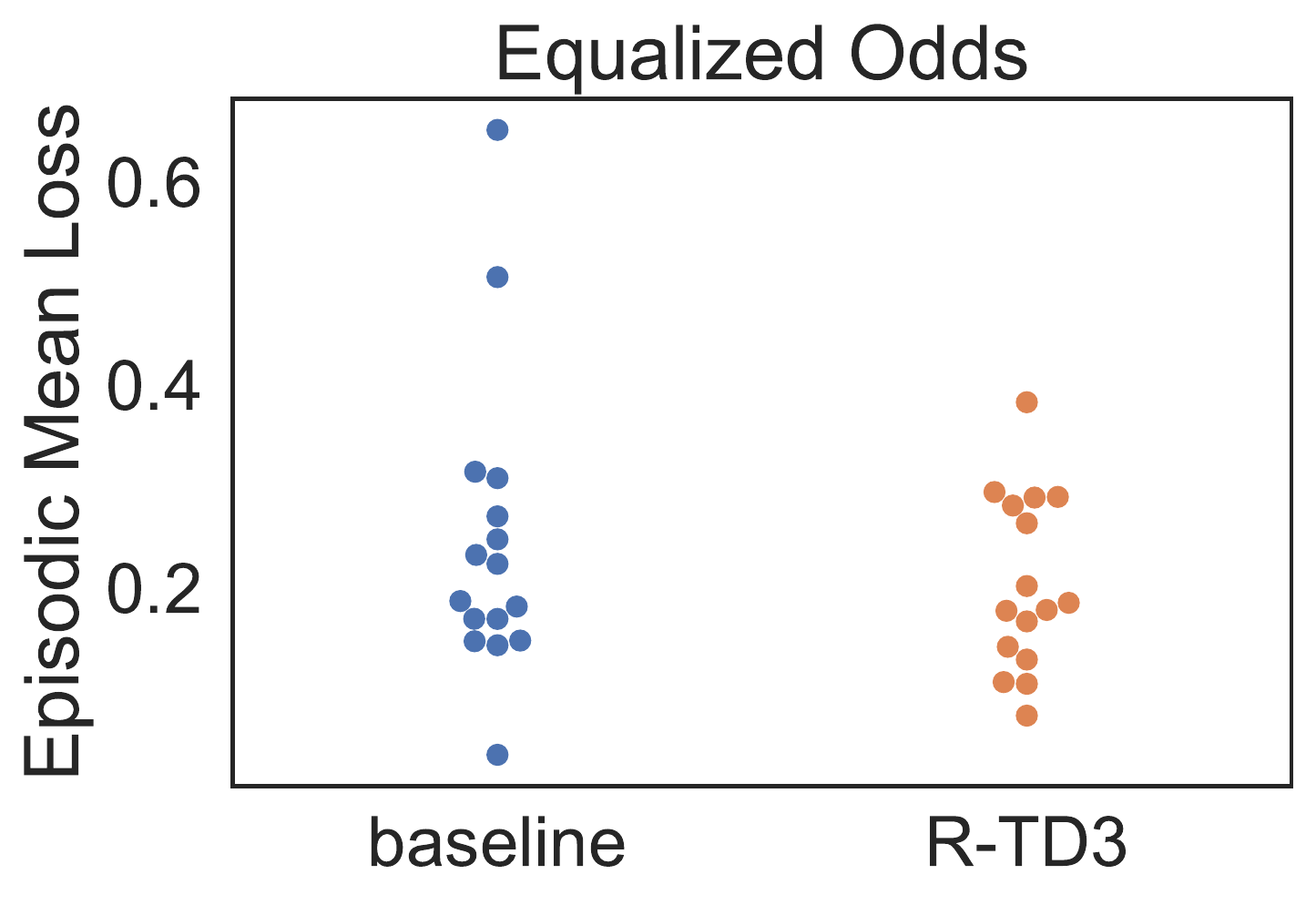}
    \end{subfigure}
     \caption{A repetition of the experiment performed in \cref{fig:3} (\ie,
with a base utility function \textbf{(true positive  fraction + 0.8 true
negative fraction)} and \(y=1\) weighted regularized disparity term), on the
\textbf{UCI ``Adult Data Set''}, as detailed in section \cref{sec:synthesis}
with groups re-weighted for equal representation.}
    \label{fig:4}
\end{figure}

\subsection{Training Curves: \ucbfair}
\subsubsection{}
\begin{figure}[H]
    \centering
    \begin{subfigure}{\textwidth}
      \centering
      \includegraphics[width=0.26\textwidth]{figures/loss_expk10000h100beta0.5tloss_tp_fraceuclidean_demographic_partity1674545114.pdf}
      \includegraphics[width=0.26\textwidth]{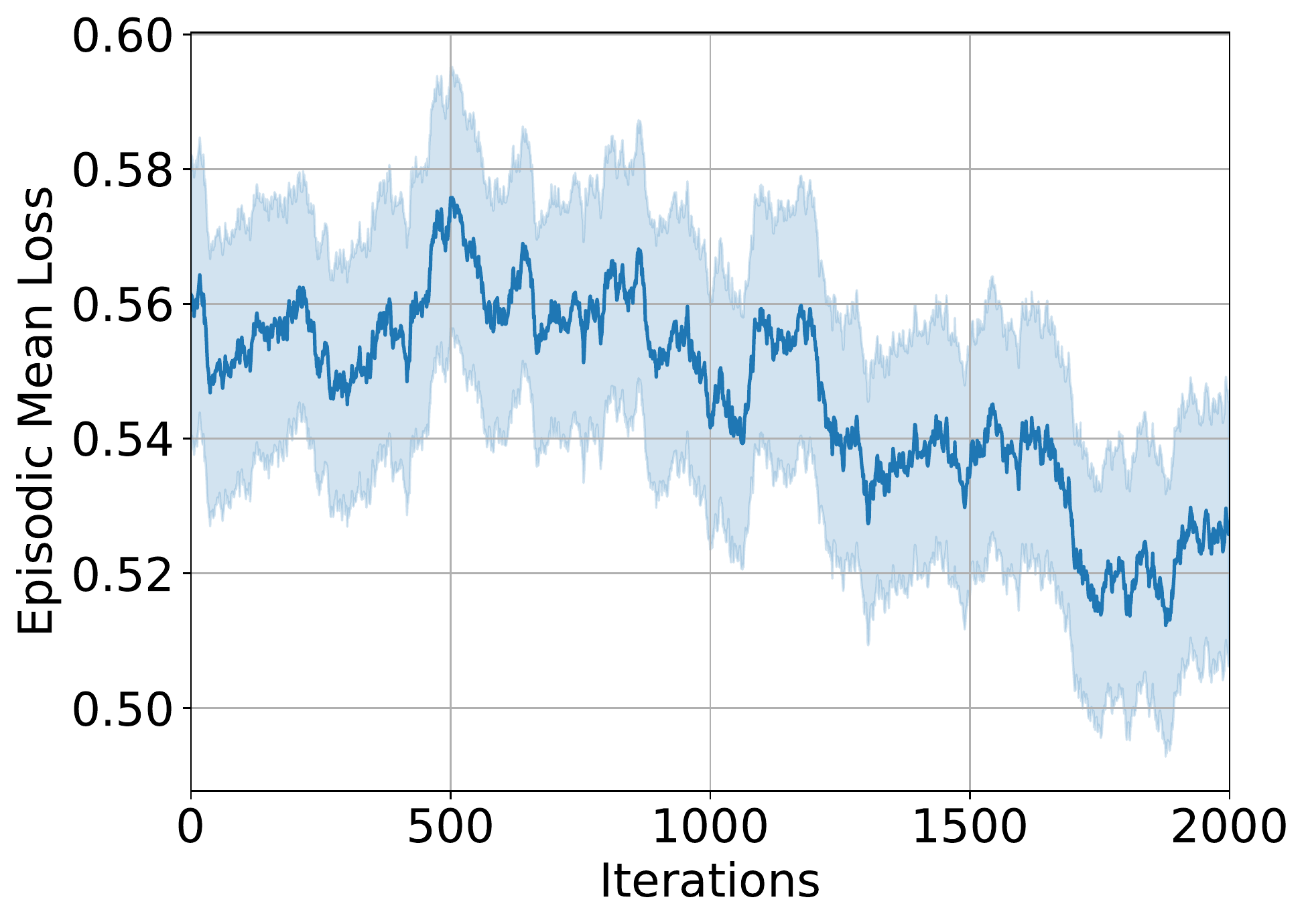}
      \includegraphics[width=0.26\textwidth]{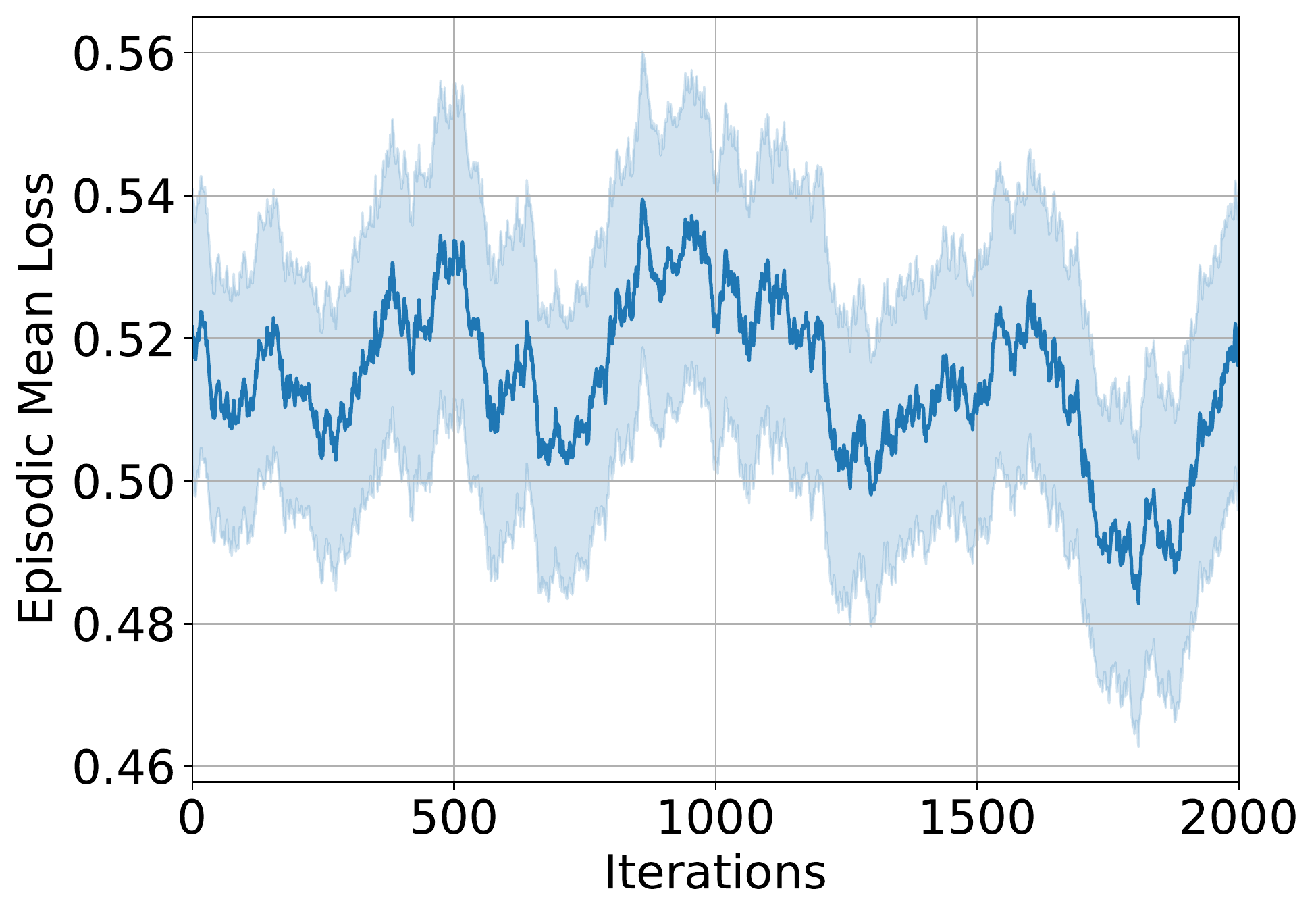}
    \end{subfigure}
    \begin{subfigure}{0.26\textwidth}
      \includegraphics[width=\textwidth]{figures/disparity_expk10000h100beta0.5tloss_tp_fracequal_opportunity_disparity1674541754.pdf}
      \caption{Demographic Parity}
    \end{subfigure}
    \begin{subfigure}{0.26\textwidth}
      \includegraphics[width=\textwidth]{figures/disparity_expk10000h100beta0.5tloss_tp_fracequal_opportunity_disparity1674541754.pdf}
      \caption{Equal Opportunity}
    \end{subfigure}
    \begin{subfigure}{0.26\textwidth}
      \includegraphics[width=\textwidth]{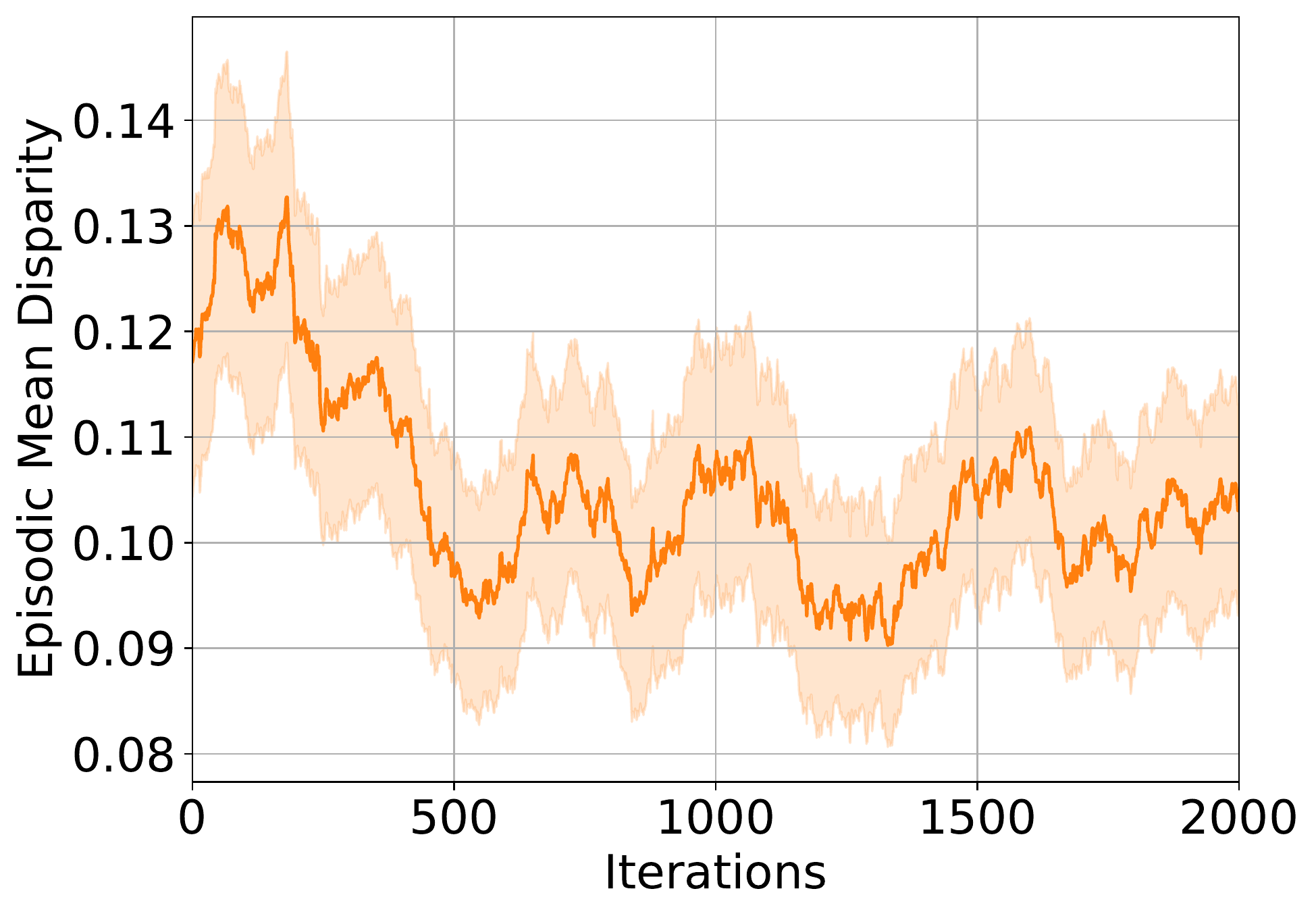}
      \caption{Equalized Odds}
    \end{subfigure}
    \caption{\ucbfair  20-step sliding mean \& std for the setting in \cref{fig:1}.}
    \label{fig:ucbfair_curve}
\end{figure}

\begin{figure}[H]
    \centering
    \begin{subfigure}{\textwidth}
      \centering
      \includegraphics[width=0.26\textwidth]{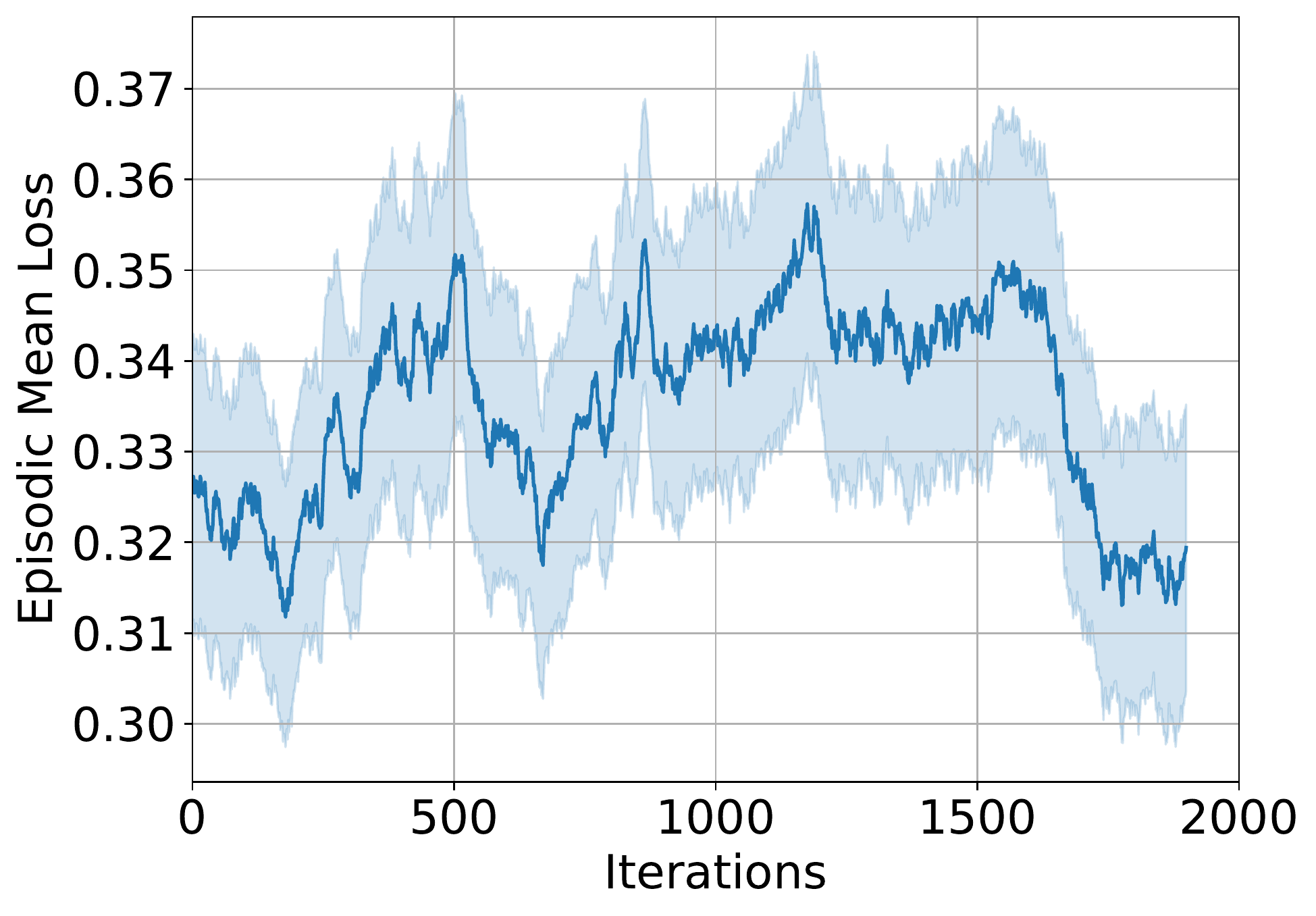}
      \includegraphics[width=0.26\textwidth]{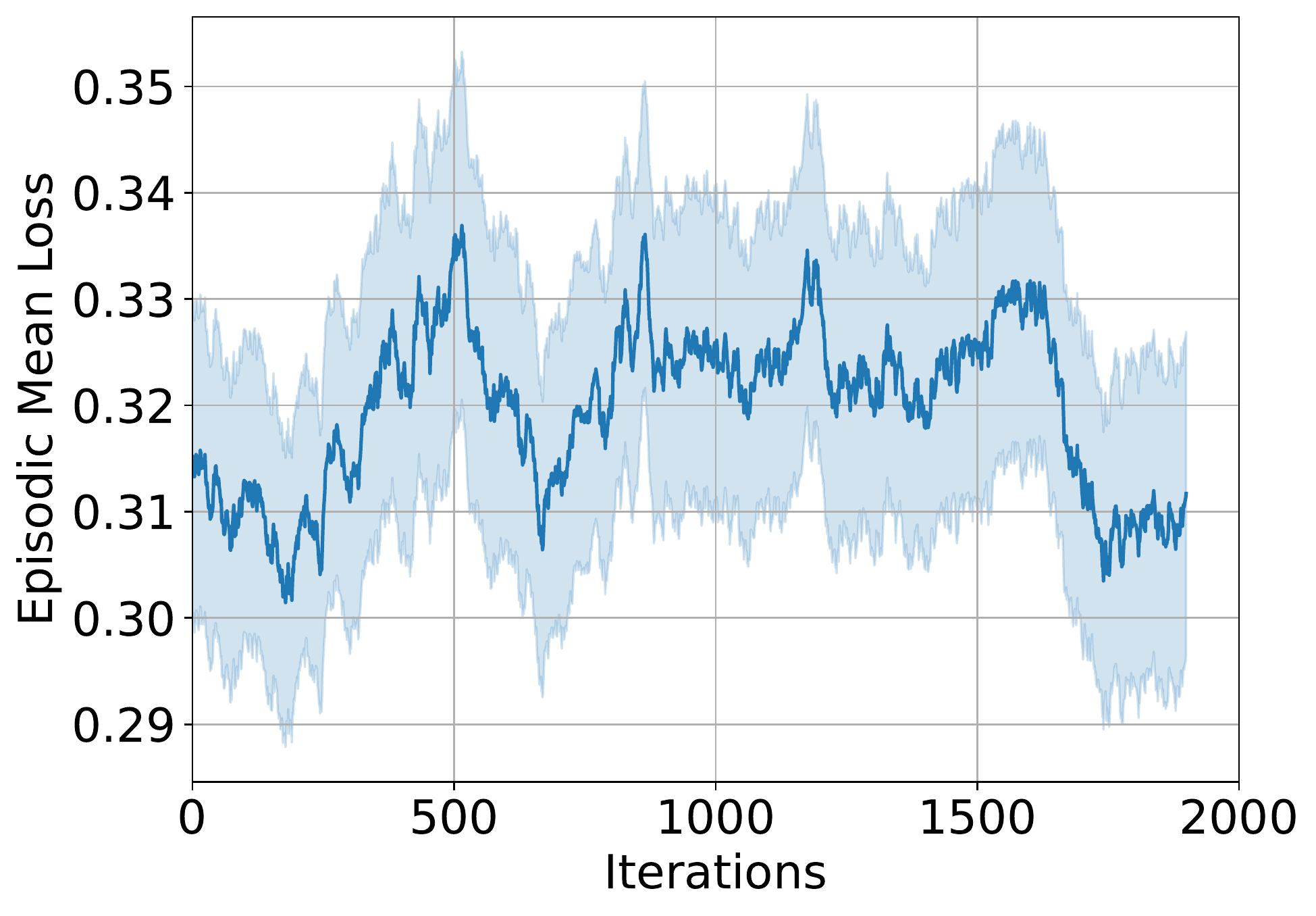}
      \includegraphics[width=0.26\textwidth]{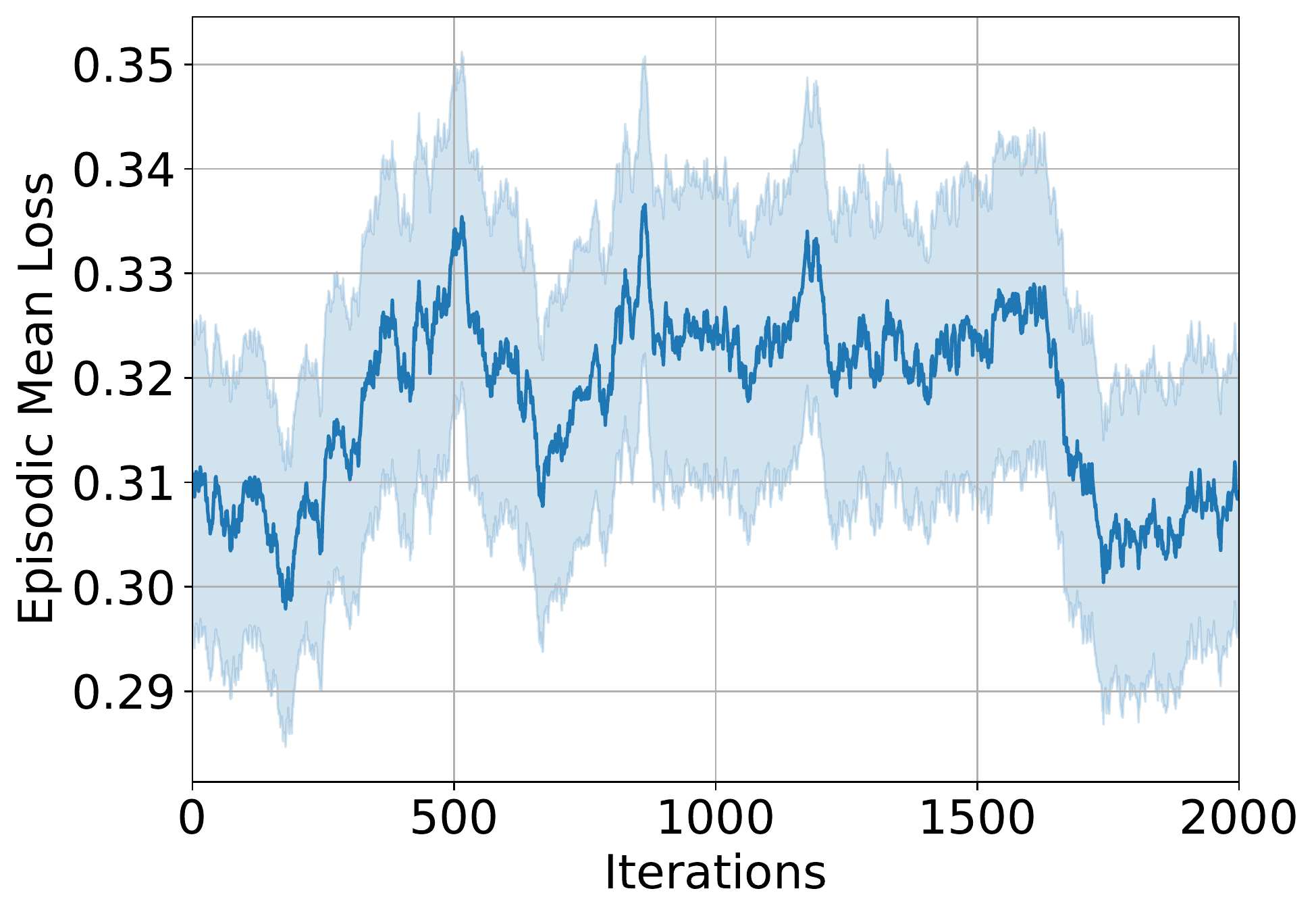}
    \end{subfigure}
    \begin{subfigure}{0.26\textwidth}
      \includegraphics[width=\textwidth]{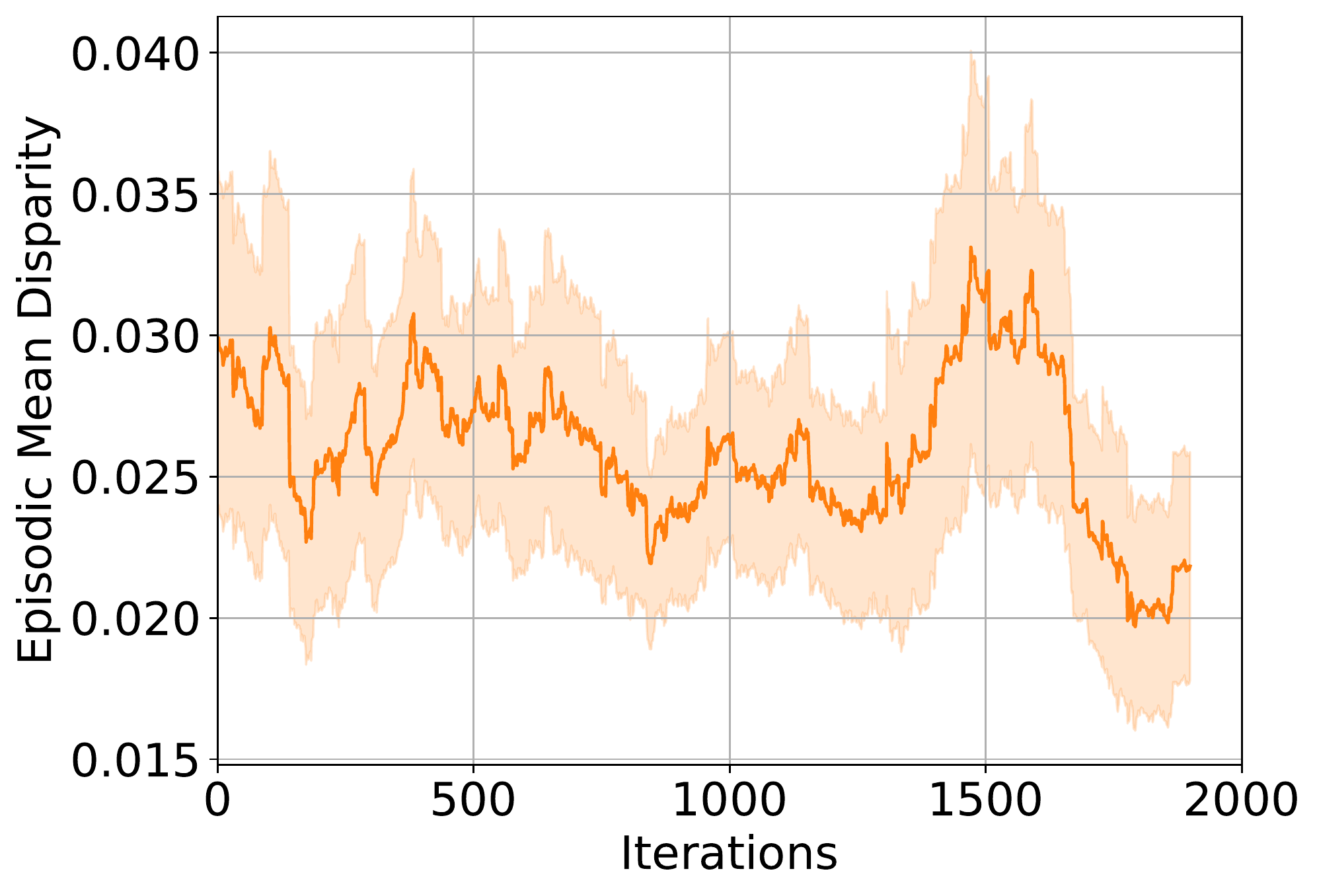}
      \caption{Demographic Parity}
    \end{subfigure}
    \begin{subfigure}{0.26\textwidth}
      \includegraphics[width=\textwidth]{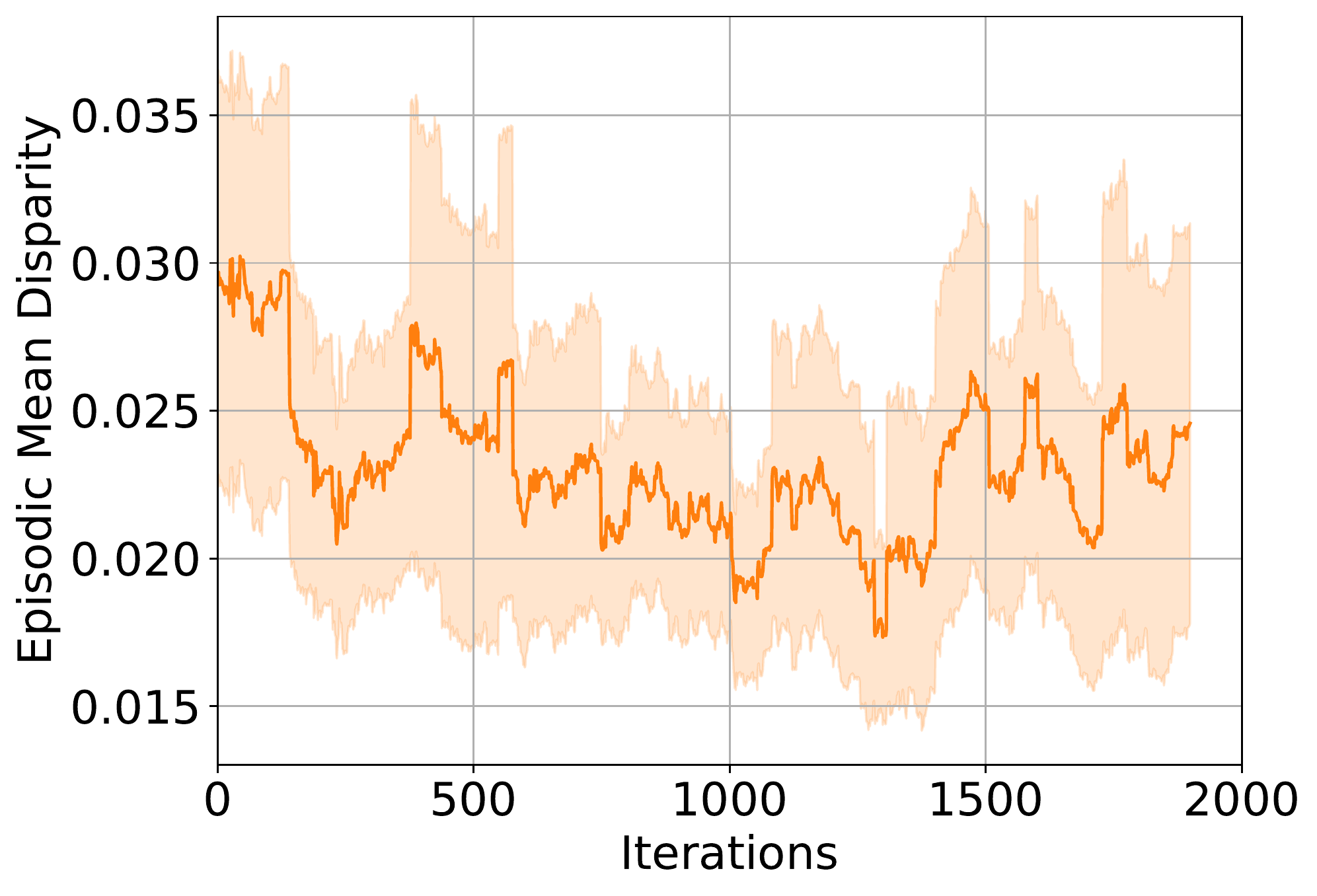}
      \caption{Equal Opportunity}
    \end{subfigure}
    \begin{subfigure}{0.26\textwidth}
      \includegraphics[width=\textwidth]{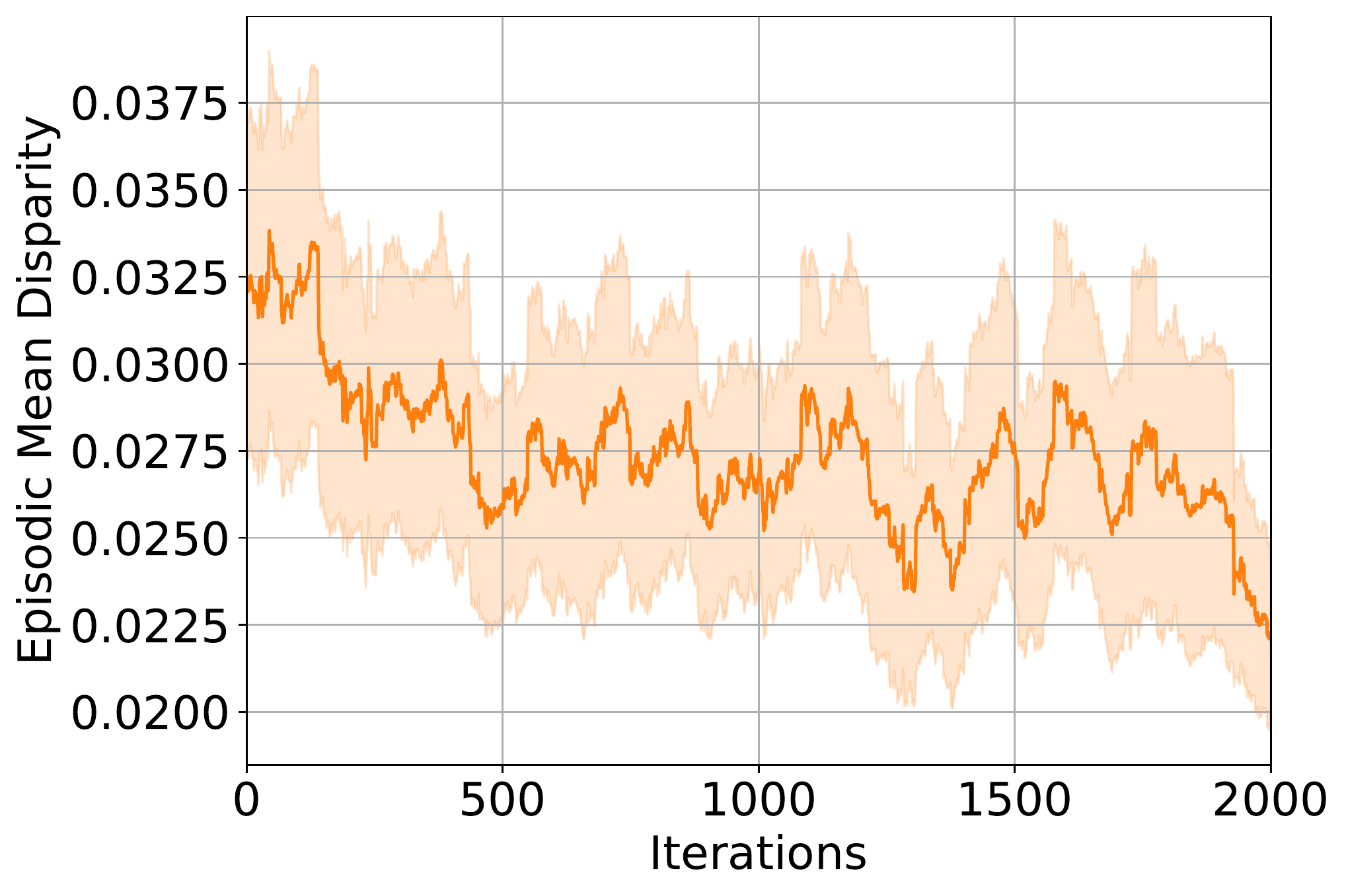}
      \caption{Equalized Odds}
    \end{subfigure}
    \caption{\ucbfair  20-step sliding mean \& std for the setting in \cref{fig:2}.}
    \label{fig:ucbfair_curve}
\end{figure}

\subsection{Training Curves: \drl}

\begin{figure}[H]
    \centering
    \begin{subfigure}{\textwidth}
      \centering
      \includegraphics[width=0.26\textwidth]{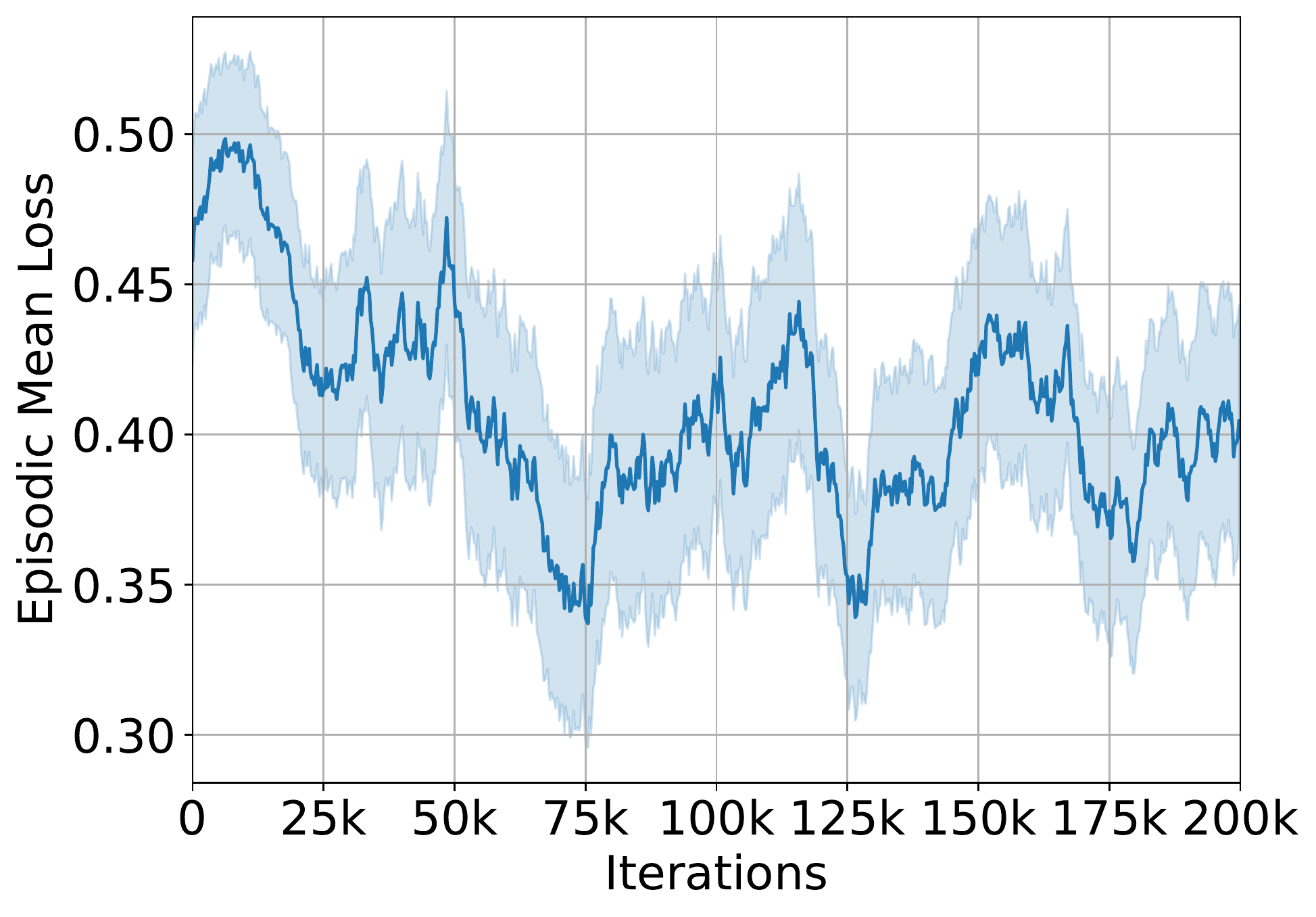}
      \includegraphics[width=0.26\textwidth]{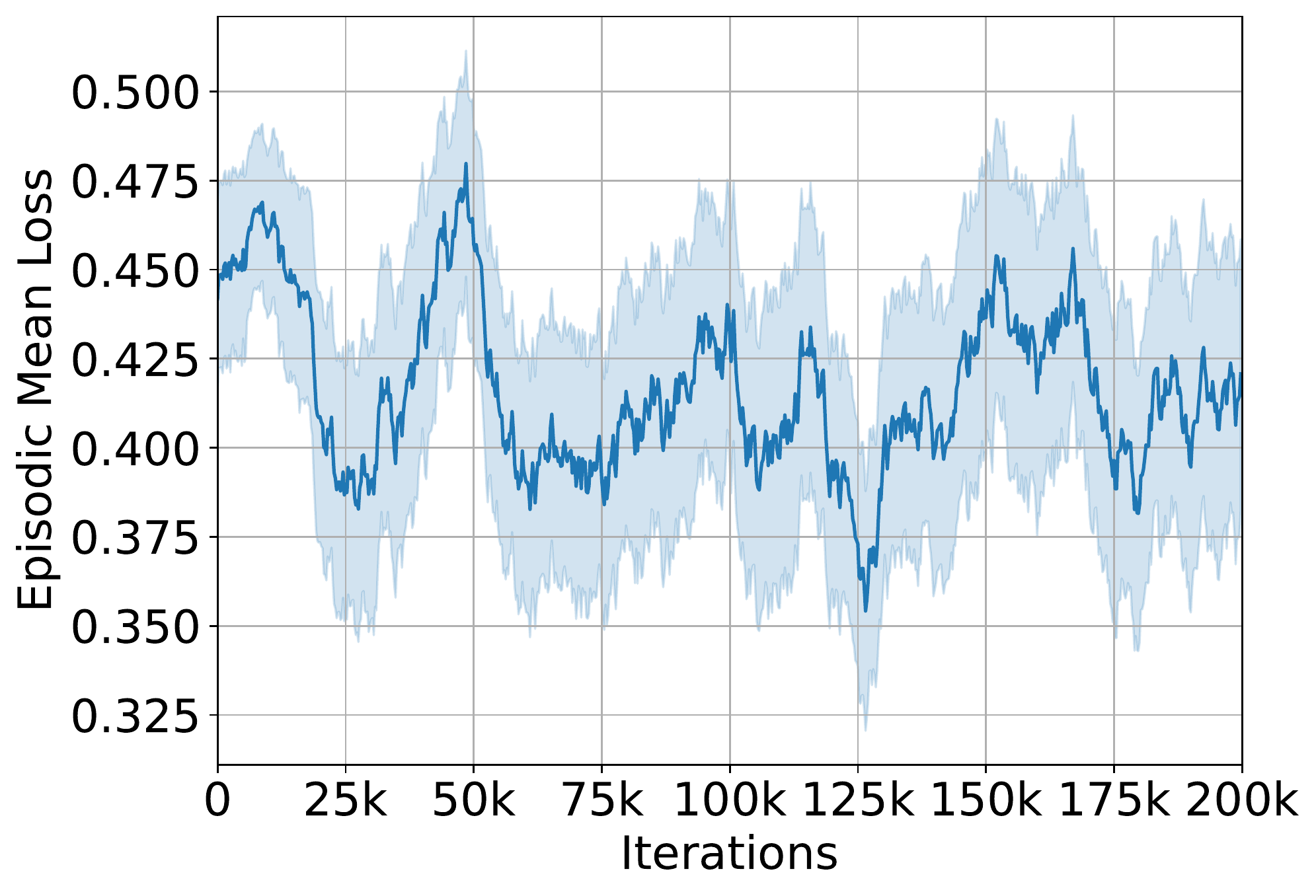}
      \includegraphics[width=0.26\textwidth]{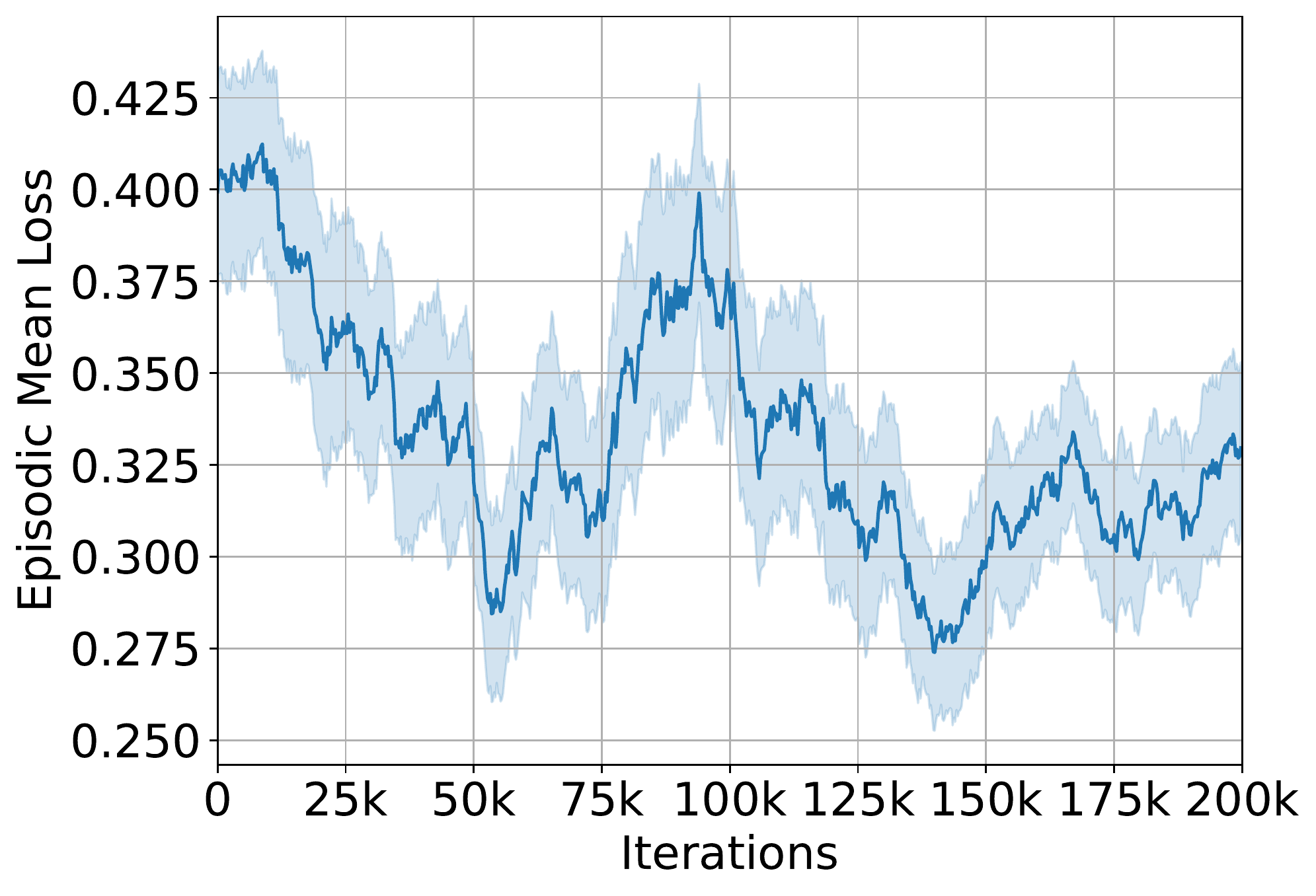}
    \end{subfigure}
    \begin{subfigure}{0.26\textwidth}
      \includegraphics[width=\textwidth]{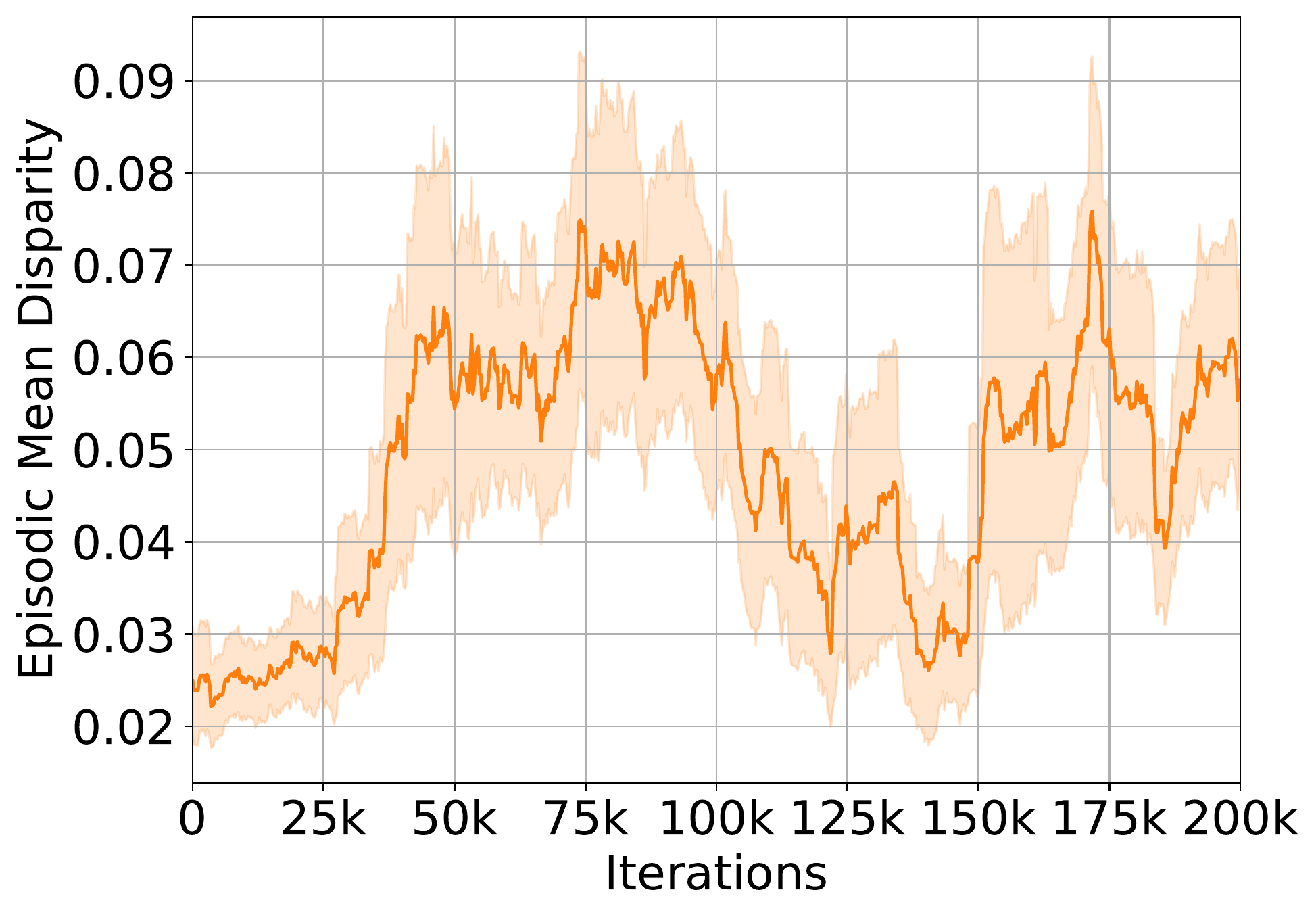}
      \caption{Demographic Parity}
    \end{subfigure}
    \begin{subfigure}{0.26\textwidth}
      \includegraphics[width=\textwidth]{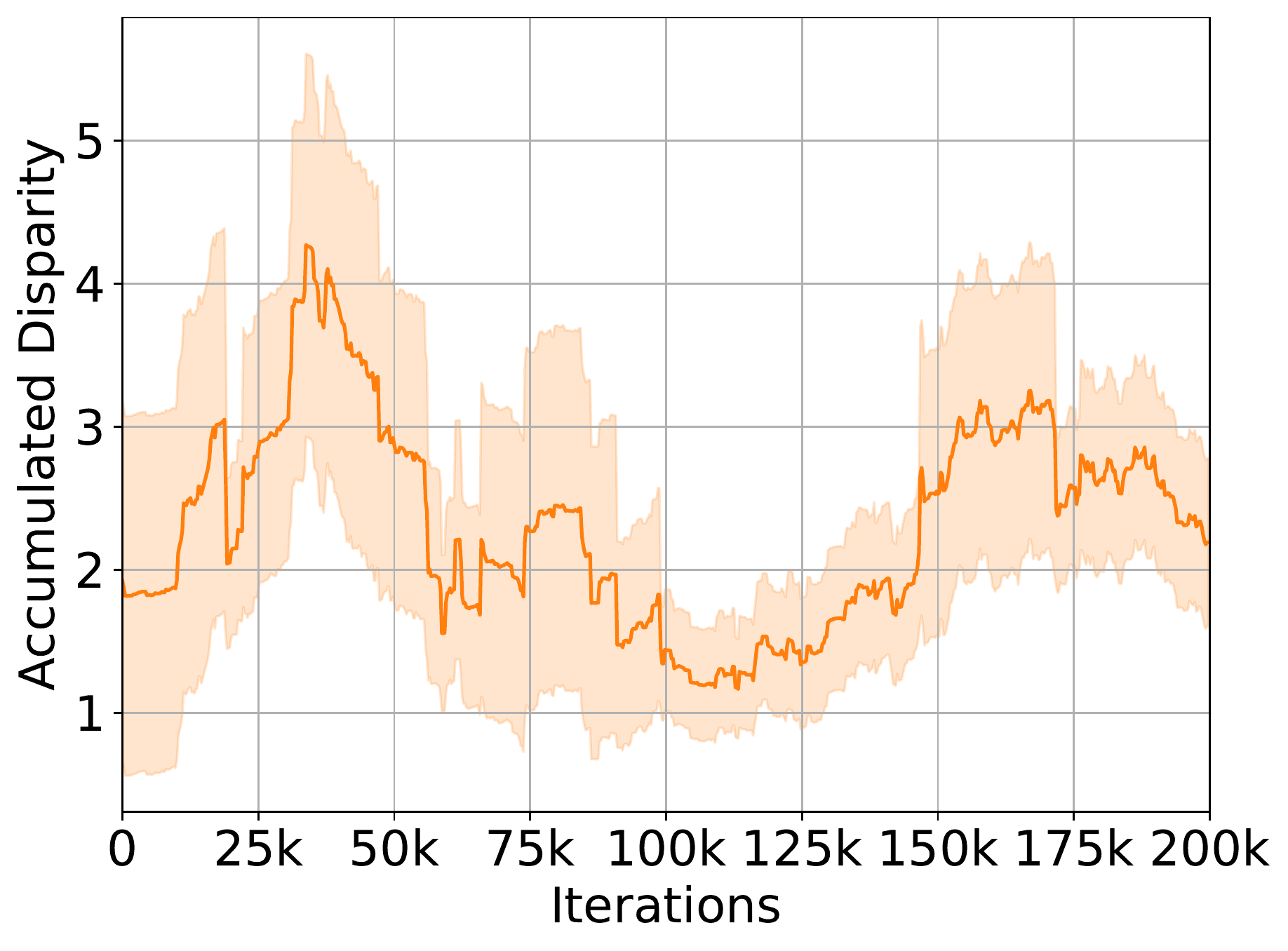}
      \caption{Equal Opportunity}
    \end{subfigure}
    \begin{subfigure}{0.26\textwidth}
      \includegraphics[width=\textwidth]{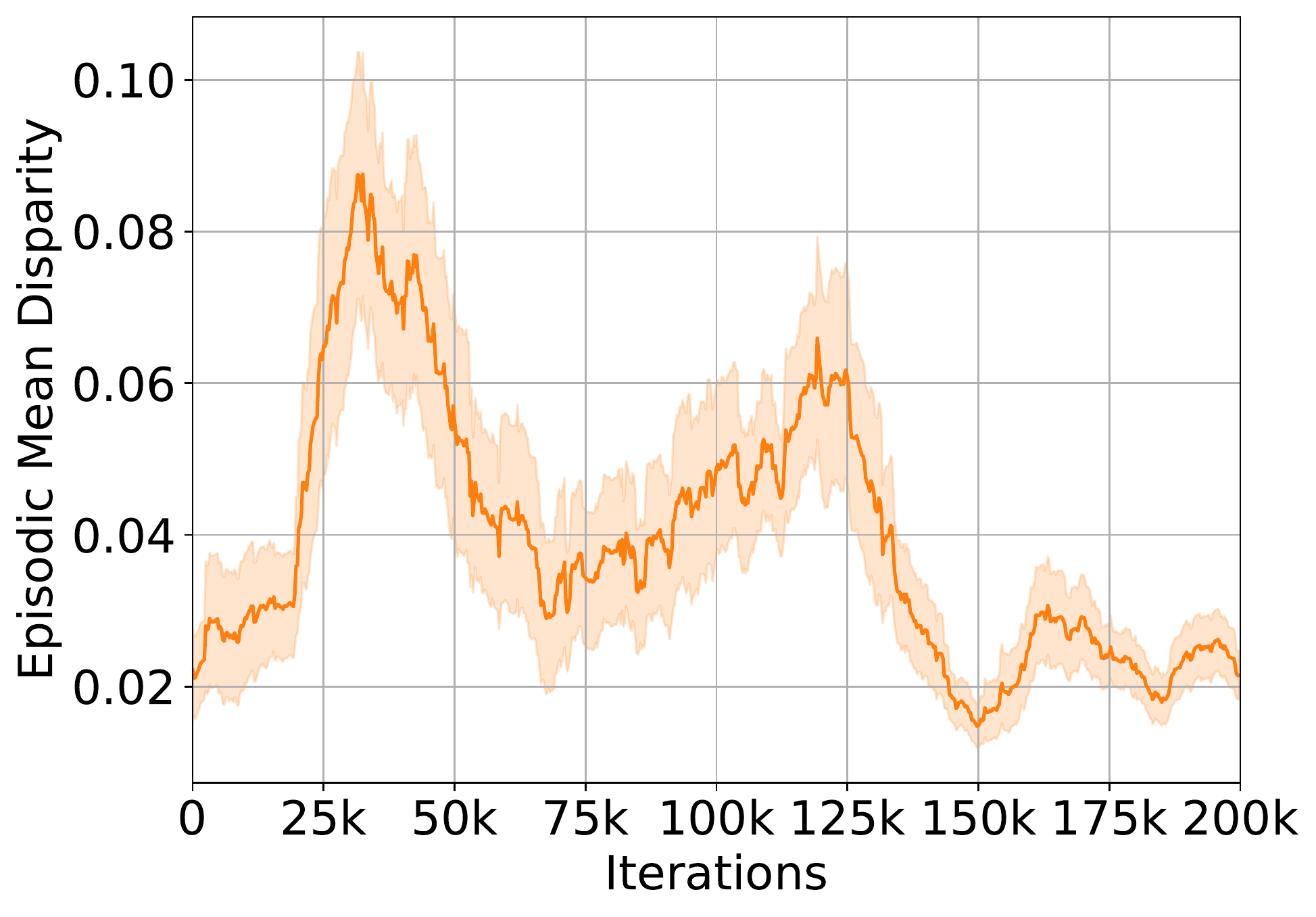}
      \caption{Equalized Odds}
    \end{subfigure}
    \caption{\drl 100-step sliding mean \& std for the setting in \cref{fig:2}.}
    \label{fig:drl_curve1}
\end{figure}

\begin{figure}[H]
    \centering
    \begin{subfigure}{\textwidth}
      \centering
      \includegraphics[width=0.26\textwidth]{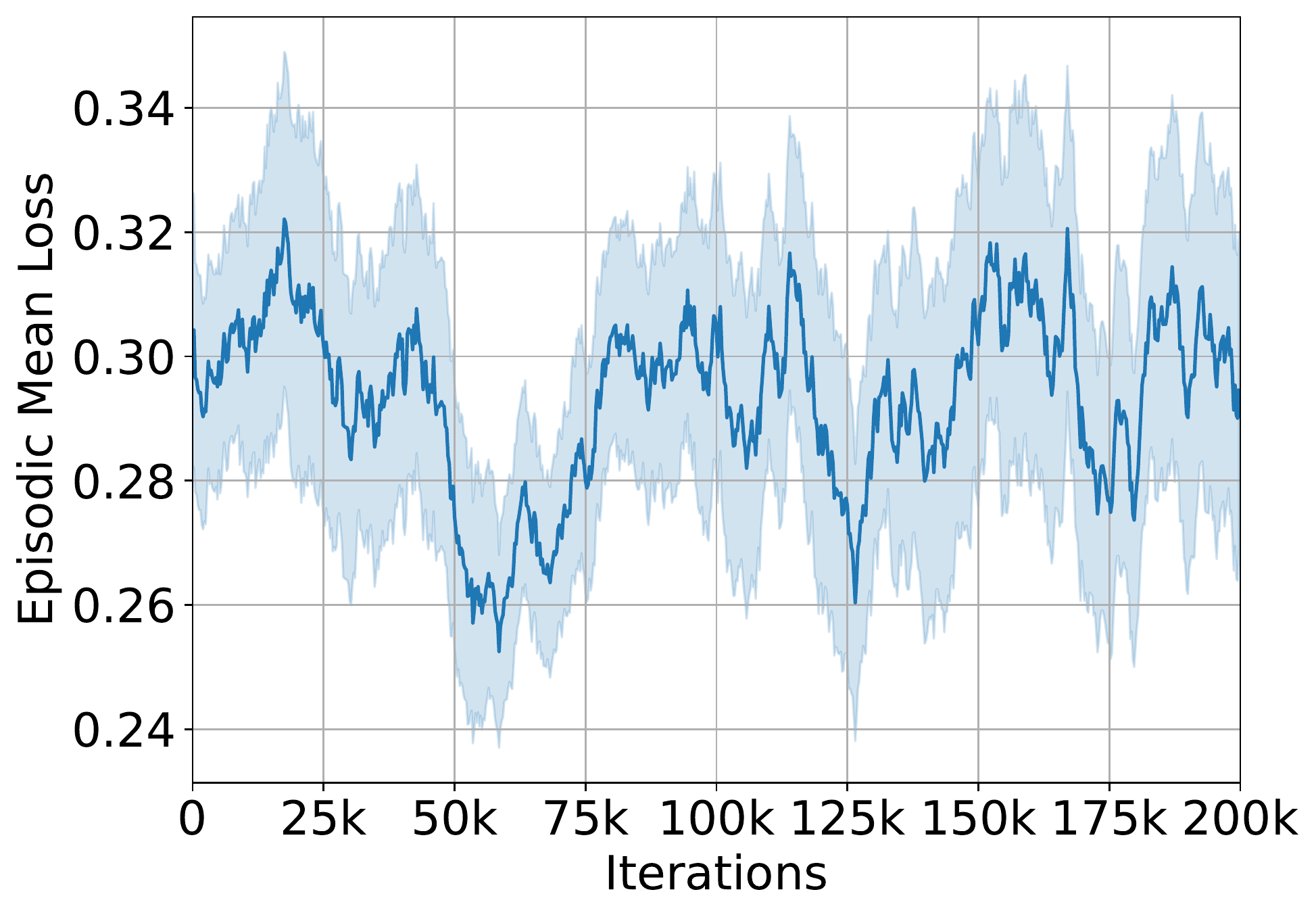}
      \includegraphics[width=0.26\textwidth]{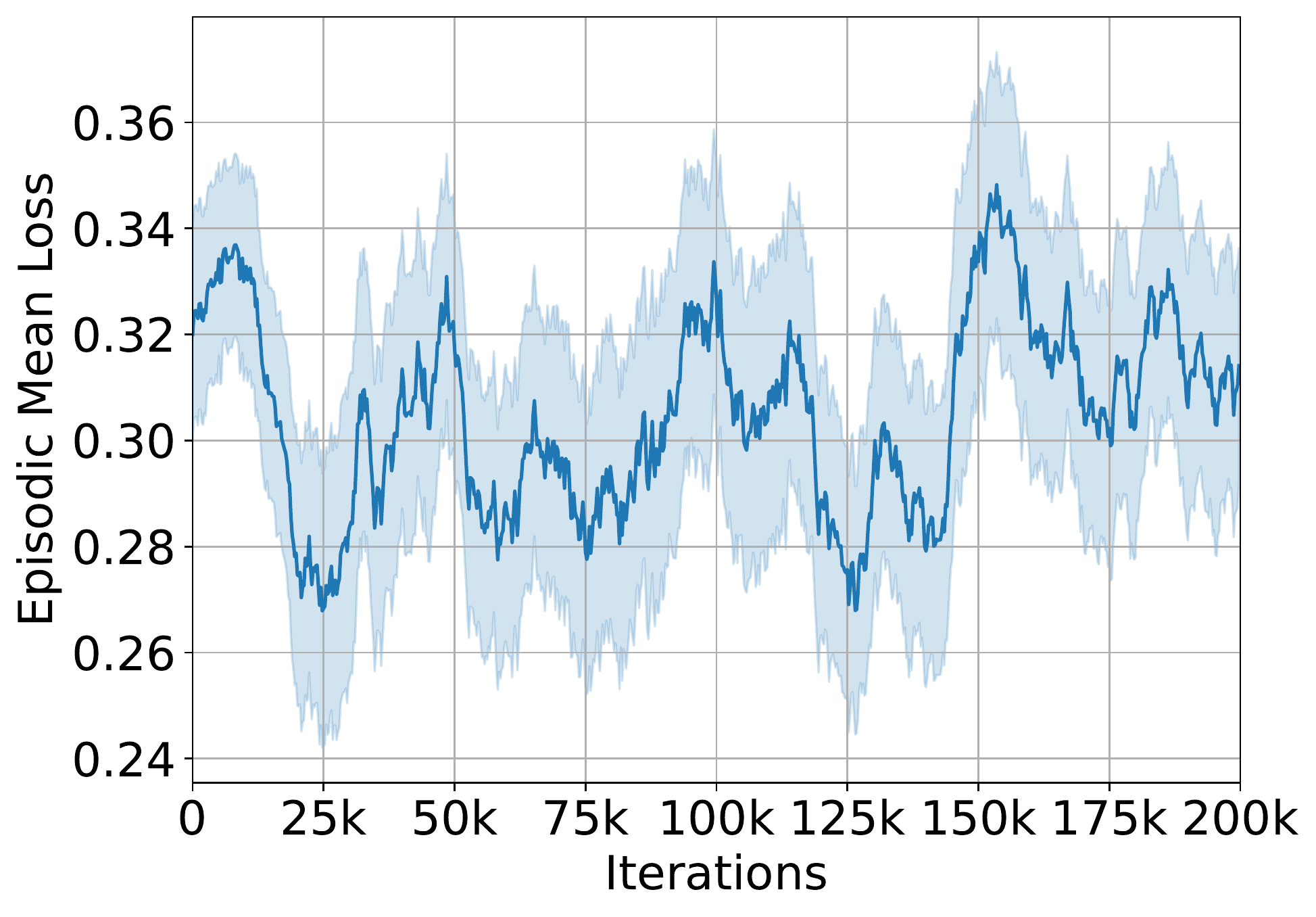}
      \includegraphics[width=0.26\textwidth]{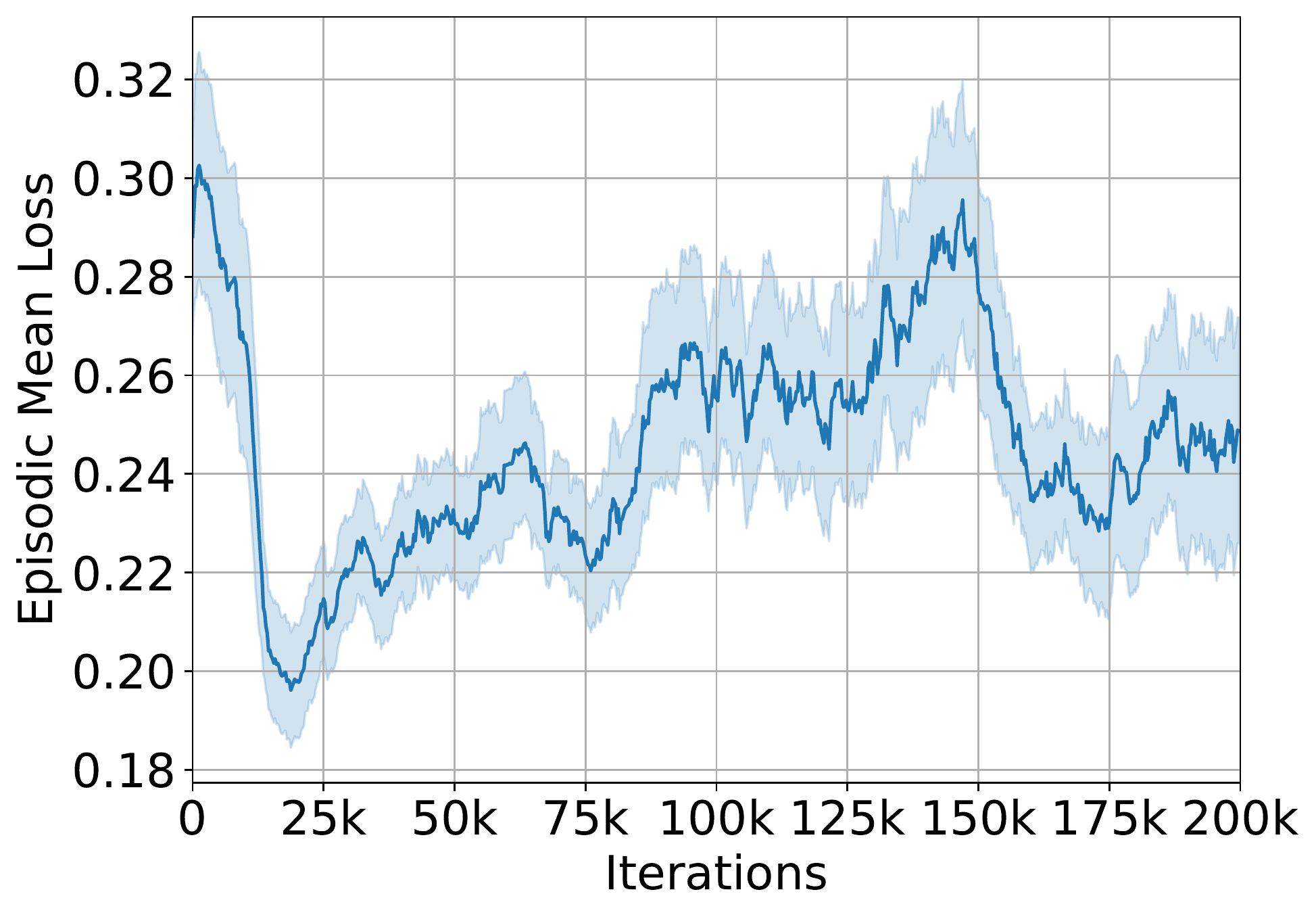}
    \end{subfigure}
    \begin{subfigure}{0.26\textwidth}
      \includegraphics[width=\textwidth]{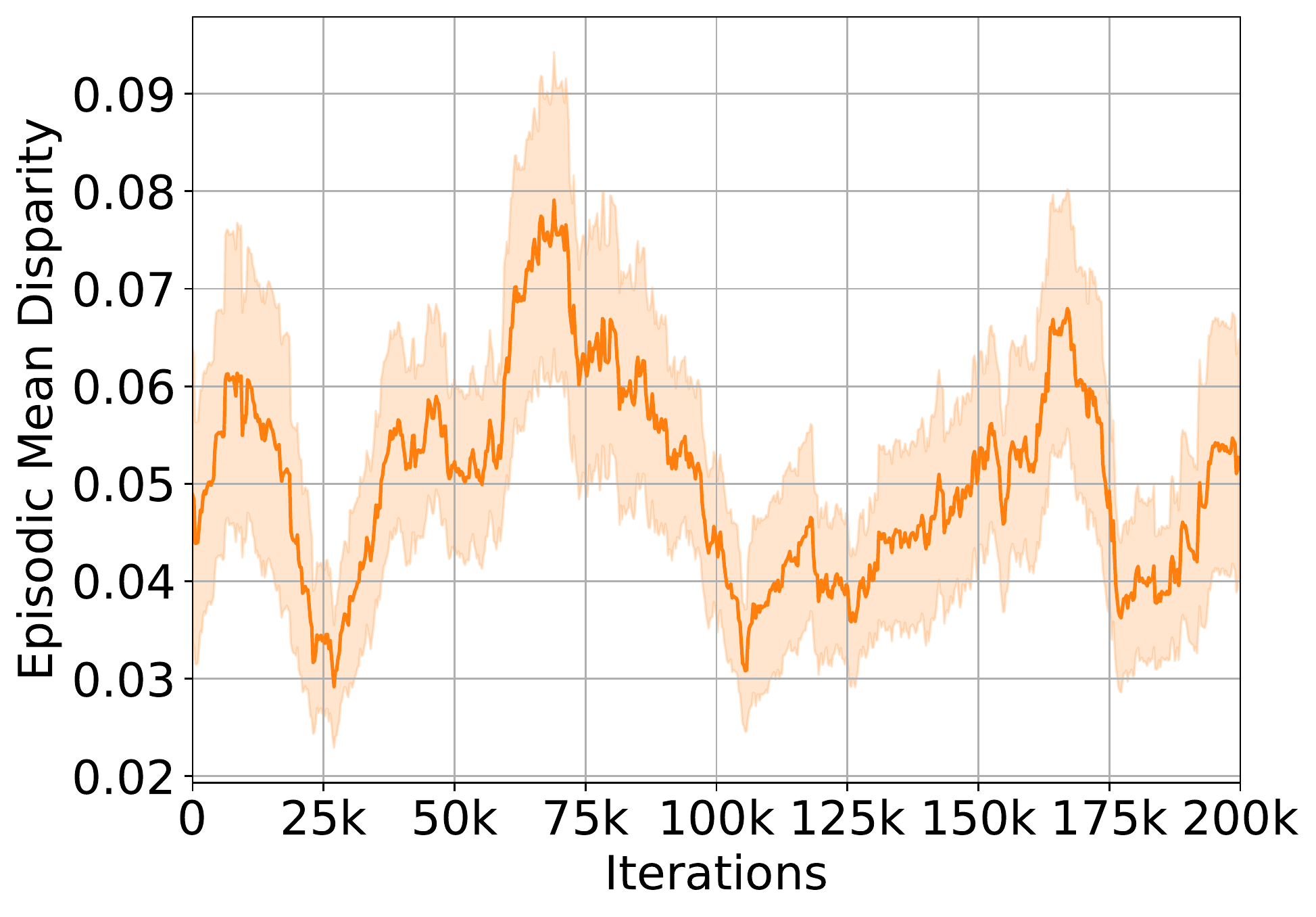}
      \caption{Demographic Parity}
    \end{subfigure}
    \begin{subfigure}{0.26\textwidth}
      \includegraphics[width=\textwidth]{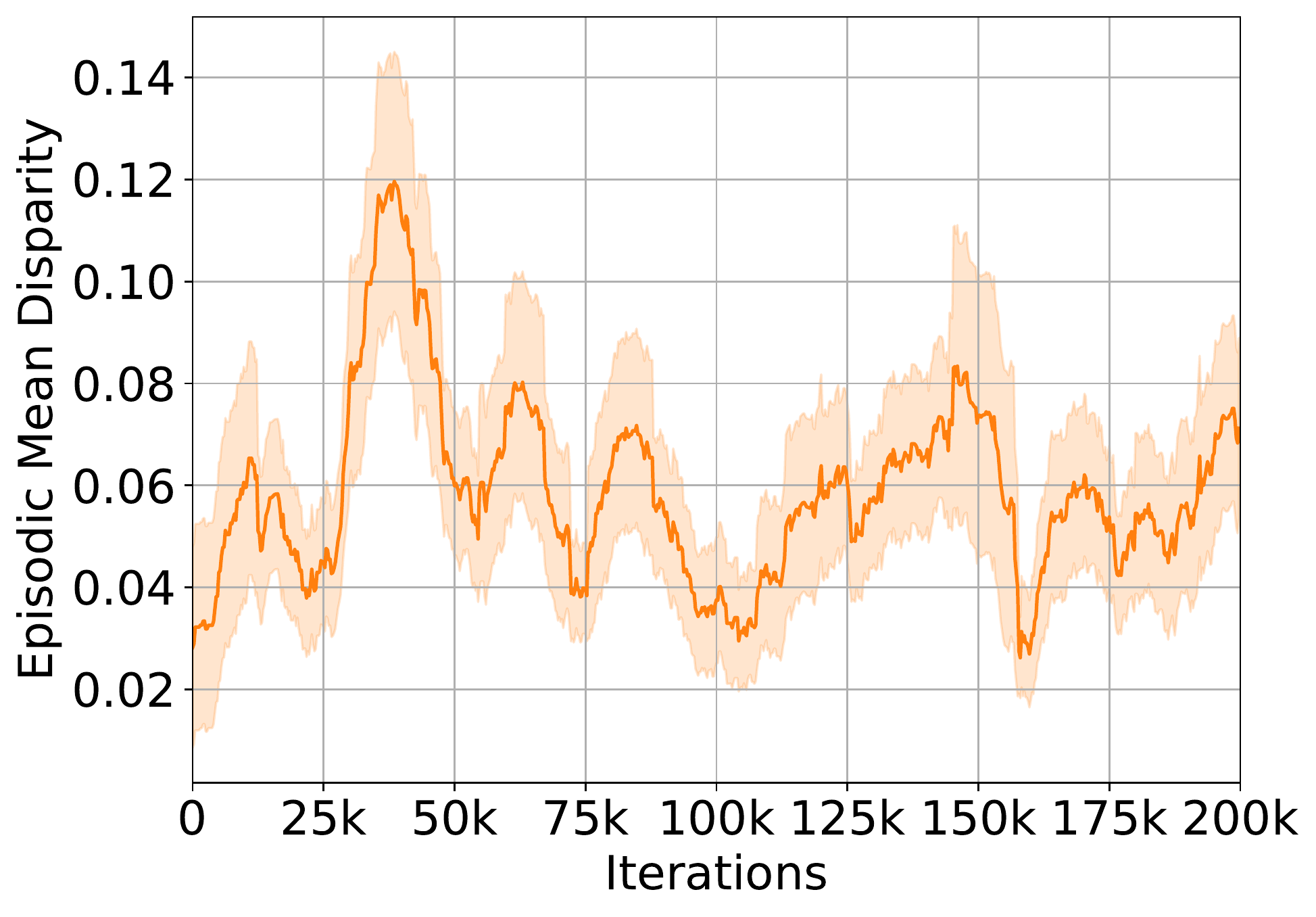}
      \caption{Equal Opportunity}
    \end{subfigure}
    \begin{subfigure}{0.26\textwidth}
      \includegraphics[width=\textwidth]{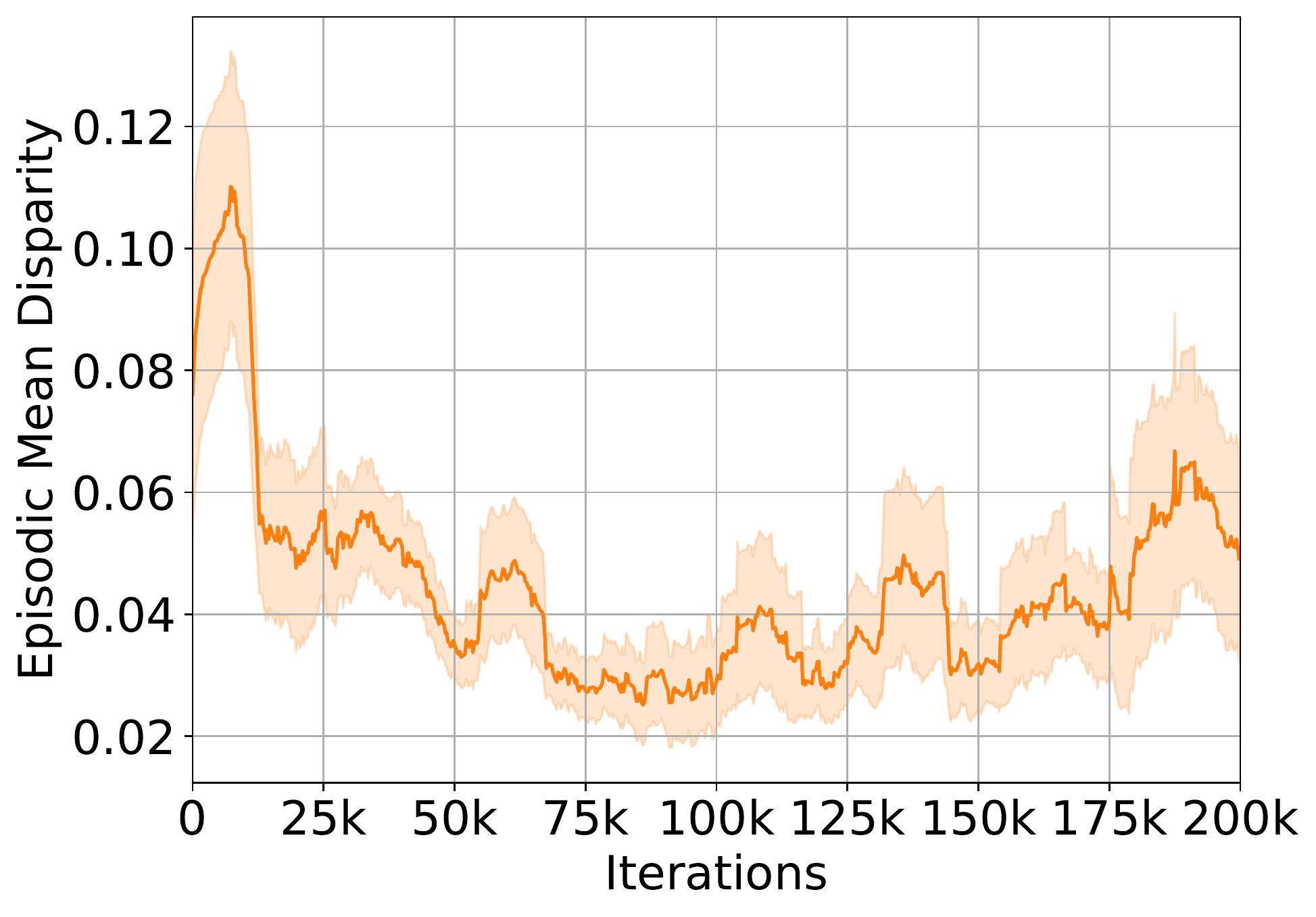}
      \caption{Equalized Odds}
    \end{subfigure}
    \caption{\drl 100-step sliding mean \& std for the setting in \cref{fig:4}.}
    \label{fig:drl_curve2}
\end{figure}

\newpage
\subsection{Reduction of Utility}
\begin{figure}[h]
    \centering
    \includegraphics[width=0.5\linewidth]{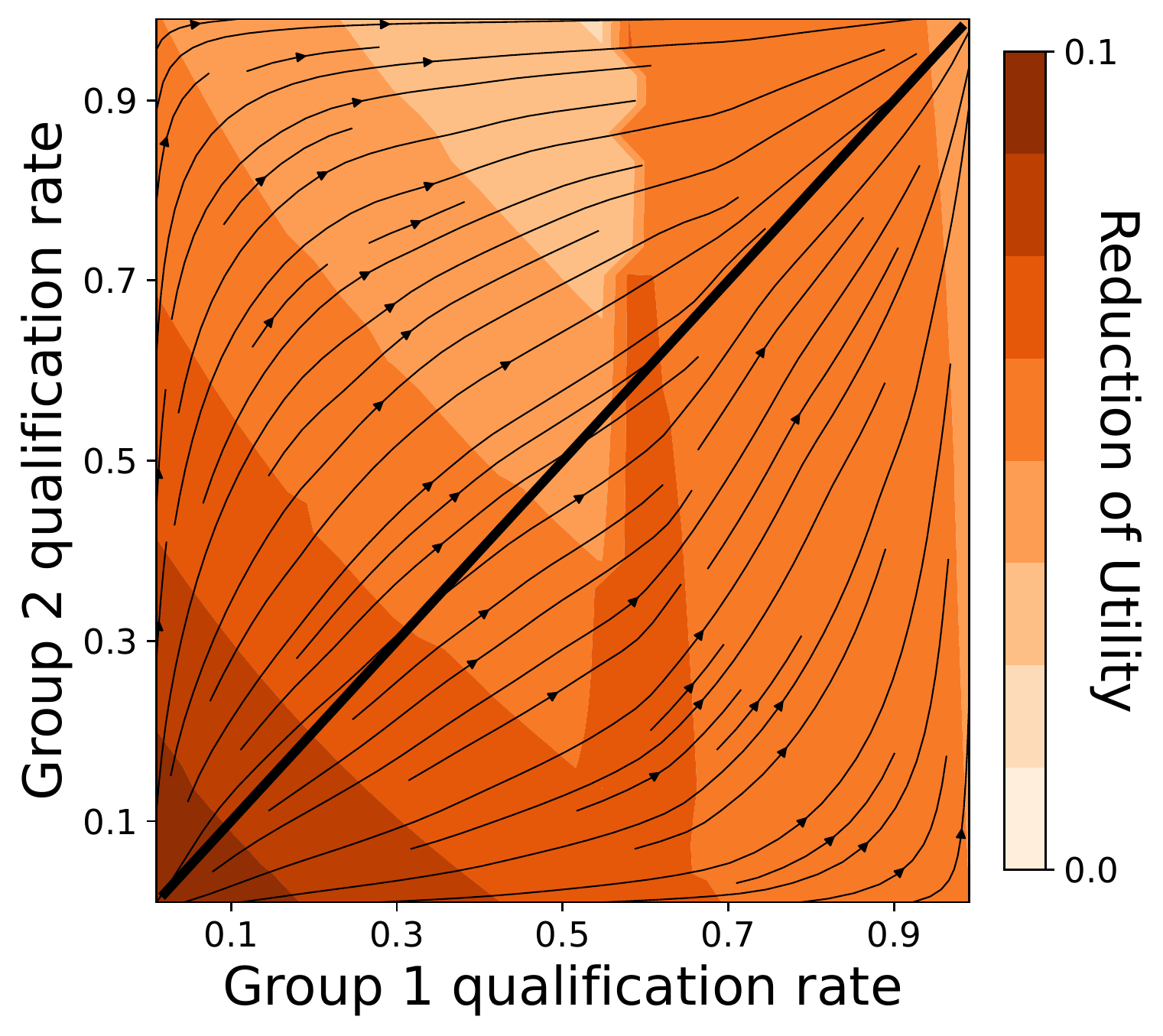}
    \caption{The figure depicts the short-term impact on utility of the \texttt{UCBFair} algorithm compared to a greedy baseline agent that operates without fairness constraints. In this experiment, both algorithms were designed to optimize the fraction of true-positive classifications, but only \texttt{UCBFair} was subject to the additional constraint of demographic parity. As the results indicate, the \texttt{UCBFair} algorithm experiences a reduction in utility compared to the greedy baseline, but it is able to drive the system towards a state that is preferable in the long term.}
    \label{fig:reduction_of_utility}
\end{figure}

\end{document}